\documentclass{article}
\usepackage{iclr2027_conference}
  \usepackage[T1]{fontenc}
  \usepackage{tgtermes}

\iclrfinalcopy 
\usepackage{amsmath,amssymb,amsthm}
\usepackage{graphicx}
\usepackage{wrapfig}
\usepackage{tikz}
\usepackage{booktabs}
\usepackage{longtable}
\usepackage{array}
\usepackage{algorithm}
\usepackage{algpseudocode}
\usepackage{multirow}
\usepackage{xcolor}
\usepackage[colorlinks=true,linkcolor=blue!55!black,
            citecolor=blue!55!black,urlcolor=blue!55!black]{hyperref}
\usepackage[capitalise,noabbrev]{cleveref}
\crefname{algorithm}{Algorithm}{Algorithms}
\Crefname{algorithm}{Algorithm}{Algorithms}
\algrenewcommand\algorithmicrequire{\textbf{Input:}}
\algnewcommand{\TEline}{\State\textcolor{blue!55!black}{$\star$}\ }
\usepackage{microtype}
\usepackage{enumitem}
\usepackage{subcaption}

\graphicspath{{figures/}}

\newcommand{\archive}{A_t}
\newcommand{\rims}{\mathrm{rim}}
\newcommand{\gated}{\mathrm{gated}}
\newcommand{\activeset}{\mathrm{active}}
\newcommand{\wtopo}{w_{\mathrm{topo}}}
\newcommand{\Blive}{B^{\mathrm{live}}_t}
\newcommand{\Shom}{S_{\mathrm{hom}}}
\newcommand{\Sloc}{S_{\mathrm{loc}}}
\newcommand{\ghat}{\hat{g}}
\newcommand{\eps}{\varepsilon^{*}}
\newcommand{\GEsub}{\mathrm{GE}}
\newcommand{\TEsub}{\mathrm{TE}}
\newcommand{\TE}{\textsc{TopoExplore}}
\newcommand{\GE}{\textsc{Go-Explore}}

\theoremstyle{plain}
\newtheorem{theorem}{Theorem}
\newtheorem{lemma}{Lemma}
\theoremstyle{definition}

\newenvironment{restatethm}[2]{%
  \par\addvspace{\medskipamount}\noindent
  \textbf{Theorem #1 (#2, restated).}\begingroup\itshape\ }%
  {\endgroup\par\addvspace{\medskipamount}}

\title{\TE: Homology as an Exploration Signal}
\author{Jason Carlson\\
\texttt{jcarlson212@gmail.com}}
\date{}

\begin{document}
\maketitle

\begin{abstract}
Exploration signals in reinforcement learning are currently computed
from what an agent has seen: visitation counts, density
estimates, or a model's prediction error at individual states. None
of these report the holes in the visitation space. To capture them, we compute the persistent homology
of the agent's own archive of visited states during exploration and
turn each detected class---an enclosed region the archive surrounds
but has not entered---into a selection bonus concentrated on the
archived states from which entry is possible, gated by an attempt
counter that retires candidate entrances and prunes sealed
structures. The method, \TE{}, is \textsc{Go-Explore} plus one
additive term, so the comparison with \textsc{Go-Explore} isolates that term. On the $189$ held-out worlds of the open-source TopoGym benchmark
of topologically varied hard-exploration environments, \TE{} finds
the goal in $167$ worlds against $152$ for \textsc{Go-Explore} and,
where both find it, with a median $126$k steps against $186$k.
On an environment that stress-tests far-apart chambers that
create holes in the archive space, \TE{} enters all six or eight
chambers across $5$ seeds, against $1$ and $2$ of $5$ seeds for
\textsc{Go-Explore}. On
Montezuma's Revenge, built over the room graph, the term helps
whenever the archive surrounds a room it has not entered, puts
33--45\% of the selections on the surrounding rooms while it is
active, and enters the room sooner in five of six cases; such
enclosures are rare on this game, so the score is unchanged. Beyond
discrete, restorable state spaces, the same signal drives an embodied
explorer that walks to every target it selects and explores from
depth alone; a vision-language-action navigation policy fine-tuned on
its trajectories succeeds $58\%$ more often on unseen scenes than one
fine-tuned on shortest-path oracle demonstrations.

\end{abstract}

\section{Introduction}
\label{sec:intro}

Hard exploration became an important problem in deep reinforcement
learning when DQN, after succeeding on many Atari games
\citep{DBLP:journals/jair/BellemareNVB13} where exploration is not
the bottleneck, scored nothing on Montezuma's Revenge
\citep{DBLP:journals/nature/MnihKSRVBGRFOPB15}, and on Pitfall!,
under sticky-action evaluation, algorithms without demonstrations
scored at most zero until \GE{}
\citep{DBLP:journals/corr/abs-1901-10995,DBLP:journals/nature/EcoffetHLSC21}.
The decade since has mostly asked what an agent should be rewarded
for seeing: rare states, by counts and their density surrogates
\citep{DBLP:conf/nips/BellemareSOSSM16,DBLP:conf/icml/OstrovskiBOM17,DBLP:conf/nips/TangHFSCDSTA17};
badly predicted states, by curiosity
\citep{DBLP:conf/icml/PathakAED17,DBLP:conf/iclr/BurdaESK19};
both at once with episodic novelty and tens of billions of frames
\citep{DBLP:conf/iclr/BadiaSVGPKTAPBB20,DBLP:conf/icml/BadiaPKSVGB20}.
All are statistics of the state the agent is in --- a count, a
density, a prediction error --- and all have the weakness of the
signal being local in state and in time. A model that
predicts a corridor well says nothing about the room it encloses, and an agent driven by a bonus such as RND's can suffer
detachment, drifting away from a frontier whose bonus it has partly
consumed and never finding its way back
\citep{DBLP:journals/corr/abs-1901-10995,DBLP:journals/nature/EcoffetHLSC21},
so the entrance to a sealed
region earns no more bonus than any wall, with nothing in the signal
proportional to how many states lie behind it.

\GE{} \citep{DBLP:journals/corr/abs-1901-10995,DBLP:journals/nature/EcoffetHLSC21}
changed how the field approaches exploration by introducing a simple
archive of visited states on which an exploration signal can act;
selecting cells by visit counts, missing neighbours and
domain-knowledge cells rather than a learned bonus, then robustifying
the trajectories it found, it surpassed human performance on both
games. Intelligent
Go-Explore \citep{DBLP:conf/iclr/LuHC25} replaces these heuristics with a
foundation model that reads the archive and chooses which state to
return to, which actions to explore from it, and which new states
are interesting enough to keep, a model call at every one of these
decisions, which is computationally expensive.
The archive remains a useful place for an exploration signal; what
it lacks is a direct, global signal of the shape of the environment
seen so far.

Persistent homology provides such a signal: the first
homology of the visited set counts exactly the regions the agent has walked around
but not into, and in higher dimensions the cavities in its coverage.
Each such region is a class, and it comes paired with a
persistence, a magnitude that grows with the region's radius
and measures how robust the region is to perturbation; computing it
needs neither a model nor a gradient that might vanish. This paper uses it as one of the main
exploration signals by treating each detected class as a hypothesis,
\emph{there is an enclosed region here that is perhaps enterable},
and giving it a selection bonus, its \emph{mass}, spread over the
archived cells best placed to test it. Each candidate cell's share
shrinks with every failed attempt; once all are exhausted, the class
is pruned and marked sealed. We use \GE{} because its archive is exactly what the
homology signal would be computed on anyway, and because it is a
proven method whose score we can change by one term.

\paragraph{Primary contributions.}
\begin{enumerate}[leftmargin=1.8em,itemsep=1pt]
  \item \textbf{The first use of persistent homology as an
    exploration signal in reinforcement learning.} The $H_1$ classes
    of the archive of visited states decide where the agent explores
    from, with no map of the environment and no planner
    (\Cref{sec:method}); prior uses of homology in robotics and
    learning are reviewed in \Cref{sec:related}.
  \item \textbf{Evidence that it explores enclosed regions faster when
    it is there.} On 189 held-out TopoGym worlds, where both find the
    goal, \TE{} finds it faster than \GE{} (median speed-up
    $1.33\times$, $1.97\times$ with the curvature term), and it finds
    it in more worlds (\Cref{sec:results:chambers}); on Montezuma's Revenge,
    built over the room graph, it entered an encircled room
    186k--853k frames sooner than the same seed without it in five of
    six paired cases, and on Pitfall!,
    which has no enclosures, the term costs little
    (\Cref{sec:results:montezuma}, \Cref{app:pitfall}).
  \item \textbf{The signal as a data generator for embodied
    policies.} Walking to each target instead of restoring states, the same explorer
    collects navigation trajectories from depth alone on which a
    vision-language-action policy succeeds $58\%$ more often on
    unseen scenes than on oracle demonstrations
    (\Cref{sec:results:embodied}).
\end{enumerate}

\section{Background and Related Work}
\label{sec:related}

We outline prior exploration signals in more detail, give the
background on persistent homology the method rests on, and then
survey how topological data analysis has been used in the field.

\paragraph{Counts and archives.}
Count-based bonuses and their density-model surrogates
\citep{DBLP:conf/nips/BellemareSOSSM16,DBLP:conf/icml/OstrovskiBOM17,DBLP:conf/nips/TangHFSCDSTA17}
score states individually and are blind to the shape of the visited
region. \GE{}
\citep{DBLP:journals/corr/abs-1901-10995,DBLP:journals/nature/EcoffetHLSC21}
keeps the visited cells in an archive and alternates returning to
a selected cell with exploring from it, the cell drawn with
probability proportional to a score over per-cell visitation
attributes plus a neighbour term; our method builds directly on it and changes only its
score, whereas Intelligent Go-Explore \citep{DBLP:conf/iclr/LuHC25}
hands selection, exploration and archiving alike to a foundation
model's judgement over the archive. Frontier
exploration \citep{DBLP:conf/cira/Yamauchi97} drives
to the boundary between known-free and unknown space. Novelty search
\citep{DBLP:journals/ec/LehmanS11} shares the archive but scores
members by behavioural distance, and MAP-Elites
\citep{DBLP:journals/corr/MouretC15} bins them by behaviour and keeps
the best performer per bin; neither scores by the geometry the
archive collectively describes.

\paragraph{Distribution objectives and curiosity.}
Maximum-entropy exploration \citep{DBLP:conf/icml/HazanKSS19} and
skill discovery \citep{DBLP:conf/iclr/EysenbachGIL19} optimise the
distribution of visited states and ignore
whether its support has holes. Prediction-error
curiosity \citep{Schmidhuber1991}, ICM
\citep{DBLP:conf/icml/PathakAED17}, RND
\citep{DBLP:conf/iclr/BurdaESK19}, disagreement
\citep{DBLP:conf/icml/PathakG019} and information gain
\citep{DBLP:conf/nips/HouthooftCCDSTA16} respond only to the present: high
where the model is currently wrong, and carrying nothing about a
region the agent has surrounded. Agent57's budgets
\citep{DBLP:conf/icml/BadiaPKSVGB20} are three orders of magnitude beyond ours;
we use RND and ICM as references at a matched step budget, though
at a far larger computational budget (\Cref{sec:results:compute}).

\paragraph{Complexes, \texorpdfstring{$H_1$}{H1}, and persistence.}
\label{sec:related:background}

A complex glues vertices, edges, faces and higher-dimensional cells,
the last of which we will not need, along their boundaries; we use only simplicial
and cubical ones. Over $\mathbb{F}_2$ the boundary map $\partial_n$
sends each $n$-cell ($n$-dimensional cell) to the sum of its
$(n-1)$-faces, $\partial_n\partial_{n+1}=0$, and the first
homology group $H_1(K)=\ker\partial_1/\operatorname{im}\partial_2$ is,
informally, closed loops modulo those that bound filled regions
\citep{EdelsbrunnerHarer2010}.
We use two standard constructions, chosen according to the structure of the
state space. For a grid or lattice, we use a cubical complex. Given a
set of lattice cells $A\subset\mathbb{Z}^2$, let $K(A)$ contain the unit
squares indexed by $A$ together with all of their edges and vertices. In the
planar case, each independent class in $H_1(K(A))$ corresponds to a bounded
connected component of the complement, i.e., an enclosed pocket of cells not
in $A$.
For a finite point set $P$ in a general metric space, the standard
construction is the Vietoris--Rips complex $\mathrm{VR}_{\varepsilon}(P)$,
which fills every subset of points whose pairwise distances are at most
$\varepsilon$. In Euclidean space, write $P_r$ for the union of the
closed balls of radius $r$ about the points of $P$, and
$\check{C}_r(P)$ for the \v{C}ech complex, which fills every subset of
points whose balls share a common point. By the nerve lemma
\citep{Borsuk1948}, $\check{C}_r(P)$ has the homotopy type of $P_r$,
and hence its homology, whereas Vietoris--Rips only approximates it,
sitting between two \v{C}ech complexes, $\check{C}_{\varepsilon/2}(P)
\subseteq \mathrm{VR}_{\varepsilon}(P) \subseteq \check{C}_{\varepsilon}(P)$.
We use the alpha complex, which has the same homotopy type as
$\check{C}_r(P)$ \citep[Thm.~3.2]{DBLP:journals/dcg/Edelsbrunner95} but in two and
three dimensions is much faster to build than either. Increasing the radius gives a nested
family of complexes, a filtration. Grid experiments use the cubical complex; the embodied explorer uses
the alpha complex, built with GUDHI \citep{DBLP:conf/icms/MariaBGY14}.
Persistent homology removes the need to rely on a single scale by
tracking homology classes throughout a filtration. A class that first appears
at scale $\varepsilon_{\mathrm b}$ and becomes filled at
$\varepsilon_{\mathrm d}$ defines an interval
$[\varepsilon_{\mathrm b},\varepsilon_{\mathrm d})$ with persistence
$\operatorname{pers}=\varepsilon_{\mathrm d}-\varepsilon_{\mathrm b}$.
Hence persistence measures how long a hole survives as the complex is
gradually thickened. In our method, $H_1$ identifies where unexplored
regions occur, while persistence provides a scale-dependent measure of how
substantial those structures are, and how robust they are to
perturbations of the space. \citet{Carlsson2009} provides a survey of the subject and
\citet{EdelsbrunnerHarer2010} is a full introduction for more
details.

\paragraph{Topological data analysis.}
The closest lineage is robotic coverage and exploration: \citet{SilvaGhrist2007}
certify, from sensor connectivity alone and without coordinates, that
a static sensor deployment leaves no hole;
\citet{DBLP:conf/iros/KimBGK13} send robots along distinct homology
classes of paths around obstacles toward the unexplored region, and \citet{DBLP:journals/trob/BhattacharyaGK15} plan
paths through such a map by their homology classes. The robotic
works assume a map with coordinates, so that any point can be
addressed, and a planner that can reach any point of it; ours has
neither. In
learning, persistence has entered as an input or a regulariser
\citep{DBLP:conf/nips/HoferKNU17,DBLP:conf/aistats/CarriereCILRU20,DBLP:conf/icml/MoorHRB20}
and as a complexity measure of the weights that decides when training
stops \citep{DBLP:conf/iclr/RieckTBMHGB19}, never as a signal for
where an agent explores, and in RL ``topological'' has meant a
graph over past experience \citep{DBLP:conf/iclr/SavinovDK18,DBLP:conf/iclr/Hong0LPA22},
which captures connectivity but not whether the visited set encloses
anything.

\section{Method}
\label{sec:method}

\TE{} is \GE{} plus one additive term in the cell score. The archive,
the return mechanism, the exploration rollout, and the base score are
inherited unchanged, so every comparison with \GE{} in
\Cref{sec:results} isolates that one term.

\paragraph{Setup.}
An agent maintains an archive $\archive$ of discretised cells, each
storing the best trajectory that reached it and per-cell counters. At
each iteration it draws a cell $c$ with probability
$\mathrm{Score}(c)/\sum_{c' \in \archive} \mathrm{Score}(c')$, returns
to it, explores for a fixed number
of steps, and adds every visited cell to the archive. \GE{}'s
score is
\begin{equation}
  \mathrm{CellScore}(c) \;=\; M(c)\Big[
  \sum_{a \in \mathrm{attrs}} \Big(w_a
  \Big(\tfrac{1}{v(c,a) + \epsilon_1}\Big)^{p_a} + \epsilon_2\Big)
  \;+\; \sum_{n} w_n \big(1 - \mathrm{HasNeighbor}(c, n)\big)
  \;+\; 1\Big],
  \label{eq:cellscore}
\end{equation}
with $\mathrm{attrs} = \{\text{times chosen},\ \text{times seen},\
\text{times chosen since new}\}$ the visitation attributes of
\citet{DBLP:journals/corr/abs-1901-10995}, the last being the number
of times the cell was chosen since exploration from it last found a
new or better cell, the neighbour term counting lattice
neighbours absent from the archive (on Montezuma's Revenge also the
cell at the same position holding one more key), and $M(c)$
\GE{}'s level multiplier, $1$ outside Montezuma's Revenge.

\subsection{The archive complex and its \texorpdfstring{$H_1$}{H1} classes}
\label{sec:method:complex}

Let $K_t$ be the complex of \Cref{sec:related:background} over
$\archive$: cubical on a lattice, where its $H_1$ classes are exactly
the components of unvisited space enclosed by visited space, and
the alpha complex of the surface points the embodied explorer has
seen (\Cref{app:embodied}). Each live
class $\gamma$ has $\rims(\gamma)$, the archived cells that are
$4$-neighbours of the enclosed unarchived region: precisely the
archived cells from which entry may be possible. Its magnitude
$b(\gamma)$ is the enclosed region's inradius, which is also the
persistence of the class under thickening.

\subsection{Gating, attempt counts, and retirement}
\label{sec:method:gates}

\begin{wrapfigure}{r}{0.3\linewidth}
\vspace{-12pt}
\centering
\begin{tikzpicture}[x=0.078\linewidth,y=0.078\linewidth,font=\scriptsize]
  \definecolor{vis}{HTML}{D6E6F4}\definecolor{wall}{HTML}{7A7A7A}
  \definecolor{rimc}{HTML}{F6C9A8}\definecolor{door}{HTML}{D55E00}
  \foreach \ox/\rim/\lab in {0/1/(a) model-free,6.8/0/(b) oracle} {
    \begin{scope}[shift={(\ox,0)}]
      \foreach \x in {0,...,5} \foreach \y in {0,...,5}
        \fill[vis] (\x,\y) rectangle ++(1,1);
      \ifnum\rim=1
        \foreach \x/\y in {1/0,2/0,3/0,4/0,0/1,0/2,0/3,0/4,5/1,5/2,5/3,5/4,1/5,3/5,4/5}
          \fill[rimc] (\x,\y) rectangle ++(1,1);
      \fi
      \foreach \x/\y in {1/1,2/1,3/1,4/1,1/2,4/2,1/3,4/3,1/4,3/4,4/4}
        \fill[wall] (\x,\y) rectangle ++(1,1);
      \fill[white] (2,2) rectangle (4,4);
      \fill[door] (2,4) rectangle ++(1,1);
      \draw[white,line width=0.3pt] (0,0) grid (6,6);
      \node[white,inner sep=0pt,font=\tiny] at (2.5,4.5) {$\star$};
      \node[font=\tiny] at (3,3) {$R$};
      \draw[black,densely dashed,line width=0.45pt] (0.5,0.5) rectangle (5.5,5.5);
      \node[anchor=north,inner sep=1.5pt,font=\tiny] at (3,0) {\lab};
    \end{scope}
  }
\end{tikzpicture}
\caption{Where the bonus goes. (a) Model-free: the whole rim
(orange), the visited cells bordering $R$; attempts find the door
($\star$). (b) Oracle, $g$ known: the door alone. Dashed: $\gamma$.}
\label{fig:rim}
\vspace{-10pt}
\end{wrapfigure}
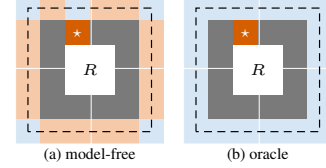
A rim cell is useful only if some action from it enters the enclosed region.
Define $g(c,\gamma) = 1$ if it does and $0$ otherwise, and
$\gated(\gamma) = \{c \in \rims(\gamma) : g(c,\gamma) = 1\}$ for
the true entrances. If $g$ were known the bonus would go to
$\gated(\gamma)$ directly, the oracle mode of \Cref{app:theory}; a
partial version of this is what the texture arm supplies when an
observed entrance affordance narrows a class's support
(\Cref{sec:method:score}). The agent in general has no model of the
environment's dynamics and cannot evaluate $g$; it can only attempt entry
and count the attempts. We call this \emph{model-free}, as opposed to
an oracle that knows $g$ (\Cref{app:theory}); every run here is
model-free: observations may narrow the rim (\Cref{app:impl:rim}),
but entrances are found by trial. It keeps an attempt count $u(c,\gamma)$, incremented on every
selection of $c$ for $\gamma$ regardless of outcome, and weights the
cell by
\begin{equation}
  \ghat(c, \gamma) \;=\; \big(1 + u(c,\gamma)\big)^{-p_u},
  \qquad
  \activeset(\gamma) \;=\; \{\, c \in \rims(\gamma) : u(c,\gamma) < u_{\max} \,\}.
  \label{eq:ghat}
\end{equation}
A cell leaves the active set after $u_{\max}$ attempts, and a class
whose active set is empty is pruned and reported as sealed. Note that
a decoy, a class around a solid and so with no true entrance
(\Cref{app:theory}), absorbs at most $|\rims(\gamma)|\,u_{\max}$
bonus draws before it is pruned (\Cref{app:theory}).

\subsection{The topological term}
\label{sec:method:score}

With $\psi(c,\gamma) = \ghat(c,\gamma)/\sum_{c' \in
\activeset(\gamma)} \ghat(c',\gamma)$ the share of a class's \emph{mass}, its total bonus
$\wtopo(t)\,b(\gamma)^{p_{\mathrm{topo}}}$, that rim cell $c$ receives,
\begin{equation}
  \Shom(c) \;=\; \wtopo(t)
  \sum_{\gamma \,\in\, \mathrm{live}(t)\,:\, c \in \activeset(\gamma)}
  b(\gamma)^{p_{\mathrm{topo}}} \; \psi(c, \gamma),
  \label{eq:shom}
\end{equation}
and the cell score becomes
\begin{equation}
  \mathrm{Score}_{\TEsub}(c) = M(c)\big[\mathrm{CellScore}(c) + \Shom(c) + \Sloc(c)\big],
  \label{eq:score}
\end{equation}
where $M(c)$ is the environment's optional cell multiplier (\GE{}'s
level weight on Montezuma's Revenge) and $\Sloc$ an optional local
term, zero in \TE{} proper. We use $\wtopo(t) = \alpha|\archive|$ so that the
topological term's share of the total score,
$\delta_t = \wtopo \sum_{\gamma\ \mathrm{live}} b(\gamma) \big/
\big(\sum_{c \in \archive} \mathrm{CellScore}(c) + \wtopo \sum_{\gamma\ \mathrm{live}} b(\gamma)\big)$
stays bounded away from zero as the archive grows (\Cref{app:theory}).

\paragraph{The \TE{} algorithm.}
\label{sec:method:algorithm}
\Cref{alg:te} (\Cref{app:impl}) states the exploration phase in full;
its starred lines are the only changes from \GE{}, and with
$\alpha = 0$ the two are identical. Phase~2 is the Backward
Algorithm of \citet{DBLP:journals/corr/abs-1812-03381} as \GE{} uses
it, unchanged; \TE{} changes only which trajectories it is given.

\paragraph{What \TE{} guarantees.}
\label{sec:method:theory}
Chamber $i$ has enclosing class $\gamma_i$, magnitude $b_i$ and
door-entry probability $q_i$; $s_{\min} = 1$ and $s_{\max}$ bound
$\mathrm{CellScore}$, $\Blive = \sum_{\gamma\ \mathrm{live}} b(\gamma)$,
$\delta_{\min} \le \delta_t$ for all $t$, and $T$ is the episode
horizon. Precise statements and proofs are in \Cref{app:theory}.

\begin{theorem}[Dilution versus anytime entry, informal]
\label{thm:dilution}
Before $\gamma_i$ is detected, any method whose per-cell score is
bounded within a fixed ratio $s_{\max}/s_{\min}$, \GE{} included,
enters chamber $i$ on the next episode
with probability at most $C(T)/|\archive|$, with $C(T)$ fixed by
$T$ and the score bounds. Once $\gamma_i$ is detected and every non-entrance rim cell has been
retired, while the class is still live, \TE{} enters with probability at least
$\delta_t\,(b_i/\Blive)\,(q_i/|\gated(\gamma_i)|)$, which
$\wtopo(t) = \alpha|\archive|$ holds constant in $t$.
\end{theorem}

\begin{theorem}[Fixed-budget separation, informal]
\label{thm:separation}
Let $k$ chambers remain in an otherwise explored world of $m$ cells.
\TE{} enters all of them in at most
$\delta_{\min}^{-1}\sum_i \beta_i$ expected iterations, with $\beta_i$
counting chamber $i$'s rim and interior: linear in what is left,
independent of $m$. At twice that budget, $N^{*}$, \TE{} enters all
$k$ with probability at least $1/2$, while \GE{} enters all $k$ with
probability at most $(e N^{*}\lambda_0)^{k}$ for
$\lambda_0 = s_{\max}\bar{n}/(s_{\min}m)$, $\bar{n}$ the largest number
of archived cells within one rollout of a door: exponentially small in
$k$ whenever the explored world is much larger than what remains,
$m \gg \sum_i \beta_i$.
\end{theorem}

\begin{theorem}[Certification, informal]
\label{thm:cert}
On a finite grid, model-free \TE{} almost surely reaches a stopping
time that is observable --- no frontier cell and no live class
remain --- at which every chamber has been entered and filled and
every surviving class is a decoy pruned by gate exhaustion. It then
reports $H_1$ of the environment with a certificate: the exhausted
gate log for each surviving class, the filled interior for each
class whose region was filled.
\end{theorem}

Informally: while no hole is detected \TE{} is \GE{}; once one is,
scaling $\wtopo$ with the archive keeps a constant fraction of
selections on the hole's rim, so
the time to enter $k$ remaining chambers grows linearly in what is
left while \GE{}'s chance of entering all $k$ decays exponentially in
$k$. Eventual coverage of a finite grid holds for either method
(\Cref{lem:cover}); the separation is in the rate, and the
certificate records why each surviving hole was declared unenterable.

\paragraph{Local-signal arms.}
$\Sloc$ carries no part of the theory; two arms test what happens
when we look at local topological properties and when we give
semantic meaning to the unvisited holes. \texttt{ricci} gives a bonus to cells whose mean
incident Ollivier--Ricci curvature \citep{Ollivier2009} on the
archive graph is negative (bottlenecks such as doorways and
corridors),
to test whether a local signal speeds up the coverage outside the
holes that \Cref{thm:separation} assumes (an otherwise explored
world), so that its separation takes hold sooner. \texttt{texture} records features the agent observed at each cell
(door sprites, key possession, skull proximity on Montezuma's
Revenge) and, when a live class's active rim contains cells showing
an entrance feature, puts the class's whole mass on them, so it also acts on $\Shom$ with a
smaller, better-placed support.

\section{Experimental Setup}
\label{sec:setup}

\paragraph{TopoGym (chamber worlds).}
TopoGym is a library that contains a benchmark of 2D worlds whose
topology is controlled by construction \ificlrfinal\citep{topogym2026}\else\citep{anon2026topogym}\fi. The hold-out benchmark is TopoGym-v1's frozen
\texttt{test} split, 189 worlds (63 units $\times$ 3 layout seeds)
spanning chambers, decoys, bottlenecks, mazes, textured scenarios and
non-trivial topologies, given 1M steps per world; TopoGym
provides a fixed \texttt{tune} split, disjoint from the hold-out
split, on which every method was tuned. Three further chamber-count
families, not in the benchmark, satisfy the hypotheses of
\Cref{thm:separation} (doors more than one rollout away, an otherwise
explored world) to different degrees, OpenFieldChamberCount most
closely; each places $k$ chambers so that reaching the next costs
more than one episode.

\paragraph{Montezuma's Revenge (Atari).}
The canonical hard-exploration game \citep{DBLP:conf/nips/BellemareSOSSM16}
of the Arcade Learning Environment \citep{DBLP:journals/jair/BellemareNVB13},
with 24 rooms in level~1, several gated behind keys, and the
domain-knowledge cells of \citet{DBLP:journals/corr/abs-1901-10995}.
Phase~1 runs deterministically with snapshot restore; archival arms
get 15M agent steps (60M frames) and the intrinsic-motivation
baselines 25M (100M frames), the sum of both \TE{} phases. The
complex is built at two scales, run as separate arms on the same
seeds: within each room, on the cell lattice, and over the rooms
themselves, on the room graph the agent learns from the passages it
crosses --- the latter so that a room the archive surrounds
forms an $H_1$ class (\Cref{sec:results:montezuma}). Pitfall!,
a chain of 255 thin rooms with little enclosed structure or reward,
is run with the same campaign and reported in \Cref{app:pitfall}.

\paragraph{Embodied navigation (MultiON on HSSD).}
The embodied experiment tests whether the trajectories a
topological explorer collects are good training data for a
navigation policy.
The task is MultiON \citep{wani2020multion} in the HSSD version of
the 2024 MultiON challenge, in its single-goal projection on Habitat \citep{DBLP:conf/iccv/SavvaMPBKMZWJSL19} with
HSSD scenes \citep{khanna2024hssd}: 42 training scenes, 20 unseen
evaluation scenes, 12 goal categories held out of training. The
explorer is \TE{}'s selection rule and gate over a complex of seen
surface points built from depth alone, with no archive to restore: it walks to every
candidate it selects and keeps a per-episode map
(\Cref{app:embodied}). The policy is InternVLA-N1-System2
\citep{internvla2025n1,wei2026ground} fine-tuned with LoRA \citep{hu2022lora}
under one recipe on one of six training sets: oracle shortest-path
demonstrations, topological-explorer episodes, two other explorers
on the same scenes and starts, and two mixtures
(\Cref{app:embodied}). The headline metric is success, stopping within
1\,m of the goal, over $\sim$325 evaluation episodes per seed and split, five
seeds per arm with seeds and episodes shared across arms.

\paragraph{Baselines and statistics.}
\GE{} is identical code with $\wtopo = 0$, so the difference from
\TE{} is exactly $\Shom$; PPO+RND
\citep{DBLP:conf/iclr/BurdaESK19} follows its reference implementation
and PPO+ICM pairs ICM's losses \citep{DBLP:conf/icml/PathakAED17} with
PPO as in \citet{DBLP:conf/iclr/BurdaEPSDE19}; frontier, random and shortest-path
planners are the embodied baselines. Aggregates are medians or
interquartile means over seeds with 95\% bootstrap intervals
\citep{DBLP:conf/nips/AgarwalSCCB21}, seeds being the unit of
replication. Every hyperparameter, its value and its provenance is
in \Cref{app:hparams}; \TE{}'s were selected once on TopoGym and
frozen before any Atari run, while RND uses its authors' values
tuned on Montezuma's Revenge (at 32 rather than 128 parallel
environments), so the in-domain tuning advantage lies with it. ICM,
never run on Atari by its authors, takes RND's PPO settings and a
TopoGym-searched intrinsic scale, and its inverse and forward losses
were weighted $0.2/0.8$ rather than the published $0.8/0.2$
(\Cref{app:hparams}).

\section{Results}
\label{sec:results}

\subsection{Topologically varied worlds}
\label{sec:results:chambers}

\begin{table}[t]
\centering
\small
\caption{Hold-out benchmark: 189 worlds, 1M steps each, methods
tuned on the disjoint \texttt{tune} split. Goal found: worlds in
which any training episode reached the goal --- a treasure chest;
each \TE{} arm is paired with
\GE{} world by world, exact sign test. Steps to first goal: over the
worlds where both found it, medians, the median per-world ratio
\GE{}/\TE{}, and worlds where \TE{} was faster / slower (sign test
$p < 10^{-3}$ for every arm). PPO+ICM's hyperparameters were set to
standard values rather than tuned (\Cref{app:hparams}), and 46 of its
worlds were lost to withdrawn cloud machines, not to failure.}
\label{tab:holdout}
{\footnotesize\setlength{\tabcolsep}{3pt}\begin{tabular}{lcccc}
\toprule
& \multicolumn{2}{c}{Goal found in training} & \multicolumn{2}{c}{Steps to first goal (both found)} \\
\cmidrule(lr){2-3}\cmidrule(lr){4-5}
Method & worlds & only TE / GE ($p$) & median TE vs GE & speed-up (faster/slower) \\
\midrule
\GE{} & 152 & --- & --- & --- \\
\TE{} (none) & 167 & 16 / 1 (0.0003) & 126k vs 186k & 1.33$\times$ (97 / 54) \\
\TE{} + ricci & 168 & 19 / 3 (0.0009) & 90k vs 180k & 1.97$\times$ (118 / 31) \\
\TE{} + texture & 168 & 17 / 1 (0.0001) & 123k vs 186k & 1.50$\times$ (101 / 50) \\
\TE{} + both & 169 & 19 / 2 (0.0002) & 85k vs 183k & 1.97$\times$ (119 / 31) \\
\midrule
PPO & 3 & \multicolumn{3}{c}{---} \\
PPO+RND & 4 & \multicolumn{3}{c}{---} \\
PPO+ICM & 49 & \multicolumn{3}{c}{143 of 189 worlds} \\
\bottomrule
\end{tabular}}

\end{table}

\begin{figure}[t]
\centering
\includegraphics[width=\linewidth]{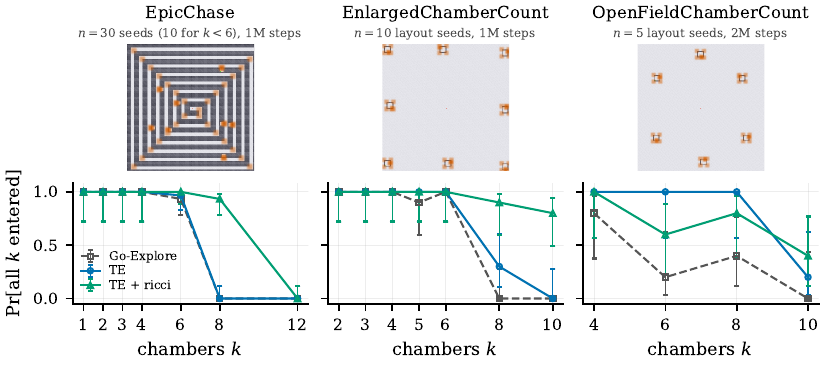}
\caption{Probability of entering all $k$ chambers within budget,
Wilson 95\% CIs, with one world of each family above its curve;
vermillion marks cells of negative Ollivier--Ricci curvature
(doorways and bottlenecks), which the ricci arm rewards. Left
to right, the walls become less consequential: a spiral whose corridors dictate the
walk, chambers on the perimeter of an open square, and the same
chambers with no reachable boundary, where nothing but the archive's
own holes marks a chamber and the theory's geometric hypotheses hold. Goal,
cost and survival curves and chamber grids in \Cref{app:chambers}.}
\label{fig:ksweep}
\end{figure}

\paragraph{The hold-out benchmark.}
\Cref{tab:holdout} is the cleanest \TE{}-over-\GE{} result in the
paper. On the roughly 150 worlds where both find the goal, \TE{}
finds it first on two thirds to four fifths of them, with a median
speed-up of $1.33\times$ for the plain term and $1.97\times$ with
the curvature term (faster on 118 of 149 worlds, slower on 31), the
medians of the raw step counts falling from $180$--$186$k to
$85$--$126$k. The distribution is right-skewed, so the ratio of means,
$2.05\times$, sits above the median. \TE{} also finds the goal in
more worlds: \GE{} in 152 of 189, every \TE{} arm in 167 to 169, and
the pairing is one-sided: plain \TE{} gains 16 worlds and loses one
($p = 3 \times 10^{-4}$), and every sign test is below $10^{-3}$
(all 63 units in \Cref{fig:app:holdout_gallery}). PPO and PPO+RND find the
goal in 3 and 4 worlds and PPO+ICM in 49 of the 143 it completed
(34\%, against 80--89\% for the archive methods), every run at the
full 1M-step budget. The archive methods explore with uniformly
random actions from the selected cell, so this compares structured
selection with random exploration against a learned policy without
selection; learning the explorer used after selection is left to
future work.

\paragraph{The $k$-sweeps.}
The three chamber-count families test \Cref{thm:separation} where its
hypotheses hold and show what the local arms add where they hold
least. Each places $k$ chambers so that a method that cannot keep
returning to distant frontiers enters all $k$ with a probability that
drops sharply in $k$, which \Cref{fig:ksweep} shows as a step. On
EpicChase the bottleneck is reaching the chambers at all, through a
spiral of tight corridors, and there the
curvature arm helps: at $k = 8$ \GE{} and plain \TE{} enter every
chamber in none of 30 seeds and \TE{}+ricci in 28, since negative curvature
marks the doorways along the wall.
EnlargedChamberCount, ten random layouts of the same design, gives
the same ordering. OpenFieldChamberCount is the family most faithful
to the theorem's assumptions: with the wall removed nothing local
remains to follow, and plain \TE{} enters every chamber in every
world up to $k = 8$ and finds the goal in 5 of 5 worlds where \GE{}
falls to 0.20--0.40 and the curvature arm to intermediate. Per-$k$
numbers, goal-found and cost curves are in \Cref{app:chambers}.

\subsection{Montezuma's Revenge: the enclosures that matter are rooms}
\label{sec:results:montezuma}

\begin{table}[t]
\centering
\small
\caption{Montezuma's Revenge, Phase 1, 60M frames. Best archived
trajectory return: median, interquartile mean (IQM)
[stratified-bootstrap 95\% CI], paired two-sided Wilcoxon against
\GE{} on shared seeds.}
\label{tab:montezuma}
\setlength{\tabcolsep}{5pt}
\begin{tabular}{lcccl}
\toprule
Method & $n$ & Median & IQM [95\% CI] & paired $p$ vs.\ \GE{} \\
\midrule
\GE{} & 200 & 31{,}400 & 31{,}629 [28{,}456, 34{,}382] & reference \\
\TE{} (plain) & 50 & 27{,}500 & 28{,}173 [23{,}308, 32{,}765] & $p = 0.699$ ($n = 50$) \\
\TE{} + texture & 200 & \textbf{36{,}000} & \textbf{35{,}057} [31{,}125, 38{,}653] & $p = 0.032$ ($n = 200$) \\
\bottomrule
\end{tabular}

\end{table}

\begin{figure}[t]
\centering
\includegraphics[width=\linewidth]{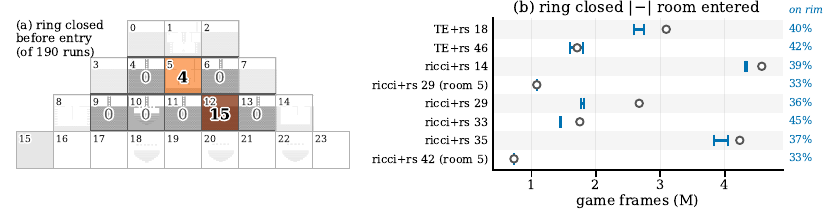}
\caption{Room-scale enclosures in Montezuma's Revenge. (a) Level~1 on the
layout learned from crossed passages; each interior room carries the
number of runs, of 190, in which the ring of visited rooms closed
around it before it was entered, over a faint screenshot of the
room. Faded rooms lie on the pyramid's edge and cannot be enclosed. (b) Each class the room-scale term
(rs) detected before its room was entered: the interval from the frame at
which the ring closed to the frame at which the room was entered, with the share of selections in that interval
that landed on the ring; the open circle is when the same seed
entered the room without the term.}
\label{fig:room12}
\end{figure}

\Cref{fig:room12} shows where homology can help on this game. The complex built within a room encloses only
platforms and walls, decoys in the sense of \Cref{app:theory}, since
doors and ladders lie on the room boundary, so nothing behind them is
enclosed
(\Cref{app:montezuma:map}); built over the room graph the agent
learns from the passages it crosses, it sometimes encloses a room the
archive has surrounded. This happens mostly for room~12,
occasionally for room~5 (\Cref{fig:room12}a). In 7 of the 100 runs with the room-scale term, the agent visited
all four rooms around room~12 (or room~5) before entering it, giving
the term an enclosed room to act on; 33 to 45\% of all selections then landed on the ring and, in five of the six
room-12 cases, entered the room sooner than the same seed without
it, by 186k to 341k frames in four and by 853k in the fifth, whose
two runs had already diverged at room~5
(\Cref{fig:room12}b, \Cref{app:montezuma:rooms}).
Since surrounding the two rooms is rare, every arm is even with
\GE{} w.r.t.\ score at 50 seeds (\Cref{tab:rooms_wave_paired}).

The texture arm adds features homology cannot see (an adjacent door
sprite, key possession) and puts a live class's mass on them
(\Cref{sec:method}). Over 200 paired seeds
\TE{}+texture scores higher than \GE{} (\Cref{tab:montezuma}:
paired two-sided Wilcoxon $p = 0.032$, median 36{,}000 against
31{,}400, ahead on 113 pairs to 84, paired $d = 0.18$), with two
caveats: (1) the selection logs show the gain came mostly through
prioritising doors and keys (\Cref{app:montezuma:doors}), and (2) a
\GE{}+texture control, which would make the attribution exact, has
not been run.

Plain \TE{} is faster early and slightly slower by the end. The term
accounts for 21\% of selections in the first 200k frames, and at one million
frames \TE{} leads \GE{} in rooms found by a paired median of two
rooms ($p = 0.01$, 28 seeds ahead to 15), reaching the fifth room at
$0.69$M frames against $0.99$M ($p = 0.03$); by one million frames
the term's share is about one percent and by two million the lead
has faded (\Cref{fig:regime}). Over all 24 rooms of level~1 the term
costs little: on the 50 shared seeds plain \TE{} reaches the last
room in 60\% of runs against \GE{}'s 62\%, at a median 43M frames
against 42M where it does, and scores 27{,}500 against 30{,}000
($p = 0.70$, \Cref{tab:montezuma}).

\subsection{Embodied navigation: explorer trajectories as training data}
\label{sec:results:embodied}

\begin{table}[t]
\centering
\small
\caption{Explorer trajectories against oracle shortest-path
demonstrations as training data for InternVLA-N1-System2 on MultiON
(HSSD). Success at 1\,m, mean over five training seeds [95\%
$t$-interval], 325--327 unseen-scene and 327--329 held-out-category
episodes per seed and arm; the best trained arm per column in bold.
The oracle+explorer mixture and every paired comparison are in
\Cref{app:embodied}.}
\label{tab:embodied}
\begin{tabular}{lccc}
\toprule
Training data & $n$ & unseen scenes & held-out categories \\
\midrule
base model (untrained) & --- & 0.01 & 0.01 \\
oracle demonstrations & 5 & 0.11 [0.10, 0.12] & 0.12 [0.10, 0.14] \\
frontier explorer data & 5 & 0.15 [0.12, 0.17] & 0.13 [0.10, 0.15] \\
random-over-candidates explorer data & 5 & 0.13 [0.11, 0.15] & 0.12 [0.09, 0.16] \\
topological explorer data & 5 & \textbf{0.18 [0.15, 0.20]} & 0.13 [0.11, 0.15] \\
top-two-band explorer mixture & 5 & 0.16 [0.13, 0.18] & \textbf{0.16 [0.13, 0.18]} \\
\bottomrule
\end{tabular}

\end{table}

\begin{figure}[t]
\centering
\includegraphics[width=\linewidth]{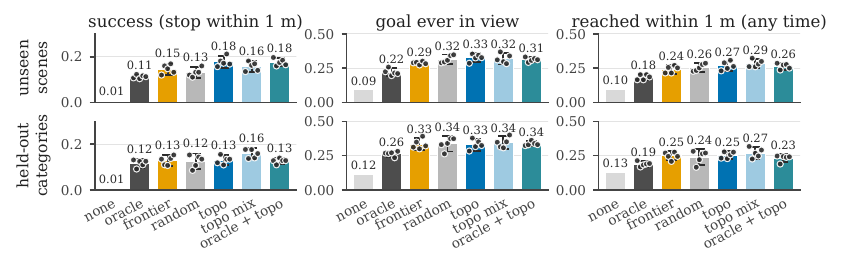}\\[1pt]
\includegraphics[width=\linewidth]{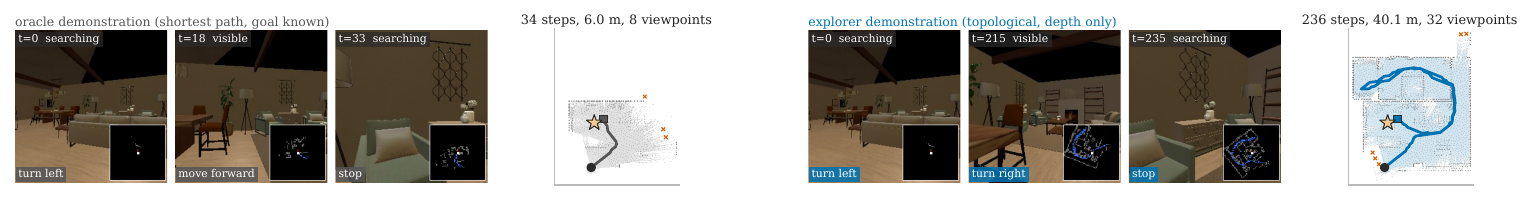}
\caption{Top: \Cref{tab:embodied} with two companion statistics
(goal ever in view, within 1\,m at any time), rows unseen scenes and
held-out categories; bars the seed mean with a 95\% $t$-interval,
dots the seeds. Explorer data, alone or mixed, generalises best to
unseen scenes; on held-out categories the explorers' data do not
separate. Bottom: one episode (``find the flower in vase'')
demonstrated by the oracle (goal known) and by the explorer (depth
only): three frames each with the egocentric map the policy sees and
the imitated action, and the top-down path each built (dots
landmarks, crosses frontier, star goal).}
\label{fig:embodied}
\end{figure}

\Cref{tab:embodied} and \Cref{fig:embodied} report the pilot.
On unseen scenes topological-explorer data
beats oracle demonstrations by 6.5 points (0.18 [0.15, 0.20] against
0.11 [0.10, 0.12], $p = 0.001$, ahead on five seeds of five), beats
the main comparator, a nearest-frontier explorer on the same
scenes and starts, by 3.2 points (0.15, $p = 0.001$, five of five),
and beats a uniform draw over its own candidate set by 4.7 points
(0.13, $p = 0.02$, five of five), so the gain comes from the mass the
homology term puts on the rim targets, not from the targets alone. It
also brings the goal into view more often (0.33 against 0.22 for
oracle data, 3.9 points over frontier data, $p = 0.02$) and stops
within a metre more often per sighting. On held-out categories the
explorer arms and the oracle overlap (0.12--0.16), and the 50/50
oracle-explorer mixture matches the explorer arm on both splits. The
bottom of \Cref{fig:embodied} shows the mechanism: an oracle
demonstration walks straight to a goal the policy cannot yet see, so
a policy imitating it never sees a search, while the explorer's
demonstration of the same episode sweeps the apartment until the goal
appears.
Absolute success is low for every trained arm (0.11--0.18 at 1\,m
with a small vision-language-action model at our compute budget,
about ten times the untrained base), so the comparison between arms
is the claim; arms share training seeds and evaluation episodes, and
every comparison is a two-sided paired $t$ over five seeds
(\Cref{app:embodied}).

\section{Conclusion}
\label{sec:discussion}

An exploration archive has a geometry of its own, and its homology
says where the archive has closed around unvisited space without
vanishing like a gradient. Used as a
selection signal, it reaches the goal first on TopoGym's 189 held-out
worlds (median $126$k steps against $186$k for \GE{}), and with the
curvature arm enters all eight adversarial chambers in $28$ of $30$
seeds where \GE{} enters them in none. Where nothing is enclosed it is
a bounded cost: on Pitfall!, and on Montezuma's Revenge unless the
complex is built over rooms, where it enters encircled rooms sooner
but leaves the score unchanged (\Cref{app:montezuma:rooms}). Beyond
restorable state spaces, its trajectories train a
vision-language-action navigation policy that succeeds $58\%$ more
often on unseen scenes than oracle demonstrations do. A natural next
step is to combine this global signal with local ones (curiosity,
visit counts, Ollivier--Ricci curvature) and to explore from the
selected cell with a learned policy rather than random actions.

\subsection*{Reproducibility statement}
Every number in the main text is regenerated by one command from the
committed summaries in the supplementary code\ificlrfinal{} (\url{https://github.com/jcarlson212/topoexplore-paper})\fi, and every figure from
the run trees the same code fetches. \Cref{app:repro} gives the
commands, the data each one reads, where the hyperparameters, proofs,
200-seed test and per-seed rows live, and the cross-architecture
determinism check that lets seeds run on different machines be combined.

\ificlrfinal
\subsubsection*{Acknowledgments}
We thank Anselm Blumer of Tufts University for reading early drafts of
this paper for writing quality.
\fi

\clearpage
\appendix
\section{Theory}
\label{app:theory}

\Cref{app:theory:model} fixes the
grid-world model and the objects the analysis is about;
\Cref{app:theory:rules} writes down the two selection rules being
compared; \Cref{app:theory:machinery} collects the lemmas;
\Cref{app:theory:t1,app:theory:t2,app:theory:t3} restate
\Cref{thm:dilution,thm:separation,thm:cert} from the main text and
prove all three; \Cref{app:theory:scope} says
what is not claimed. Everything here concerns the exploration
phase alone. \GE{} explores and then robustifies, and \TE{} changes
only cell selection, so the two methods can differ only in how they
explore.

\subsection{The model}
\label{app:theory:model}

\paragraph{Grid world.}
The world is a rectangle $G \subset \mathbb{Z}^2$, finite unless
stated otherwise, partitioned into free cells $F$ and obstacle cells
$O$; the agent occupies one free cell at a time. $F$ is $4$-connected,
and obstacle components are the $4$-connected components of $O$.
Dynamics are deterministic with
$\mathcal{A} = \{\text{up},\text{down},\text{left},\text{right}\}$:
action $a$ in cell $c$ moves to $c + a$ if that cell is free and
leaves the agent at $c$ otherwise. A \emph{door} is a free cell with
obstacle cells on two opposite sides and free cells on the other two,
so every door has width one and passing through one is ordinary
movement. We assume $O$ is well-composed, meaning it contains no
diagonal pinch: there is no pair of free cells meeting only at a
corner point with obstacles on the other diagonal. Well-composedness
is what makes the discrete Jordan curve theorem unambiguous and, at
the Rips scale fixed below, prevents an edge from leaking through a
corner.

\paragraph{Features.}
Let $W \subseteq O$ be a union of obstacle components, none of which
touches the outer boundary of $G$. For a set of cells $S$, write
$\mathrm{enc}(S)$ for the union of bounded complementary components of
$S$ in $\mathbb{Z}^2$. Let $D(W)$ be the set of doors whose adjacent
obstacle cells lie in $W$, and require $W \cup D(W)$ to be a simple
ring: $4$-connected, with $\mathrm{enc}(W \cup D(W))$ connected.
Define the enclosed free region
\begin{equation}
  I(W) \;:=\; \big(\mathrm{enc}(W \cup D(W)) \cap F\big) \setminus D(W).
  \label{eq:interior}
\end{equation}
$W$ is a \emph{chamber} if $I(W) \neq \emptyset$ and a \emph{decoy} if
$I(W) = \emptyset$. A chamber necessarily has $D(W) \neq \emptyset$:
$W$ alone enclosing free cells would make them unreachable and
contradict $4$-connectivity of $F$, so every enclosure of reachable
free space is door-mediated. A decoy is solid, or its enclosed region
is entirely obstacle; nothing is inside it and the feature is purely
topological. A \emph{pocket} is a transient object of exploration
rather than of the environment: a set of unvisited free cells enclosed
by visited cells, whose perimeter includes non-door free cells.
Decoys are born when circumnavigated and never die; chambers are born
when circumnavigated and die when their interior is filled; pockets
are born and die during ordinary exploration. The main text calls the
region a class encloses, for a chamber, decoy or pocket alike, its
enclosed region.

\paragraph{Exploration primitive.}
Both algorithms share one explorer: from the selected archived cell,
take $L \geq 1$ i.i.d.\ uniform random actions ($\pi_{\mathrm{explore}}$
of \Cref{alg:te} is uniform in every run of this paper). Write
$q_0 := 1/|\mathcal{A}| = 1/4$ for the probability that one step
equals a prescribed action. Because the actions are i.i.d.\ uniform,
for any $4$-connected path $(c_0, \ldots, c_n)$ of free cells starting
at the launch cell, with realising actions $a_1, \ldots, a_n$,
\begin{equation}
  \Pr\big[\text{step } k{+}1 \text{ is } a_{k+1} \;\big|\;
    \text{the first } k \text{ steps traversed } c_0, \ldots, c_k\big]
  \;=\; q_0
  \qquad (0 \leq k < n),
  \label{eq:q0}
\end{equation}
the conditioning being vacuous.

\paragraph{Notation.}
$A_t \subseteq F$ is the archive at iteration $t$, each cell stored
with a returning trajectory rooted at the origin $x_0$, and
$A_0 = \{x_0\}$. $N = |F|$ on a finite grid. $s_{\min}$ and $s_{\max}$
are uniform bounds on the \GE{} cell score, fixed in
\Cref{app:theory:rules}. A \emph{frontier cell} at iteration $t$ is an
archived cell with a free neighbour outside $A_t$. For a cycle
$\gamma$, a closed $4$-connected path of archived cells, $R(\gamma)$
is the bounded region of $\mathbb{Z}^2$ it encloses, well defined by
the Jordan curve theorem and unambiguous under well-composedness. The
graph distance $d_G(x, y)$ is the number of moves in a shortest
$4$-connected free path, so $d_G(x,y) \geq \|x-y\|_1 \geq \|x-y\|_2$,
with equality on the left exactly when no obstacle forces a detour.

\paragraph{The complex.}
For a finite cell set $P$ and a scale $\eps$, the Vietoris--Rips
complex $\mathrm{Cl}_{\eps}(P)$ has vertex set $P$, an edge
$\{p,q\}$ whenever the centre distance $d(p,q) \leq \eps$, and a
filled triangle on every triple that is pairwise joined. Centre
distances near a cell take the values $1$, $\sqrt{2}$ and $2$, which
makes $\eps \in [\sqrt{2}, 2)$ the valid range: at
$\eps \geq \sqrt{2}$ a $2 \times 2$ block of free cells is filled by
triangles rather than left as a spurious hole, at $\eps < 2$ no edge
crosses a wall of thickness one, and well-composedness rules out the
remaining leak through diagonal corners. We fix $\eps \in [\sqrt{2}, 2)$ throughout and
write $K_t := \mathrm{Cl}_{\eps}(A_t)$. At this scale the clique
complex of a well-composed cell set is the filled-squares complex, so
the cubical complex the lattice backend builds (\Cref{app:impl}) has
the same $H_1$ and the same enclosed regions, and the analysis applies
to either. The clique complex is here to help with the analysis, not
for the actual implementation, which computes the cubical $H_1$
directly by flood fill.

At this scale $H_1(K_t)$ is computable by flood fill: each enclosed
region containing no cell of $A_t$ is one class. Its representative
$\gamma$ is a shortest enclosing loop of cells and its rim is
\begin{equation}
  \rims(\gamma) \;:=\;
  \{\, p \in A_t : p \text{ is $4$-adjacent to a cell of }
    R(\gamma) \setminus A_t \,\}.
  \label{eq:rim}
\end{equation}
Rim and representative differ where it matters. Once a door cell is
visited but the interior is not, the tightest enclosing cycle passes
in front of the door while the true entrance is the visited door cell
itself, one step inward: on the rim but on no simple enclosing cycle.
Placing the bonus on $\gamma$ rather than $\rims(\gamma)$ would leave
such a chamber with no entrance and prune it while it is still live.
We write $\Gamma_t$ for the set of representatives of $H_1(K_t)$, one
per class. As $A_t$ grows, a class is born when its region is first
enclosed and dies when the region fills; the interval between the two
is its bar.

\paragraph{Return modes.}
Canonical \GE{} returns to the selected archived cell either by
restoring simulator state or, the dynamics being deterministic, by
replaying its stored trajectory. Under \emph{reset-return} (RR) the
environment forces a reset to $x_0$ every $T$ steps and return costs
$\ell(c, A_t)$ replay steps, leaving $T - \ell(c, A_t)$ for
exploration; every archived cell then lies within graph distance $T$
of the origin, so the effective universe is finite with
$|U_T| \leq 2T^2 + 2T + 1$. Under \emph{teleport-return} (TR) the
selected cell's state is restored directly, no replay cost is paid,
and reach is unbounded, but at most $T$ cells are archived per
episode, so deterministically
\begin{equation}
  |A_t| \;\leq\; 1 + tT .
  \label{eq:archivecap}
\end{equation}
The default elsewhere is the base model above, used by both methods:
replay return with a full $L$-step rollout and no episode structure,
which coincides with (TR) at $T = L$. Each result states the mode it
assumes.

\subsection{The two selection rules}
\label{app:theory:rules}

\paragraph{\GE{}.}
The score is \Cref{eq:cellscore}, with count attributes
$\mathrm{attrs} = \{\text{times chosen}, \text{times seen},
\text{times chosen since a new cell}\}$ and neighbour slots
$\mathrm{nbrs} = \{\text{left}, \text{right}, \text{up},
\text{down}\}$, where $\mathrm{HasNeighbor}(c, n) = 1$ iff the
position $c+n$ is archived. Selection is proportional:
$\mathrm{CellProb}(c) = \mathrm{CellScore}(c) /
\sum_{c' \in A_{t-1}} \mathrm{CellScore}(c')$. With
$n_{\max} := 2w_h + 2w_v$ the maximum total neighbour score, every
cell and every time satisfy
$1 \leq \mathrm{CellScore}(c) \leq s_{\max}$ with
\begin{equation}
  s_{\min} := 1,
  \qquad
  s_{\max} := \sum_{a}\big(w_a \epsilon_1^{-p_a} + \epsilon_2\big)
    + n_{\max} + 1 ,
  \label{eq:scorebounds}
\end{equation}
and therefore the \emph{selection floor}
\begin{equation}
  P(\text{select } c \mid A_{t-1})
  \;\geq\; \frac{s_{\min}}{s_{\max}\,|A_{t-1}|} .
  \label{eq:floor}
\end{equation}
\Cref{eq:floor} is the single fact behind every eventual-coverage
claim below.

\paragraph{Gates.}
The gate of a class $\gamma$ asks of each rim cell whether some action
from it still reaches unexplored interior:
\begin{equation}
\begin{gathered}
  g(c, \gamma) :=
  \begin{cases}
    1 & \text{if some action takes } c \text{ to an unvisited free
        cell of } R(\gamma),\\
    0 & \text{otherwise},
  \end{cases}\\[3pt]
  \gated(\gamma) := \{\, c \in \rims(\gamma) : g(c,\gamma) = 1 \,\}.
\end{gathered}
  \label{eq:gate}
\end{equation}
In \emph{oracle mode} the algorithm selects from $\gated(\gamma)$
directly. In \emph{model-free mode} it cannot evaluate $g$ and
instead keeps the attempt count $u(c,\gamma)$, the number of times $c$
has been selected for $\gamma$ whatever the outcome, weights rim cells
by the surrogate $\ghat(c,\gamma) = (1 + u(c,\gamma))^{-p_u}$, and
retires a cell at $u_{\max} \geq 1$, giving $\activeset(\gamma)$ as in
\Cref{eq:ghat}. A class is \emph{live} if $\gated(\gamma) \neq
\emptyset$ (oracle) or $\activeset(\gamma) \neq \emptyset$
(model-free), and \emph{pruned} (reported as sealed in the main
text) otherwise. Analyses of either mode may
refer to $\gated(\gamma)$, which is ground truth; only the selection
rule is mode-specific.

\paragraph{Cycle weights and the \TE{} score.}
$b(\gamma) > 0$ is the magnitude of a live class; the default is the
inradius of the unvisited part of $R(\gamma)$, equivalently the
persistence of the class under the thickening filtration of the
archive, since thickening by that radius is exactly what fills the
hole. Any positive weight preserves correctness and the choice affects
only rates. The score is \Cref{eq:score} with $\Shom$ as in
\Cref{eq:shom}. Throughout the analysis $M(c) \equiv 1$; with a
multiplier bounded in $[M_{\min}, M_{\max}]$ every bound below
survives with $s_{\min} \to M_{\min} s_{\min}$ and
$s_{\max} \to M_{\max} s_{\max}$. We take $p_{\mathrm{topo}} = 1$
throughout, the value used in every experiment (\Cref{tab:hparams});
for general $p_{\mathrm{topo}}$ replace $b(\gamma)$ by
$b(\gamma)^{p_{\mathrm{topo}}}$ below. The index set of the sum in
\Cref{eq:shom} depends on the mode: in oracle mode it runs over
$c \in \gated(\gamma)$ with $\psi(c,\gamma) = 1/|\gated(\gamma)|$,
in model-free mode over $c \in \activeset(\gamma)$ with
$\psi(c,\gamma) = \ghat(c,\gamma) /
\sum_{c' \in \activeset(\gamma)} \ghat(c',\gamma)$,
so in either mode each live class contributes total mass
$\wtopo\, b(\gamma)$ and $\Shom \equiv 0$ off live classes. Write
\begin{equation}
  \Blive := \sum_{\gamma \text{ live}} b(\gamma),
  \qquad
  \Sigma^{\GEsub}_t := \sum_{c \in A_t} \mathrm{CellScore}(c),
  \qquad
  \delta_t := \frac{\wtopo \Blive}{\Sigma^{\GEsub}_t + \wtopo \Blive} .
  \label{eq:delta}
\end{equation}
$\delta_t$ is the \emph{realised topological share}. The proportional
draw over $\mathrm{Score}_{\TEsub}$ can be read as a draw over $2|A_t|$
entries, each cell contributing a \GE{} entry of mass
$\mathrm{CellScore}(c)$ and a topological entry of mass $\Shom(c)$;
$\delta_t$ is exactly the probability that the draw lands on a
topological entry, and the mass of a set of cells is the sum of their
masses.

The local term $\Sloc$ carries no part of the theory. The results
below do not require $\Sloc \equiv 0$, only a uniform per-cell bound
$\Sloc(c) \leq s_{\mathrm{loc}} < \infty$: the total mass it
contributes is then at most $s_{\mathrm{loc}}|A_t|$, the same order as
$\Sigma^{\GEsub}_t \in [\,|A_t|, s_{\max}|A_t|\,]$, so every appearance
of $s_{\max}$ below absorbs it through
$s_{\max} \mapsto s_{\max} + s_{\mathrm{loc}}$. Floors and ceilings
survive with changed constants.

One structural remark on \Cref{eq:shom}. Even with shortest
representatives the sum can have more than one term in two dimensions.
In a theta configuration, an outer visited ring with a visited strip
through the middle enclosing two unvisited pockets, the two classes'
representatives are the left and right loops, the strip cells lie on
both, and a strip cell with an unvisited neighbour in each pocket is
gated for both: it is a doorway to two distinct unexplored regions and
receives both contributions. The bounds below depend only on the total
live mass $\wtopo \Blive$ and are insensitive to the choice of
representatives.

\subsection{Machinery}
\label{app:theory:machinery}

\begin{lemma}[Wall-blindness, \GE{} and \TE{}]
\label{lem:wallblind}
$\mathrm{HasNeighbor}(c,n)$ tests archive membership of the position
$c+n$, so a neighbour absent because it is a wall and one absent
because it is unexplored free space are indistinguishable.
Consequently \emph{(i)} every wall-adjacent cell retains a neighbour
score forever, since wall positions are never archived, and along the
rim of a pocket or chamber that score is carried by all
$|\rims(\gamma)|$ cells alike; and \emph{(ii)} the neighbour score
raises any single cell's score by at most $n_{\max}$, a factor
$(1 + n_{\max})$ over the floor $s_{\min} = 1$, on a lattice without
levels or keys ($M \equiv 1$), as in TopoGym.
\end{lemma}

\begin{proof}
Both claims use the definition of $\mathrm{HasNeighbor}$ in
\Cref{eq:cellscore}. For (i), a wall position is not a free cell and
so is never added to the archive, hence
$\mathrm{HasNeighbor}(c,n) = 0$ for that slot at every $t$, and the
test returns the same value for an unexplored free neighbour. For
(ii), with $M \equiv 1$ and only lattice neighbours, the neighbour
sum is at most $\sum_{n} w_n = 2w_h + 2w_v = n_{\max}$, and
$\mathrm{CellScore}(c) \geq 1$ by the $+1$ in \Cref{eq:cellscore}.
\end{proof}

The implication of \Cref{lem:wallblind} is that nothing in \GE{}'s score
singles out a door front. A cell in front of a door and a cell against
a blank wall carry the same neighbour score for the same reason.

\begin{lemma}[Floor and share]
\label{lem:floorshare}
$\mathrm{Score}_{\TEsub}(c) \geq 1$ always, and conditional on any
history the \TE{} kernel satisfies
\begin{equation}
  P(\text{select } c) \;\geq\;
    \frac{s_{\min}}{s^{\TEsub}_{\max}(t)\,|A_t|},
  \qquad
  P(\text{select } c) \;\geq\;
    (1 - \delta_t)\,\frac{\mathrm{CellScore}(c)}{\Sigma^{\GEsub}_t},
  \label{eq:tefloor}
\end{equation}
where $s^{\TEsub}_{\max}(t) := s_{\max} + \max_c \Shom(c)
\leq s_{\max} + \wtopo \Blive$.
\end{lemma}

\begin{proof}
$\mathrm{Score}_{\TEsub}(c) = \mathrm{CellScore}(c) + \Shom(c) + \Sloc(c)
\geq \mathrm{CellScore}(c) \geq 1$ since both added terms are
non-negative. For the first inequality of \Cref{eq:tefloor}, the
numerator is at least $s_{\min} = 1$ and each of the $|A_t|$
denominator terms is at most $s^{\TEsub}_{\max}(t)$. For the second,
split the normaliser as $\Sigma^{\GEsub}_t + \wtopo \Blive$, so that the
\GE{} entry of cell $c$ alone gives
\begin{align*}
  P(\text{select } c)
  \;\geq\; \frac{\mathrm{CellScore}(c)}{\Sigma^{\GEsub}_t + \wtopo \Blive}
  &\;=\; \frac{\mathrm{CellScore}(c)}{\Sigma^{\GEsub}_t}
    \cdot \frac{\Sigma^{\GEsub}_t}{\Sigma^{\GEsub}_t + \wtopo \Blive}\\
  &\;=\; (1 - \delta_t)\,
    \frac{\mathrm{CellScore}(c)}{\Sigma^{\GEsub}_t},
\end{align*}
using the definition of $\delta_t$ in \Cref{eq:delta}.
\end{proof}

The second inequality says \TE{} behaves like \GE{} damped by
$(1 - \delta_t)$, and exactly like \GE{} conditional on the draw
landing on non-topological mass.

\begin{lemma}[Immediate decoy pruning, oracle mode]
\label{lem:oracleprune}
In oracle mode a decoy's class is pruned as soon as it is detected:
every rim cell has all interior-side neighbours obstructed and, once
the surrounding corridor is visited, no unvisited free neighbour.
Decoys therefore absorb zero bonus mass.
\end{lemma}

\begin{proof}
Detection of a decoy $W$ means the archive encloses $W$, so every cell
of $\rims(\gamma)$ is archived and every cell of $R(\gamma)$ is an
obstacle by $I(W) = \emptyset$ in \Cref{eq:interior}. No action from a
rim cell reaches an unvisited free cell of $R(\gamma)$, so
$g(c,\gamma) = 0$ for every $c$ by \Cref{eq:gate},
$\gated(\gamma) = \emptyset$, the class is pruned, and $\Shom$ is zero
on it by the convention that $\Shom$ vanishes off live classes.
\end{proof}

\begin{lemma}[Crossover]
\label{lem:crossover}
Since $\Sigma^{\GEsub}_t \in [\,|A_t|,\ s_{\max}|A_t|\,]$,
\begin{equation}
  \delta_t \;\geq\;
  \frac{\wtopo \Blive}{s_{\max}|A_t| + \wtopo \Blive},
  \label{eq:crossover}
\end{equation}
so $\delta_t = \Theta(1)$ while
$s_{\max}|A_t| \lesssim \wtopo \Blive$ and
$\delta_t = \Theta\big(\wtopo \Blive / (s_{\max}|A_t|)\big)$ beyond.
\end{lemma}

\begin{proof}
$\Sigma^{\GEsub}_t$ sums $|A_t|$ terms each in $[1, s_{\max}]$ by
\Cref{eq:scorebounds}, giving the stated range; substituting the upper
end into \Cref{eq:delta} gives \Cref{eq:crossover}, which is
decreasing in $s_{\max}|A_t|$ and takes the two stated regimes
according to which term dominates the denominator.
\end{proof}

\Cref{lem:crossover} is the dilution phenomenon on the \TE{} side.
$\mathrm{CellScore}$ sums per-attribute terms over the whole archive,
so its normaliser grows like $|A_t|$; against a fixed $\wtopo$ the
topological share goes to zero and \TE{} degenerates to \GE{}. Past
the crossover \TE{}'s advantage at a detected entrance is the
multiplicative factor $\Theta(\wtopo b_i / s_{\max})$ rather than
unbounded, unless $\wtopo = \Omega(|A_t|)$. This is why the schedule
$\wtopo(t) = \alpha|A_t|$, and not a constant, is the default.

\begin{lemma}[Eventual full exploration]
\label{lem:cover}
In the finite grid of \Cref{app:theory:model}, both \GE{} and \TE{}
visit every cell of $F$ in almost surely finite time, and
\begin{equation}
  \mathbb{E}[\tau_{\mathrm{cover}}(\GEsub)]
    \leq \frac{s_{\max}}{s_{\min}} \cdot \frac{N^2}{q_0},
  \qquad
  \mathbb{E}[\tau_{\mathrm{cover}}(\TEsub)]
    \leq \frac{s^{\TEsub}_{\max}}{s_{\min}} \cdot \frac{N^2}{q_0},
  \label{eq:cover}
\end{equation}
with $s^{\TEsub}_{\max} = s_{\max} + \wtopo \sup_t \Blive$.
\end{lemma}

\begin{proof}
At any iteration with $A_t \neq F$, connectivity of $F$ supplies a
frontier cell $c^{*} \in A_t$. By \Cref{eq:floor} it is selected with
probability at least $s_{\min}/(s_{\max}N)$, and by
\Cref{eq:tefloor} the same holds for \TE{} with
$s_{\max} \mapsto s^{\TEsub}_{\max}$. Conditional on selecting $c^{*}$,
the first exploration step moves into an unvisited neighbour with
probability at least $q_0$ by \Cref{eq:q0}. Hence every iteration with
an incomplete archive discovers at least one new cell with probability
at least
\begin{equation}
  p_0 \;:=\; \frac{q_0\, s_{\min}}{s_{\max} N},
  \label{eq:p0}
\end{equation}
and the same with $s^{\TEsub}_{\max}$ for \TE{}.

Let $\mathcal{F}_t$ be the $\sigma$-field generated by the first $t$
iterations. Call iteration $t$ a discovery if at least one new cell is
archived during it, let $T_0 := 0 < T_1 < \cdots < T_K$ be the
discovery iterations, and let $D_j := T_j - T_{j-1}$. Every discovery
adds at least one cell and $|A_0| = 1$, so coverage completes at the
$K$-th discovery with $K \leq N - 1$.

Fix $j$ and condition on $\mathcal{F}_{T_{j-1}}$ restricted to
$\{j \leq K\}$, on which the archive is incomplete throughout epoch
$j$ and \Cref{eq:p0} applies at every step of it. We claim
\begin{equation}
  P\big(D_j > m \mid \mathcal{F}_{T_{j-1}}\big)
  \;\leq\; (1 - p_0)^m
  \qquad (m \geq 0),
  \label{eq:geomdom}
\end{equation}
by induction on $m$. The case $m = 0$ is trivial. For $m \geq 1$,
$\{D_j > m\}$ is $\{D_j > m-1\}$ intersected with the event that
iteration $T_{j-1}+m$ is not a discovery, and
$\{D_j > m-1\} \in \mathcal{F}_{T_{j-1}+m-1}$, so by the tower
property
\begin{align*}
  P\big(D_j > m \mid \mathcal{F}_{T_{j-1}}\big)
  &= \mathbb{E}\Big[\mathbf{1}\{D_j > m-1\}\,
     P\big(\text{no discovery at } T_{j-1}{+}m \mid
       \mathcal{F}_{T_{j-1}+m-1}\big) \;\Big|\;
     \mathcal{F}_{T_{j-1}}\Big]\\
  &\leq (1 - p_0)\, P\big(D_j > m-1 \mid \mathcal{F}_{T_{j-1}}\big),
\end{align*}
the inequality using \Cref{eq:p0}, which is uniform over histories.
That uniformity is why no independence between epochs is needed.
Summing \Cref{eq:geomdom} gives
$\mathbb{E}[D_j \mid \mathcal{F}_{T_{j-1}}] = \sum_{m \geq 0}
P(D_j > m \mid \mathcal{F}_{T_{j-1}}) \leq 1/p_0$ on $\{j \leq K\}$.

Now $\tau_{\mathrm{cover}} = \sum_{j=1}^{K} D_j
= \sum_{j=1}^{N-1} D_j \mathbf{1}\{j \leq K\}$ and
$\{j \leq K\} = \{A_{T_{j-1}} \neq F\} \in \mathcal{F}_{T_{j-1}}$, so
\[
  \mathbb{E}[\tau_{\mathrm{cover}}]
  = \sum_{j=1}^{N-1} \mathbb{E}\Big[\mathbf{1}\{j \leq K\}\,
      \mathbb{E}\big[D_j \mid \mathcal{F}_{T_{j-1}}\big]\Big]
  \;\leq\; \sum_{j=1}^{N-1} \frac{P(j \leq K)}{p_0}
  \;\leq\; \frac{N}{p_0}
  \;=\; \frac{s_{\max}}{s_{\min}} \cdot \frac{N^2}{q_0},
\]
which is \Cref{eq:cover}, and identically for \TE{}. Finiteness of the
expectation forces $\tau_{\mathrm{cover}} < \infty$ almost surely.
\end{proof}

The guarantee in \Cref{lem:cover} comes from the $+1$ floor and
connectivity alone. Every form of shaping, whether counts, neighbours
or topology, moves rates and never the guarantee. The $N^2$ comes from
crediting a single frontier cell per iteration at the floor rate
$\sim 1/N$ for each of the $\sim N$ discoveries needed; in open
regions many cells are frontier, so the bound is loose.

\begin{lemma}[Bounded per-class cost, model-free]
\label{lem:decoytax}
Each class $\gamma$ absorbs at most $|\rims(\gamma)|\, u_{\max}$ bonus
selections before it is pruned or its region is entered; for a decoy,
pruning is certain.
\end{lemma}

\begin{proof}
$u(c,\gamma)$ increments on every selection of $c$ for $\gamma$
regardless of outcome. After $u_{\max}$ selections of a given cell
that cell leaves $\activeset(\gamma)$ by \Cref{eq:ghat}. There are
$|\rims(\gamma)|$ such cells, so unless the region is entered first,
after at most $|\rims(\gamma)|\,u_{\max}$ selections
$\activeset(\gamma)$ is empty and the class is pruned, after which
$\Shom$ vanishes on it. For a decoy every rim cell fails
deterministically, since its interior-side neighbours are obstacles
and its exterior-side neighbours are archived, so entry never occurs
and pruning is certain.
\end{proof}

\Cref{lem:decoytax} is the model-free analogue of
\Cref{lem:oracleprune} and is what makes the term safe to leave on:
the budget sunk into sealed structure is finite and independent of run
length. The \GE{} component of the score still selects such cells at
its usual floor rate; what is bounded is the bonus mass a decoy can
absorb.

\begin{lemma}[Almost-sure completeness under (TR)]
\label{lem:trcomplete}
Under teleport-return, for any fixed cell $x$, in particular any
chamber-interior cell, both \GE{} and \TE{} archive $x$ in almost
surely finite time; consequently $|A_t| \to \infty$ almost surely.
\end{lemma}

\begin{proof}
Fix a free path $x_0 = y_0, y_1, \ldots, y_k = x$ and let $y_i$ be the
last archived cell on it at the current time. Selecting $y_i$ has
probability at least $s_{\min}/(s^{\TEsub}_{\max}|A_t|)$ by
\Cref{eq:tefloor}, and stepping to $y_{i+1}$ has probability at least
$q_0$ by \Cref{eq:q0}. With \Cref{eq:archivecap} the conditional
probability of progress at episode $t$ is therefore at least
\[
  \frac{q_0\, s_{\min}}{s^{\TEsub}_{\max}} \cdot \frac{1}{1 + tT},
  \qquad
  \sum_{t \geq 1}
  \frac{q_0\, s_{\min}}{s^{\TEsub}_{\max}\,(1 + tT)} = \infty .
\]
By the second Borel--Cantelli lemma in L\'evy's conditional extension
\citep{Williams1991}, which states that
$\sum_t P(E_t \mid \mathcal{F}_{t-1}) = \infty$ forces infinitely many
$E_t$ to occur almost surely, progress events occur infinitely often,
and $k$ of them reach $x$.
\end{proof}

\subsection{Theorem 1: dilution versus anytime entry}
\label{app:theory:t1}

Chamber $i$ has enclosing class $\gamma_i$, magnitude $b_i$, and
door-entry probability $q_i \geq q_0$ from a gated cell.

\begin{restatethm}{1}{dilution versus anytime entry}
Conditional on any history $\mathcal{F}_t$:
\emph{(i) Ceiling, either return mode, and \TE{} before detection.}
If chamber $i$ is undiscovered and
$M := \max_{c \in A_t} d_G(c, x_0)$, then
\begin{equation}
  P(\text{enter chamber } i \text{ next episode} \mid \mathcal{F}_t)
  \;\leq\; \frac{s_{\max}\, n_D\, h_T}{s_{\min}\,|A_t|}
  \;\leq\; \frac{C(T)}{M},
  \label{eq:t1ceiling}
\end{equation}
where $n_D$ is the number of cells from which the door is reachable
within the exploration budget, $h_T \leq 1$ bounds the probability
that a rollout from such a cell reaches the door before the budget
runs out, and $C(T) := s_{\max} n_D h_T / s_{\min}$. Under
space-filling exploration, $|A_t| \geq \kappa M^2$ for a covered
fraction $\kappa \in (0, 1]$, the bound improves to $C(T)/(\kappa M^2)$.
\emph{(ii) Floor, either return mode.} If $\gamma_i$ is detected and
its interior unvisited, then
\begin{equation}
\begin{aligned}
  P(\text{enter chamber } i \text{ next episode} \mid \mathcal{F}_t)
  &\;\geq\; \frac{\wtopo\, b_i}{s_{\max}|A_t| + \wtopo \Blive}
    \cdot \frac{q_i}{|\gated(\gamma_i)|}\\
  &\;=\; \delta_t \cdot \frac{b_i}{\Blive}
    \cdot \frac{q_i}{|\gated(\gamma_i)|},
\end{aligned}
  \label{eq:t1floor}
\end{equation}
which is constant in $t$ under $\wtopo(t) = \Omega(|A_t|)$. In
model-free mode the gated factor is reached by exhaustion rather than
by an oracle: each non-gated rim cell leaves $\activeset(\gamma_i)$
after $u_{\max}$ draws, so
$\activeset(\gamma_i) \subseteq \gated(\gamma_i)$ after at most
$|\rims(\gamma_i)|\,u_{\max}$ rim draws.
\end{restatethm}

\begin{proof}
\emph{(i)} Entering chamber $i$ during one episode requires selecting
one of the $n_D$ cells from which the door is reachable within the
budget, and the rollout then reaching the door, which has probability
at most $h_T$. A union bound over those cells gives
\[
\begin{gathered}
  P(\text{enter chamber } i \mid \mathcal{F}_t)
  \;\leq\; n_D\, h_T \cdot \max_{c \in A_t} P(\text{select } c),\\[2pt]
  P(\text{select } c)
  = \frac{\mathrm{CellScore}(c)}{\Sigma^{\GEsub}_t}
  \;\leq\; \frac{s_{\max}}{s_{\min}\,|A_t|},
\end{gathered}
\]
using $\mathrm{CellScore}(c) \leq s_{\max}$ in the numerator and
$\mathrm{CellScore}(c') \geq s_{\min} = 1$ for each of the $|A_t|$
denominator terms, both from \Cref{eq:scorebounds}. Chaining the two
gives the first inequality of \Cref{eq:t1ceiling}. For the second,
$|A_t| \geq M$: the stored trajectory of the farthest archived cell
$c$ visits only archived cells and, being a walk from $x_0$ to a cell
with $d_G(c, x_0) = M$, visits at least $M + 1$ distinct ones.
Substituting and writing $C(T) := s_{\max} n_D h_T / s_{\min}$ gives
$C(T)/M$, and substituting $|A_t| \geq \kappa M^2$ instead gives
$C(T)/(\kappa M^2)$.

For \TE{} before detection, no live class encloses the chamber, so
$\Shom = 0$ on its door front while the normaliser
$\Sigma^{\GEsub}_t + \wtopo \Blive$ only grows, and the same ceiling
holds.

\emph{(ii)} The selection rule factors the entry event into three
conditional probabilities. By \Cref{eq:delta} and
$\Sigma^{\GEsub}_t \leq s_{\max}|A_t|$,
\begin{align*}
  P(\text{draw lands on } \gamma_i\text{'s mass} \mid \mathcal{F}_t)
  &= \frac{\wtopo\, b_i}{\Sigma^{\GEsub}_t + \wtopo \Blive}
   \;\geq\; \frac{\wtopo\, b_i}{s_{\max}|A_t| + \wtopo \Blive},\\
  P(\text{select } c \mid \text{draw on } \gamma_i)
  &= \psi(c, \gamma_i) = \frac{1}{|\gated(\gamma_i)|}
   \quad \text{(oracle mode)},\\
  P(\text{enter chamber } i \mid \text{select } c) &\geq q_i ,
\end{align*}
and multiplying the three gives \Cref{eq:t1floor}. For the bound to
stay away from zero the factor $b_i/\Blive$ must stay bounded below,
which it does: by \Cref{lem:oracleprune} a decoy contributes zero to
$\Blive$, and a pocket contributes $b(\gamma)$ only until it is
entered, so $\Blive$ is bounded by the total magnitude of the genuine
unentered chambers plus transient pocket terms.

In model-free mode the oracle share is replaced by
$\psi(c,\gamma_i) = \ghat(c,\gamma_i) / \sum_{c' \in
\activeset(\gamma_i)} \ghat(c',\gamma_i)$. Before any attempts
$\ghat \equiv 1$, so $\psi(c,\gamma_i) = 1/|\rims(\gamma_i)|$, and
cells with equal counts keep equal weights, so the allocation over
$\activeset(\gamma_i)$ stays near-uniform as counters climb. Every
cell, gated or not, leaves $\activeset(\gamma_i)$ after exactly
$u_{\max}$ selections, so the rim absorbs at most
$|\rims(\gamma_i)|\,u_{\max}$ draws in total by
\Cref{lem:decoytax}; the non-gated cells exhaust after at most
$|\rims(\gamma_i) \setminus \gated(\gamma_i)|\,u_{\max}$ of them, and
from that point $\activeset(\gamma_i) \subseteq \gated(\gamma_i)$, so
the gated share of the normalised weights is exactly $1$. Each gated
selection enters with probability at least $q_i$ and the door front
receives up to $|\gated(\gamma_i)|\,u_{\max}$ selections before it too
exhausts, so the class is pruned before first entry with probability
at most
\begin{equation}
  (1 - q_i)^{u_{\max}} \;\leq\; e^{-q_i u_{\max}} .
  \label{eq:sweepfail}
\end{equation}
We call this the model-free exhaustion sweep and reuse it in
\Cref{app:theory:t2}.

For the behaviour in $t$: \Cref{eq:crossover} is $\Theta(1)$ while
$s_{\max}|A_t| \lesssim \wtopo \Blive$ and decays like
$\wtopo \Blive / (s_{\max}|A_t|)$ beyond, and \Cref{lem:trcomplete}
gives $|A_t| \to \infty$ almost surely. So for fixed $\wtopo$ the
condition eventually fails and \Cref{eq:t1floor} becomes
$\Theta\big(\wtopo b_i q_i / (s_{\max}|A_t|\,|\gated(\gamma_i)|)\big)$,
while under $\wtopo(t) = \Omega(|A_t|)$ it holds for all $t$,
$\delta_t = \Theta(1)$ uniformly, and \Cref{eq:t1floor} is a
feature-dependent constant.
\end{proof}

For \GE{}, and for \TE{} before detection, no selection mass sits on
the hole: by \Cref{lem:wallblind} the door front is just another
archived cell among $|A_t|$, so the more the archive grows the less
likely any single episode is to enter the chamber. For \TE{} after
detection the class carries mass $\wtopo b(\gamma_i)$ concentrated on
its $O(1)$ gated cells no matter how large the archive has grown. The
only thing that erodes that is the crossover of
\Cref{lem:crossover}, which the schedule removes.

\subsection{Theorem 2: all-but-\texorpdfstring{$k$}{k} holes}
\label{app:theory:t2}

\paragraph{Setup.}
Finite grid, teleport-return with episode horizon $T = L$. There are
$k$ chambers with pairwise-disjoint walls $W_1, \ldots, W_k$ and doors
$d_1, \ldots, d_k$, and the doors are at pairwise graph distance
greater than $L$, so a single rollout enters at most one
previously-unentered chamber. Condition both algorithms on the same
late-exploration event $E_m$: every free cell outside
$\bigcup_i I(W_i)$ is archived, $m := |A_0| = N - \sum_i |I(W_i)|$,
every $\gamma_i$ has been circumnavigated and is therefore detected
and live, and every interior is unvisited. No nesting is assumed.

\begin{restatethm}{2}{fixed-budget separation}
Model-free \TE{} with a schedule maintaining
$\delta_t \geq \delta_{\min} > 0$, and a gate budget
$u_{\max} \geq q_i^{-1}\log(s^{\TEsub}_{\max}N/s_{\min})$ for every $i$.
Conditional on $E_m$, with the per-chamber budget
\begin{equation}
  \beta_i := |\rims(\gamma_i)|\,u_{\max} + 2q_i^{-1}
    + |I(W_i)|\,(1 + u_{\max}),
  \label{eq:beta}
\end{equation}
\emph{(i)} $\mathbb{E}[\tau_{\mathrm{all}}(\TEsub)] \leq
\delta_{\min}^{-1} \sum_{i=1}^{k} \beta_i$, linear in the total
remaining rim size and interior volume and independent of $m$;
\emph{(ii)} for \GE{}, every iteration and every unentered chamber
satisfy, conditional on the history, $P(\text{enter } i) \leq \lambda_0 :=
s_{\max}\bar{n}/(s_{\min}m)$ with $\bar{n} := \max_i n_i$ and $n_i$
the number of cells from which door $d_i$ is reachable within one
rollout, so the
number of distinct chambers entered within $N$ iterations is
stochastically dominated by $\mathrm{Binomial}(N, k\lambda_0)$ and
$P(\GEsub \text{ enters all } k \text{ within } N) \leq
(eN\lambda_0)^{k}$;
\emph{(iii)} at $N^{*} := \lceil (2/\delta_{\min})\sum_i \beta_i
\rceil$, \TE{} enters all $k$ with probability at least $1/2$ while
\GE{}'s probability is at most $(eN^{*}\lambda_0)^{k}$, exponentially
small in $k$ whenever $m \gg \sum_i \beta_i$.
\end{restatethm}

\begin{proof}
\emph{(i)} Under $E_m$ every $\gamma_i$ is detected and live at
$t = 0$, so $\Blive = \sum_{i=1}^{k} b_i$ and each draw lands on
topological mass with probability $\delta_t \geq \delta_{\min}$ by the
schedule assumption. A chamber's bar dies when its interior is filled
and not at first entry, so an entered chamber retains positive live
mass until its interior is archived. Count the topological draws until
all $k$ chambers are entered and classify each into exactly one of
three types: $X$ draws select a non-gated rim cell of a
still-unentered class, $Y$ draws select a gated cell of a
still-unentered class, and $Z$ draws land on the live structure of an
already-entered chamber, meaning its rim or the classes its
partially-filled interior forms.

For $X$: every non-door rim cell fails deterministically, since its
wall-side neighbours are obstacles and its corridor-side neighbours
are archived under $E_m$, so each such draw increments
$u(c,\gamma_i)$ and after $u_{\max}$ failures that cell leaves
$\activeset(\gamma_i)$ by \Cref{lem:decoytax}. Class $i$ has
$|\rims(\gamma_i)|$ rim cells, so
\begin{equation}
  \mathbb{E}[X] \;\leq\;
  \sum_{i=1}^{k} |\rims(\gamma_i)|\, u_{\max} .
  \label{eq:EX}
\end{equation}

For $Y$: a gated draw on class $i$ enters chamber $i$ with probability
at least $q_i$, and $Y$ counts draws on class $i$ only until first
entry, so Wald's identity applied per class gives
\begin{equation}
  \mathbb{E}[Y] \;\leq\; \sum_{i=1}^{k} \frac{1}{q_i} .
  \label{eq:EY}
\end{equation}
This accounting presumes class $i$ stays live until first entry.
Gated cells also accumulate failures, so the door front can exhaust
$u_{\max}$ failures before a success and prune the class; by
\Cref{eq:sweepfail} and the $u_{\max}$ hypothesis this has probability
at most $e^{-q_i u_{\max}} \leq s_{\min}/(s^{\TEsub}_{\max}N)$. On that
event entry still occurs through the selection floor of
\Cref{eq:tefloor}: a door-front cell is selected with probability at
least $s_{\min}/(s^{\TEsub}_{\max}N)$ per iteration and enters with
probability at least $q_i$, so in expected time at most
$s^{\TEsub}_{\max}N/(s_{\min}q_i)$. The product of the two is at most
$q_i^{-1}$ per chamber, the second $q_i^{-1}$ in \Cref{eq:beta}.

For $Z$: under $E_m$ every free cell outside the interiors is
archived, so a type-$Z$ draw on entered chamber $i$ either archives a
new interior cell, possible at most $|I(W_i)|$ times, or fails on a
cell of an interior pocket rim and increments that cell's counter;
those rim cells are interior cells, so at most $|I(W_i)|$ of them
absorb at most $u_{\max}$ failures each, giving
\begin{equation}
  \mathbb{E}[Z] \;\leq\;
  \sum_{i=1}^{k} |I(W_i)|\,(1 + u_{\max}) .
  \label{eq:EZ}
\end{equation}

Each iteration is a topological draw with probability at least
$\delta_{\min}$ conditional on the history, so by Wald's identity
again, and combining \Cref{eq:EX,eq:EY,eq:EZ} with the pruning term,
\[
  \mathbb{E}[\tau_{\mathrm{all}}(\TEsub)]
  \;\leq\; \frac{\mathbb{E}[X] + \mathbb{E}[Y] + \mathbb{E}[Z]}
    {\delta_{\min}}
    + \sum_{i=1}^{k} e^{-q_i u_{\max}}
      \frac{s^{\TEsub}_{\max}N}{s_{\min}q_i}
  \;\leq\; \frac{1}{\delta_{\min}} \sum_{i=1}^{k} \beta_i ,
\]
and no $m$ dependence survives. The gate-budget hypothesis is a
convenience that makes $\beta_i$ clean. The runs use $u_{\max} = 8$,
below it; the middle expression then still bounds
$\mathbb{E}[\tau_{\mathrm{all}}(\TEsub)]$, with the pruning sum in
place of the absorption it bounds, and it is that sum, not $m$, that
the smaller gate budget enlarges.

\emph{(ii)} This is \Cref{eq:t1ceiling} conditional on $E_m$, with two
additions. The archive only grows, so $|A_t| \geq m$ for all $t$ and
the ceiling is uniform in $t$:
\[
  P(\text{enter chamber } i \text{ this iteration})
  \;\leq\; \frac{s_{\max}\, n_i\, h_T}{s_{\min}\,|A_t|}
  \;\leq\; \frac{s_{\max}\,\bar{n}}{s_{\min}\, m}
  \;=\; \lambda_0 ,
\]
using $h_T \leq 1$ and $n_i \leq \bar{n}$. And by
\Cref{lem:wallblind}, under $E_m$ chamber $i$'s door front carries the
same neighbour score as every other wall-adjacent cell, so nothing in
the score singles it out. A union bound over the at most $k$ unentered
chambers gives
$P(\text{enter some new chamber this iteration}) \leq k\lambda_0$, and
the separation assumption caps entries at one per iteration, which
gives the binomial domination. For the tail bound, if
$\mathrm{Bin}(N,p) \geq k$ then some specific set of $k$ trials all
succeeded, so a union bound over the $\binom{N}{k}$ such sets gives
\[
  P(\mathrm{Bin}(N,p) \geq k)
  \;\leq\; \binom{N}{k} p^k
  \;\leq\; \frac{N^k}{k!} p^k
  \;\leq\; \Big(\frac{eNp}{k}\Big)^{k},
\]
where the last step uses $k! \geq (k/e)^k$, itself from
$e^k = \sum_{j \geq 0} k^j/j! \geq k^k/k!$. Substituting
$p = k\lambda_0$ cancels the $k$ and leaves $(eN\lambda_0)^k$.

\emph{(iii)} For \TE{}, part (i) gives
$\mathbb{E}[\tau_{\mathrm{all}}(\TEsub)] \leq
\delta_{\min}^{-1}\sum_i \beta_i \leq N^{*}/2$, so Markov's inequality
gives $P(\tau_{\mathrm{all}}(\TEsub) > N^{*}) \leq
\mathbb{E}[\tau_{\mathrm{all}}(\TEsub)]/N^{*} \leq 1/2$. For \GE{},
substitute $N^{*}$ into the tail bound of (ii).
\end{proof}

\TE{} wastes at most $|\rims(\gamma_i)|\,u_{\max}$ failed tries per
class, because a constant share $\delta_{\min}$ of all draws lands on
the rim and the attempt counters retire each rim cell after $u_{\max}$
draws, concentrating that share onto the door front while it remains
active. Entering a chamber begins to reduce its bar, and the remaining
interior absorbs at most $|I(W_i)|(1 + u_{\max})$ further draws before
the bar dies and its share passes to the classes that are left. \GE{}
has no concentration mechanism: by \Cref{eq:t1ceiling} each remaining
door front holds a $\sim 1/m$ share of attention forever, so at a
budget sized to \TE{}'s sweeps \GE{} must win $k$ independent
$1/m$-lotteries. Nesting is unnecessary; dilution alone gives the
result.

This is the statement \Cref{fig:ksweep} tests. The chamber-count
families were built to instantiate its hypotheses: $k$ chambers whose
doors sit beyond one episode horizon in a world the agent has
otherwise explored, so that $m$ grows while $\sum_i \beta_i$ does not.
The predicted shape is a step in $k$ for \GE{} and a flat curve for an
arm that keeps $\delta_t$ from decaying, which is what the left and
middle panels of \Cref{fig:ksweep} show. The right panel is the
control: the open-field family removes the wall a curvature signal can
follow, and the ordering inverts.

\subsection{Theorem 3: certification}
\label{app:theory:t3}

\begin{restatethm}{3}{certification}
Finite grid, model-free \TE{}. Almost surely there is a finite
$\tau_{\mathrm{cert}}$ at which (i) every chamber and pocket bar has
died because its interior is visited, (ii) every decoy class is pruned
by gate exhaustion, and (iii) $\Gamma_t$ consists of exactly the $k_d$
decoy classes. $\tau_{\mathrm{cert}}$ is observable: stop when no
frontier cell and no live class remains. At $\tau_{\mathrm{cert}}$ the
algorithm outputs $H_1 = \mathbb{F}_2^{k_d}$ with a certificate,
namely the exhausted gate log for each reported class and the explicit
interior filling for each killed one.
\end{restatethm}

\begin{proof}
\Cref{lem:cover} gives full coverage almost surely.

\emph{(i)} The archive only grows, so the complexes form a filtration
$K_{t_1} \subseteq K_{t_2} \subseteq \cdots$ and persistent $H_1$
assigns each class a bar $[t_b, t_d)$, with $t_b$ the first time the
class exists and $t_d$ the first time it maps to zero under the
inclusions, $t_d = \infty$ if never \citep{Carlsson2009}. For a
pocket, at its birth time $t_b$ the archived cells first contain a
closed chain of Rips edges
\[
  z = \{c_1c_2\} + \{c_2c_3\} + \cdots + \{c_rc_1\},
  \qquad [z] \neq 0 \text{ in } H_1(K_{t_b}),
\]
winding once around the enclosed unarchived cells. The class is
non-zero because no $2$-chain of $K_{t_b}$ has boundary $z$: at
$\eps < 2$ no edge crosses a wall of thickness one, so every simplex
realises inside the union of archived cells and $|K_{t_b}|$ misses the
centre $p$ of each enclosed unarchived cell; a filling $2$-chain would
contract $z$ in $\mathbb{R}^2 \setminus \{p\}$, contradicting winding
number $1$ about $p$. At the fill time $t_d$ the enclosed cells are
archived and the Rips triangles on them, which at $\eps < 2$ are
exactly the triples that are mutually $8$-adjacent, assemble into a
$2$-chain $c$ with $\partial c = z$, so $[z] = 0$ in $H_1(K_{t_d})$
and the bar is $[t_b, t_d)$. Which triangulation is immaterial; only
that one exists. A chamber is the same computation with enclosed
region $W_i \cup I(W_i)$ and $t_b$ the circumnavigation time, with
archiving $I(W_i)$ supplying the filling $2$-chain. At full coverage
every interior is archived, so every chamber and pocket bar has died
and no frontier cell remains. Both conditions are functions of the
archive and are therefore observable.

\emph{(ii)} A decoy's enclosed region is obstacle by
$I(W) = \emptyset$, so its cells are never archived, the winding
obstruction above is permanent, no filling $2$-chain ever exists, and
the bar is $[t_b, \infty)$. Its class is pruned rather than killed:
each rim cell keeps selection probability at least
$s_{\min}/(s^{\TEsub}_{\max}|A_t|) > 0$ at every iteration by
\Cref{eq:tefloor}, so the conditional Borel--Cantelli lemma
\citep{Williams1991}, as in \Cref{lem:trcomplete}, gives infinitely
many selections almost surely. Every one fails, so the counters
exhaust after at most $|\rims(\gamma)|\,u_{\max}$ selections of the
class by \Cref{lem:decoytax} and the class prunes.

\emph{(iii)} At the almost surely finite time $\tau_{\mathrm{cert}}$
when both (i) and (ii) have occurred, $\Gamma_t$ consists of exactly
the $k_d$ decoy classes. The stopping rule is sound: with no frontier
cell, no future archive growth is possible, so no new bar can be born
and no pruned bar can revive.
\end{proof}

\GE{} has no analogue. Counts never certify that a rim is exhausted
rather than unlucky, because a count that has not produced a discovery
is indistinguishable from one that cannot.

\subsection{What is not claimed}
\label{app:theory:scope}

These are statements about the selection distribution and about the
budget a decoy can absorb, conditional on detection and on the gate
budget $u_{\max} \geq q_i^{-1}\log(s^{\TEsub}_{\max}N/s_{\min})$. They do
not assert a sample-complexity separation from \GE{} in general
environments, and the empirical sections are not presented as
verifying one. The chamber-count families of
\Cref{sec:results:chambers} were constructed to satisfy the
geometric hypotheses (\Cref{sec:setup}). Montezuma's
Revenge and the embodied task do not satisfy them: rooms are entered
through doors the agent has often already passed, detection is
partial, and the episode horizon is not matched to the door geometry.
The results there are reported as evidence about the signal in
practice, not about the theorems, though they lend the theorems some
colour.

\section{Implementation details}
\label{app:impl}

On grids, $H_1$ of the cubical complex is computed exactly by flood
fill, each bounded component of unvisited cells being one class, and a
class's persistence under thickening is its inradius, computed by
breadth-first search; the embodied explorer's alpha complex is built
with GUDHI \citep{DBLP:conf/icms/MariaBGY14}.
\Cref{alg:te} states \TE{}'s exploration phase in full.

\begin{algorithm}[h]
\caption{\TE{}, exploration phase. Starred lines ($\star$) are the
additions to \GE{}.}
\label{alg:te}
\small
\begin{algorithmic}[1]
\Require start cell $x_0$; rollout length $L$; \GE{} score constants
  (\Cref{eq:cellscore}); topological weight $\alpha$, exponents
  $p_{\mathrm{topo}}, p_u$, retirement threshold $u_{\max}$,
  rebuild cadence $\tau$
\State $A_0 \gets \{x_0\}$ with the empty trajectory; visitation
  counts and attempt counts $u \gets 0$
\For{$t = 1, 2, \ldots$ until the step budget is spent}
  \TEline every $\tau$ iterations: rebuild $K_t$ on $A_{t-1}$; read
    its live classes $\gamma$, rims $\rims(\gamma)$, magnitudes $b(\gamma)$
    \Comment{\S\ref{sec:method:complex}}
  \TEline $\activeset(\gamma) \gets \{c \in \rims(\gamma) :
    u(c,\gamma) < u_{\max}\}$; prune (seal) $\gamma$ if
    $\activeset(\gamma) = \emptyset$
    \Comment{Eq.~\eqref{eq:ghat}}
  \TEline $\wtopo \gets \alpha\,|A_{t-1}|$; compute $\Shom(c)$ for
    $c \in A_{t-1}$, zero off active rims
    \Comment{Eq.~\eqref{eq:shom}}
  \State \textbf{select} $c \in A_{t-1}$ with probability
    $\mathrm{Score}_{\TEsub}(c) \big/ \sum_{c'} \mathrm{Score}_{\TEsub}(c')$
    \Comment{Eq.~\eqref{eq:score}}
  \State \textbf{return} to $c$: restore its stored state, or walk
    its trajectory
  \For{$\ell = 1, \ldots, L$}
    \Comment{explore}
    \State take $a_\ell \sim \pi_{\mathrm{explore}}$; observe cell $x$;
      \textbf{if} $x \notin A$ \textbf{then} add $x$ with its trajectory
  \EndFor
  \State \textbf{update} visitation counts; replace stored trajectories
    by shorter or higher-scoring ones
  \TEline $u(c,\gamma) \mathrel{+}= 1$ for every live $\gamma$ with
    $c \in \activeset(\gamma)$
    \Comment{regardless of outcome}
\EndFor
\State \textbf{Phase 2:} robustify the best archived trajectories into
  a policy (Backward Algorithm)
  \Comment{not analysed here}
\end{algorithmic}
\end{algorithm}

\subsection{Backends}
\textbf{Lattice (Atari).} Cells carry an integer site within a room
(one lattice per level--room pair). Per room we build the cubical
complex of visited sites over a walled base and read $H_1$ classes as
\{cycle, rim, pocket\} triples; class magnitude is the enclosed
region's inradius, computed as the maximum BFS depth from its boundary.
Class identity across passes is (level--room pair, minimum pocket site), which is
stable while any of the pocket survives, so gate counters persist
across recomputation.
\paragraph{The rim in each backend.}
\label{app:impl:rim}
\Cref{app:theory} defines $\rims(\gamma)$ as the archived cells
$4$-adjacent to an unvisited cell of $R(\gamma)$, walls included. The
TopoGym runs narrow it to the archived cells $4$-adjacent to unvisited
\emph{free} cells the agent has already seen, for every arm: a
wall-facing cell carries no bonus, a chamber has no rim until some of
its interior has been seen, so the entry guarantees of
\Cref{app:theory} apply from that point on, and a decoy, which has no
free interior, never receives the bonus rather than absorbing up to
$|\rims(\gamma)|\,u_{\max}$ draws before it is pruned. The lattice
backend on Atari widens it to every archived cell $8$-adjacent to the
enclosed region, adding the cells that touch it only at a corner; the
counting bounds of \Cref{app:theory} then hold with $|\rims(\gamma)|$
replaced by the size of that shell, which still contains every true
entrance. The embodied backend places the bonus on the landmarks
within the rim support radius of \Cref{tab:hparams}.

\subsection{Embodied adaptation: walking instead of restoring}
\label{app:embodied}

\begin{figure}[h]
\centering
\includegraphics[width=0.7\linewidth]{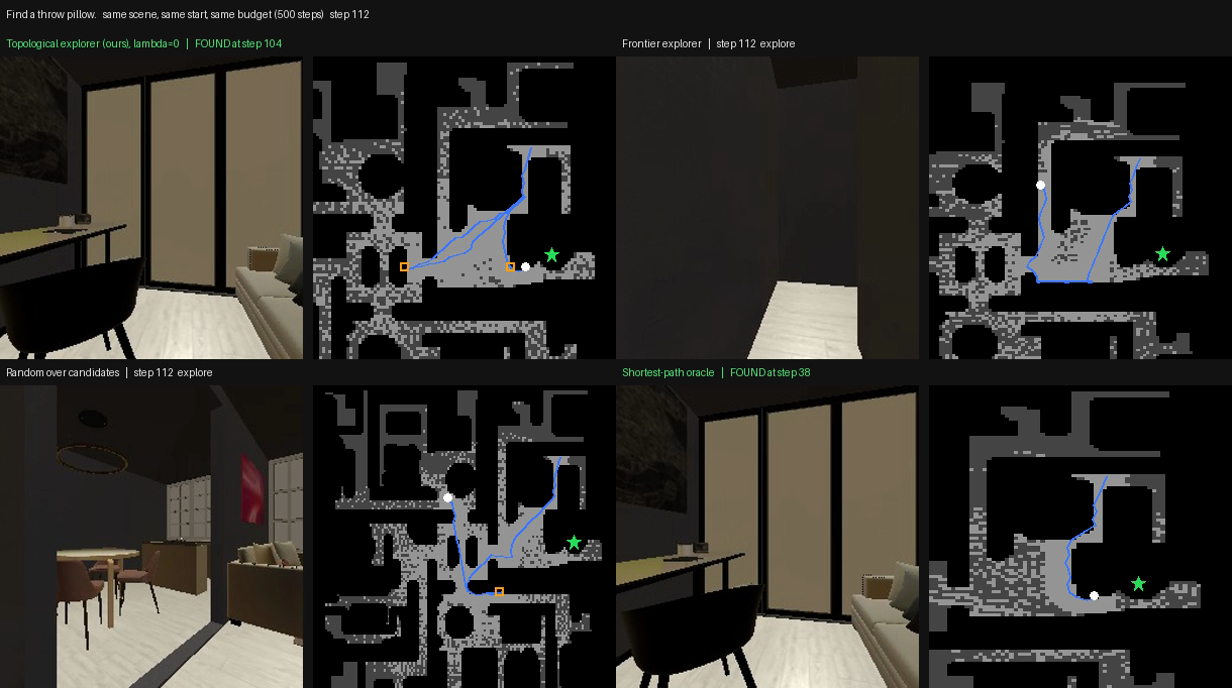}
\caption{The four explorers that generated the training pools on one
episode of the fixed simulator, same scene, start, and 500-step
budget: camera and top-down map per explorer, goal starred. Find a
throw pillow: the topological explorer (top left) has it in view at
step 104 and the shortest-path oracle (bottom right) at step 38; the
frontier and random explorers have not found it by the end of the
budget.}
\label{fig:app:explorers}
\end{figure}

The embodied explorer of \Cref{sec:results:embodied} keeps
\Cref{eq:score}'s selection rule and the gate of
\Cref{sec:method:gates} and changes what they operate on. The
Atari and TopoGym versions rest on an archive of restorable states:
a selected cell is returned to by snapshot restore, so travel is
free and the archive persists for the whole run. A navigation policy
has neither. The adaptation below is what the pilot ran; its
reference planners (nearest-frontier, random walk, and the
shortest-path oracle that produced the demonstration arm) share the
same episode loop and action space, and \Cref{fig:app:explorers}
shows all four on one episode. Evaluation counts behind
\Cref{tab:embodied}: 325--327 unseen-scene and 327--329 held-out-category
episodes per seed for every trained arm; the untrained base on one
evaluation of 325 and 328.

\begin{figure}[t]
\centering
\includegraphics[width=\linewidth]{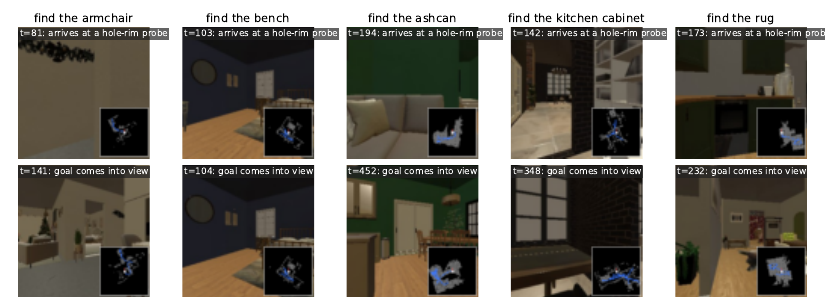}
\caption{What the embodied explorer sees at its own decisions, five
HSSD scenes. Top: the frame at which it arrives at a hole-rim probe,
the topological term's target, with the 64-px egocentric map the
policy is later trained on in the corner. Bottom: the first frame of
the same episode in which the goal object is in view.}
\label{fig:app:embodied_states}
\end{figure}

\paragraph{The map replaces the archive.}
Each episode starts a fresh map and nothing persists across
episodes. Every second step the depth frame (256$^2$, $79^\circ$
HFOV, range 5\,m, hits above 2.2\,m discarded) is unprojected into
surface hits, and the map is a $\delta$-net of those hits with
$\delta = 0.2$\,m, each landmark carrying its sensor-facing normal
and its acquisition time. Rays that returned nothing within range
contribute their far ends to a \emph{frontier}: a 0.5\,m net of
open-ray ends that must snap to the navmesh and retire once a sweep
passes within 1\,m of them.

\paragraph{The complex and its classes.}
The coverage complex is the subcomplex of the alpha complex
\citep{DBLP:journals/dcg/Edelsbrunner95} of the landmarks at radius
$2\delta$, built with GUDHI, that keeps only simplices whose vertex
normals agree, so that the two sides of a wall never merge; the balls
it is built from cover the witnessed surface at the map's
resolution. Filtration is by acquisition time. The classes
the planner acts on are the \emph{rims} of unacquired regions:
boundary landmarks whose tangent-plane neighbourhood has an unseen
side, clustered one rim per region and given a stable identity
across passes by a registry. Indoors a rim is a doorway into an
unseen room, the edge of an occlusion shadow behind furniture, or the
edge of what the depth range reached. A pass runs every 20 steps and
on every arrival.

\paragraph{Candidates and supports.}
A candidate is a live frontier point, a rim's \emph{probe} (a
navigable point at its mouth, snapped to the navmesh), or, failing
both, a pose the rim was first seen from (a re-sweep view). A rim's
support is its frontier points within 1.5\,m, its probe, or its
re-sweep views, in that order. The score is \Cref{eq:score} with the
count term $w_a (1+u)^{-p_a}$ ($w_a = 0.3$, $p_a = 0.5$) over how
often the candidate was chosen, a frontier term of constant weight
1, and the homology term $w_{\mathrm{topo}} \sum_{\gamma\,:\, c
\in \mathrm{active}(\gamma)} (\mathrm{perimeter}(\gamma)/\delta)\,
\psi(c, \gamma)$ with $\psi$ the gate's normalised share over the
rim's active support. $w_{\mathrm{topo}}$ is the constant 1 rather
than the archive-size schedule of the grid version, because
candidate counts indoors are small. The gate is unchanged: $u$ is
incremented on every selection of the pair, the share is weighted
by $(1+u)^{-p_u}$ with $p_u = 1$, a candidate retires at $u_{\max} =
6$, a rim whose support is fully retired is pruned and never
re-nominated, and rim progress resets its counters.

\paragraph{Return is walking.}
The draw is proportional to score times $\exp(-\lambda d)$ with $d$
the geodesic distance to the candidate; the pilot's arm used
$\lambda = 0$, so selection is the proportional rule of the grid
version and walking distance is not penalised. The selected candidate is reached
by a greedy geodesic follower in at most 150 primitive steps
(0.25\,m forward, $30^\circ$ turns), every one of which is executed
and recorded in the policy's own action space, so the trajectory the
policy later imitates is exactly what the explorer did. On arrival
the explorer takes four random moves, senses, and runs a pass; a
frontier candidate retires once visited and a probe once reached.
When no candidate exists it takes a burst of ten random primitives
so that the episode can still end. Episodes are capped at 500 steps.

\paragraph{Approach and stop.}
Once the goal is visible within 4\,m the episode switches to an
approach phase: the follower walks to the reachable goal view point
nearest the object and calls stop. This phase uses the episode's
goal view points, which the oracle demonstrations use throughout;
the exploration phase, which is the part the topological signal
controls and the part that differs between arms, uses depth alone.
Success is the ObjectNav rule (stop within 1\,m geodesic of a goal
view point) and is what \Cref{tab:embodied} reports for the
fine-tuned policies, which drive the same primitive interface at
evaluation.

\paragraph{Comparator arms.}
The six training sets, all trained with the recipe of \Cref{tab:hparams}
and a stop cue in the answer format, are: 455 oracle shortest-path
demonstrations; 352 topological-explorer episodes; the two
comparison explorers below; a 953-episode mixture of the top two
quality bands over eleven explorer variants; and a 642-episode 50/50
mixture of oracle and explorer episodes, the mixtures budgeted to the
topological arm's imitated steps. The
arms are the successful non-holdout episodes of two other explorers
from the same trajectory pool as the topological arm, same scenes,
starts, explorer seed and 500-step budget: the nearest-frontier
explorer (292 episodes, 54.8k imitated steps) and a uniform draw over
the topological planner's candidate set of frontier points and rim
probes (327 episodes, 45.5k steps), against the topological arm's 352
episodes and 54.7k steps; training recipe, seeds 0--4 and evaluation
sets as for every other arm. Primary test: success at 1\,m on unseen
scenes, topological minus frontier, paired by training seed,
two-sided paired $t$ over five seeds at $\alpha = 0.05$. Secondary,
same rule: topological minus random, the held-out-category split,
and goal ever in view. Decision rule: both rows enter the main table
whatever the sign, and the 50/50 oracle+explorer mixture moves to
this appendix with its evaluation extended to the same counts, which
it now has (\Cref{tab:app:embodied_all}).
\Cref{tab:app:embodied_tests} lists every paired comparison;
the per-seed values behind them are in
\texttt{figures/embodied/pilot\_per\_seed.csv} and the tests in
\texttt{pilot\_paired\_tests.csv}, both written by
\texttt{analysis/embodied/pilot\_table.py}.

\begin{table}[H]
\centering
\small
\caption{Every training set of the embodied pilot, including the
oracle+explorer mixture. Success at 1\,m, mean over five training
seeds [95\% $t$-interval], 325--327 unseen-scene and 327--329
held-out-category episodes per seed and arm; the best trained arm per
column in bold.}
\label{tab:app:embodied_all}
\begin{tabular}{lccc}
\toprule
Training data & $n$ & unseen scenes & held-out categories \\
\midrule
base model (untrained) & --- & 0.01 & 0.01 \\
oracle demonstrations & 5 & 0.11 [0.10, 0.12] & 0.12 [0.10, 0.14] \\
frontier explorer data & 5 & 0.15 [0.12, 0.17] & 0.13 [0.10, 0.15] \\
random-over-candidates explorer data & 5 & 0.13 [0.11, 0.15] & 0.12 [0.09, 0.16] \\
topological explorer data & 5 & \textbf{0.18 [0.15, 0.20]} & 0.13 [0.11, 0.15] \\
top-two-band explorer mixture & 5 & 0.16 [0.13, 0.18] & \textbf{0.16 [0.13, 0.18]} \\
oracle + explorer, 50/50 & 5 & 0.18 [0.16, 0.19] & 0.13 [0.11, 0.14] \\
\bottomrule
\end{tabular}

\end{table}

\begin{table}[H]
\centering
\footnotesize
\setlength{\tabcolsep}{4pt}
\caption{Paired-by-seed comparisons between training sets on success
at 1\,m (first named minus second): mean difference over the five
shared training seeds with its 95\% $t$-interval, the two-sided
paired $t$-test $p$-value, and the number of seeds on which the first
arm is ahead. The first row is the primary comparison.}
\label{tab:app:embodied_tests}
\begin{tabular}{llcc}
\toprule
Comparison & split & difference [95\% CI] & $p$ (seeds ahead) \\
\midrule
topological $-$ frontier & unseen scenes & +0.032 [+0.021, +0.044] & 0.001 (5/5) \\
 & held-out categories & +0.002 [-0.030, +0.034] & 0.866 (2/5) \\
topological $-$ random & unseen scenes & +0.047 [+0.011, +0.083] & 0.022 (5/5) \\
 & held-out categories & +0.006 [-0.042, +0.054] & 0.757 (2/5) \\
topological $-$ oracle & unseen scenes & +0.065 [+0.043, +0.088] & 0.001 (5/5) \\
 & held-out categories & +0.014 [-0.026, +0.054] & 0.386 (3/5) \\
frontier $-$ oracle & unseen scenes & +0.033 [+0.012, +0.054] & 0.012 (5/5) \\
 & held-out categories & +0.012 [-0.012, +0.036] & 0.237 (4/5) \\
frontier $-$ random & unseen scenes & +0.015 [-0.023, +0.053] & 0.339 (3/5) \\
 & held-out categories & +0.004 [-0.018, +0.026] & 0.666 (2/5) \\
oracle + explorer 50/50 $-$ topological & unseen scenes & -0.002 [-0.035, +0.031] & 0.884 (2/5) \\
 & held-out categories & -0.004 [-0.038, +0.031] & 0.782 (2/5) \\
\bottomrule
\end{tabular}

\end{table}

\begin{table}[H]
\centering
\footnotesize
\setlength{\tabcolsep}{4pt}
\caption{The same paired comparisons on whether the goal ever came
into view.}
\label{tab:app:embodied_tests_seen}
\begin{tabular}{llcc}
\toprule
Comparison & split & difference [95\% CI] & $p$ (seeds ahead) \\
\midrule
topological $-$ frontier & unseen scenes & +0.039 [+0.009, +0.070] & 0.024 (4/5) \\
 & held-out categories & -0.004 [-0.041, +0.033] & 0.775 (3/5) \\
topological $-$ random & unseen scenes & +0.011 [-0.026, +0.048] & 0.452 (3/5) \\
 & held-out categories & -0.005 [-0.090, +0.080] & 0.871 (1/5) \\
topological $-$ oracle & unseen scenes & +0.106 [+0.071, +0.142] & 0.001 (5/5) \\
 & held-out categories & +0.066 [+0.010, +0.123] & 0.030 (5/5) \\
frontier $-$ oracle & unseen scenes & +0.067 [+0.042, +0.092] & 0.002 (5/5) \\
 & held-out categories & +0.071 [+0.006, +0.136] & 0.039 (5/5) \\
frontier $-$ random & unseen scenes & -0.028 [-0.062, +0.005] & 0.079 (0/5) \\
 & held-out categories & -0.001 [-0.070, +0.068] & 0.963 (2/5) \\
oracle + explorer 50/50 $-$ topological & unseen scenes & -0.015 [-0.063, +0.033] & 0.443 (1/5) \\
 & held-out categories & +0.006 [-0.042, +0.054] & 0.744 (3/5) \\
\bottomrule
\end{tabular}

\end{table}

\begin{figure}[H]
\centering
\includegraphics[width=\linewidth]{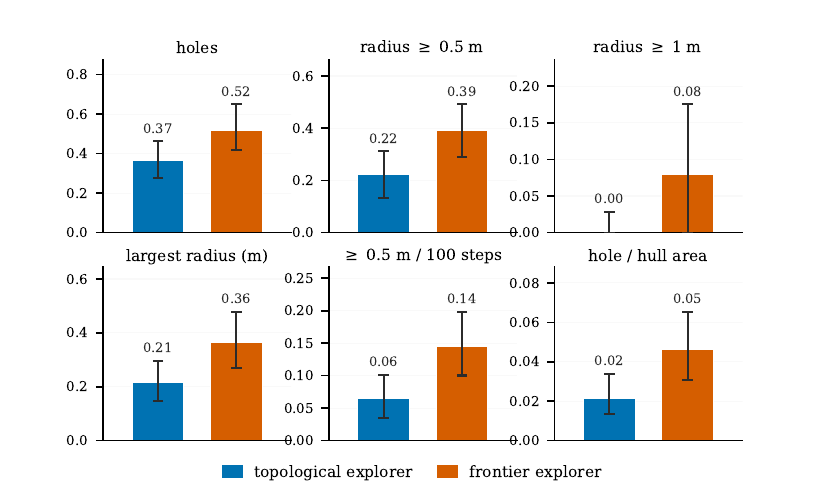}
\caption{Holes in the viewpoint sets of goal-reaching training
trajectories, topological against frontier explorer: interquartile
mean over trajectories with a 95\% stratified-bootstrap interval
(422 and 353 trajectories). Holes are $H_1$ classes of the alpha
filtration over the viewpoint set with persistence at least 0.1\,m;
a hole's radius is that of the largest empty disc it encloses; hole
area is the sum of those discs over the area of the viewpoint set's
convex hull.}
\label{fig:app:viewpoint_holes}
\end{figure}

\paragraph{Holes in the viewpoint sets of the training trajectories.}
The viewpoint set of a trajectory is the set of positions it sensed
from (every second step and every arrival, one point per 0.25\,m
cell). Its holes are loops of viewpoints that enclose floor the
trajectory never stood on: the $H_1$ classes of the alpha filtration
over the two-dimensional viewpoint set, a class born at radius $b$
and dying at radius $d$ being a hole whose loop closes once its
viewpoints are within $2b$ of one another and which is filled by
discs of radius $d$, so $d$ is the radius of the largest empty disc
it encloses. Classes with $d - b < 0.1$\,m, the gap between two
consecutive poses, are not counted. \Cref{fig:app:viewpoint_holes}
and \Cref{tab:app:viewpoint_holes} compare the goal-reaching
trajectories of the topological and frontier explorers, the
trajectories the two arms were trained on, on rliable's
interquartile mean over trajectories with a 95\% stratified-bootstrap
interval \citep{DBLP:conf/nips/AgarwalSCCB21}. Topological
trajectories leave fewer and smaller holes on every statistic: the
probability that a frontier trajectory has more than a topological
one is 0.54 to 0.55 with intervals above 0.5 throughout, holes of
radius at least 1\,m occur in 22\% of topological trajectories against
29\% of frontier ones (interquartile means 0.00 and 0.08), and the enclosed hole area is 2\% of the
viewpoint set's hull for topological trajectories against 5\% for
frontier. The difference is not one of range: frontier trajectories
reach farther (extent radius 3.5\,m against 3.1) and take longer
(173 steps against 117), and the gap persists per 100 steps. On
trajectories that failed to reach the goal, all at the 500-step cap,
the three explorers do not separate (2 to 3 holes each), so the
difference is in how the goal is reached, not in how much ground is
covered.

The random-over-candidates explorer shows as few holes as the
topological one (0.33 against 0.37, hole area 2\% of the hull) and
yet its data trains the weakest explorer policy
(\Cref{tab:app:embodied_all}), for a reason the pool makes visible.
Random succeeds on 73\% of pool episodes against 79\% for the
topological explorer, the gap being on far goals (0.52 against
0.66), and the episodes it does finish are the shorter and more
direct ones (105 steps, fewest decisions): conditioning on success
selects the trajectories in which the uniform draw happened to send
it toward the goal early. That tidiness is luck rather than a
strategy, since nothing in the observation the policy conditions on
produced it, and unconditionally random's trajectories have the same
number of holes as topological ones (0.72 against 0.70 over all
episodes). A rim probe, by contrast, is a target the map itself
defines, so the same egocentric map tends to produce the same move
and the policy sees a consistent mapping from map to action. The
evaluation rows agree: policies trained on random data sight the goal
and come within a metre of it as often as topologically trained ones
(0.23 to 0.29 against 0.23 to 0.31 on unseen scenes) and then fail to
stop (stop rate 0.60 to 0.70 against 0.70 to 0.82).
\Cref{tab:app:viewpoint_holes} is computed by
\texttt{analysis/embodied/viewpoint\_set\_figure.py} from the
per-trajectory CSV \texttt{figures/embodied/viewpoint\_set\_holes.csv},
which the source repository's \texttt{scripts/viewpoint\_set\_holes.py}
writes from the recorded poses of every pool episode.

\begin{table}[H]
\centering
\footnotesize
\setlength{\tabcolsep}{4pt}
\caption{The statistics of \Cref{fig:app:viewpoint_holes} for all
four training sets (goal-reaching trajectories, interquartile mean
[95\% stratified-bootstrap interval]), and the size of the
trajectories.}
\label{tab:app:viewpoint_holes}
\begin{tabular}{lcccc}
\toprule
per goal-reaching trajectory & topological & frontier & random & oracle \\
 & (422) & (353) & (391) & (536) \\
\midrule
holes & 0.37 [0.28, 0.46] & 0.52 [0.42, 0.65] & 0.33 [0.23, 0.43] & 0.00 \\
holes of radius $\ge 0.5$\,m & 0.22 [0.13, 0.31] & 0.39 [0.29, 0.49] & 0.17 [0.08, 0.27] & 0.00 \\
holes of radius $\ge 1$\,m & 0.00 [0.00, 0.03] & 0.08 [0.00, 0.18] & 0.00 [0.00, 0.02] & 0.00 \\
largest hole radius (m) & 0.21 [0.15, 0.30] & 0.36 [0.27, 0.48] & 0.19 [0.11, 0.27] & 0.00 \\
holes $\ge 0.5$\,m per 100 steps & 0.06 [0.03, 0.10] & 0.14 [0.10, 0.20] & 0.06 [0.02, 0.10] & 0.00 \\
hole area / hull area & 0.02 [0.01, 0.03] & 0.05 [0.03, 0.07] & 0.02 [0.01, 0.03] & 0.00 \\
\midrule
viewpoints & 42 & 48 & 38 & 18 \\
steps & 117 & 173 & 105 & 39 \\
extent radius (m) & 3.05 & 3.50 & 3.07 & 1.94 \\
\bottomrule
\end{tabular}

\end{table}

\subsection{Per-game specifics}
Montezuma's Revenge: cells $(\text{level}, \text{room}, \text{keys},
x/8, y/16)$ from RAM, latched across the black transition frames where
room and position update at different times; termination on life loss;
\GE{}'s level multiplier and more-keys neighbour term. Pitfall!:
cells $(\text{room}, x/8, y/16)$ with room from the byte of the game's linear-feedback shift register (LFSR) that generates the world;
\emph{no} termination on life loss (death is a usable left-edge
teleport); rewards scaled $\times 10^{-3}$ and never clipped in the
learned baselines.

\subsection{Hyperparameters}
\label{app:hparams}

\Cref{tab:hparams} lists every hyperparameter behind a reported
number, its final value, and its provenance. Three sources recur.
\emph{Search}: a staged Optuna search of \TE{}'s knobs on TopoGym's
\texttt{tune} split, run once and frozen before any Atari run; the
same winners were selected on both TopoGym suites, which is what
licensed freezing them. \emph{Go-Explore}: the per-game
domain-knowledge cell-score weights of \citet{DBLP:journals/corr/abs-1901-10995},
Appendix A.6, Table 2, used verbatim by both \GE{} and \TE{} so both methods run with identical Atari settings. \emph{Reference}: a baseline's own
paper. Values marked \emph{default} are implementation defaults
never tuned; \emph{hand-set} values are ours, with the reason given.

\begingroup\footnotesize
\setlength{\tabcolsep}{4pt}
\begin{longtable}{>{\raggedright\arraybackslash}p{0.33\linewidth} >{\raggedright\arraybackslash}p{0.27\linewidth} >{\raggedright\arraybackslash}p{0.33\linewidth}}
\caption{Hyperparameters, final values, and provenance.}
\label{tab:hparams} \\
\toprule
Hyperparameter & Value & Source \\
\midrule
\endfirsthead
\toprule
Hyperparameter & Value & Source \\
\midrule
\endhead
\midrule \multicolumn{3}{r}{\emph{continued on the next page}} \\
\endfoot
\bottomrule
\endlastfoot
\multicolumn{3}{l}{\textbf{Archive methods on Atari (\GE{} and every \TE{} arm)}} \\
Budget per run & 15M agent steps (60M frames) & matched to \GE{}'s level-1 solve with domain knowledge (57.6M frames) \\
Frameskip; sticky actions; episode cap & 4; off (deterministic); 100k steps & \GE{} Phase 1; cap hand-set (\GE{} ends episodes on life loss or game over) \\
Explore rollout $L$ & 100 steps & \GE{} \\
Cell (Montezuma) & (level, room, keys, $x/8$, $y/16$) & \GE{} domain-knowledge cells \\
Cell (Pitfall!) & (room byte, $x/8$, $y/16$) & \GE{}, room from the LFSR byte \\
Count exponent $p_a$; $\epsilon_1$, $\epsilon_2$ & 0.5; $10^{-3}$, $10^{-5}$ & \GE{} \\
Montezuma count / neighbour weights & 0, 0, 0 / 0.3 (horiz.), 0.1 (vert.), 10 (more keys) & \GE{} App.~A.6, Table 2 \\
Montezuma level weight $M(c)$ & $0.1^{\,\max\mathrm{level} - \mathrm{level}(c)}$ & \GE{} \\
Pitfall! count / neighbour weights & 1.0, 0, 0.5 / 1.0, 0 & \GE{} App.~A.6, Table 2 \\
\midrule
\multicolumn{3}{l}{\textbf{The topological term (all \TE{} arms, all domains)}} \\
Schedule gain $\alpha$ in $\wtopo(t) = \alpha|A_t|$ & 4.597 & search \\
Magnitude exponent $p_{\mathrm{topo}}$ & 1 & default \\
Gate decay $p_u$; retirement $u_{\max}$ & 0.293; 8 & search \\
Class magnitude $b(\gamma)$ & inradius (lattice); persistence (point cloud) & \Cref{sec:method} \\
Rebuild cadence $\tau$; region & every iteration; one room & default; \Cref{app:impl} \\
\midrule
\multicolumn{3}{l}{\textbf{Local-signal arms}} \\
$w_{\mathrm{ricci}}$ & 115.98 & search \\
Texture slot weights (TopoGym-canonical) & door 19.91, hallway 0.559, wormhole 1.851, drop-adjacent 1.851, water 0.368, ladder 0.057, bridge 0.056 & search \\
Montezuma slots & \texttt{has\_key} 1.0; \texttt{door\_adjacent}, \texttt{openable\_door} = door; \texttt{skull\_adjacent} = drop-adjacent & adaptor default; mapped to the canonical slots \\
Montezuma door detector & colour (232,204,99), $\ge 12$ px/column, $\ge 3$ columns; near = 14 px; skull near = 24 px & hand-set from the sprite \\
Pitfall! slot & \texttt{underground} 0.5 & adaptor default, untuned \\
\midrule
\multicolumn{3}{l}{\textbf{TopoGym hold-out and chamber sweeps (archive arms)}} \\
Budget per world & 1M steps (2M on OpenFieldChamberCount) & benchmark \\
GE base $w_a$, $p_a$, $w_n$ & 0.3, 0.5, 3.0 & search (inherited and confirmed) \\
\TE{} knobs and arm weights & as above & search \\
\midrule
\multicolumn{3}{l}{\textbf{TopoGym gradient baselines (RLlib PPO)}} \\
Budget; early stopping & 1M steps per world; none & benchmark; \Cref{app:repro} \\
Train batch; env runners & 4{,}000; 2 $\times$ 2 envs & default \\
PPO: lr; entropy coefficient & $3 \times 10^{-4}$; 0.01 & grid on \texttt{tune} split: lr $\{10^{-4}, 3 \times 10^{-4}\}$ $\times$ entropy $\{10^{-3}, 10^{-2}\}$ \\
PPO+RND: lr; intrinsic scale & $3 \times 10^{-4}$; 0.05 & grid on \texttt{tune} split: lr $\times$ scale $\{0.05, 0.5\}$ \\
PPO+ICM: lr; $\eta$; $\beta$; feature dim & $3 \times 10^{-4}$; 0.5; 0.2; 64 & standard values, not searched: lr as PPO/RND chose; $\eta$ unstated in \citet{DBLP:conf/icml/PathakAED17}, 0.5 beat 0.05 on one single-layout search; $\beta$ reference; dim default \\
\midrule
\multicolumn{3}{l}{\textbf{Atari gradient baselines (PPO+RND, PPO+ICM)}} \\
Budget & 25M agent steps (100M frames) & sum of both \TE{} phases \\
Observation & 84$\times$84 grey, 4-frame stack, sticky $p = 0.25$, no no-ops & reference \citep{DBLP:conf/iclr/BurdaESK19} \\
Envs; rollout; epochs; minibatches & 32; 128; 4; 4 & reference, except envs: RND's preliminary 32 (its default is 128) \\
lr; clip; entropy; value coef. & $10^{-4}$; 0.1; 0.001; 0.5 & reference; value coef.\ as the reference code's $0.5\cdot$MSE \\
$\gamma_{\mathrm{ext}}$, $\gamma_{\mathrm{int}}$, $\lambda$; advantage coefs. & 0.999, 0.99, 0.95; 2.0 ext, 1.0 int & reference \\
RND predictor update proportion & 0.25 & RND's value for 128 envs; its code uses 1.0 at 32 (deviation) \\
ICM $\beta$; $\eta$ & 0.2; 0.5 & $\beta$ of \citet{DBLP:conf/icml/PathakAED17}, but applied to the inverse loss (losses weighted 0.2/0.8, not 0.8/0.2); $\eta$ as TopoGym \\
Observation-normalisation warm-up & 50 steps & hand-set (RND's code: 6{,}400) \\
\midrule
\multicolumn{3}{l}{\textbf{Embodied explorer (MultiON on HSSD)}} \\
Depth sensor & 256$^2$, $79^\circ$ HFOV, 5 m range, hits above 2.2 m dropped & simulator config \\
Sensing / topology cadence & every 2 / every 20 steps & hand-set \\
Landmark net $\delta$; frontier net & 0.2 m; 0.5 m & hand-set \\
Rim support radius; walk cap; episode cap & 1.5 m; 150 steps; 500 steps & hand-set \\
Count term $w_a$, $p_a$; frontier weight & 0.3, 0.5; 1 & as TopoGym base \\
$\wtopo$; $p_u$; $u_{\max}$ & 1 (constant); 1; 6 & hand-set (few candidates indoors) \\
Distance discount $\lambda$ & 0 & pilot arm \\
\midrule
\multicolumn{3}{l}{\textbf{VLA fine-tuning}} \\
Base model & InternVLA-N1-System2 (Qwen2.5-VL-7B) & \citet{internvla2025n1} \\
LoRA rank; $\alpha$; dropout & 16; 32; 0.05 & default \\
Optimiser; lr; schedule; weight decay & AdamW; $10^{-4}$; cosine to zero after 100 linear warm-up steps; 0 & hand-set \\
Steps; batch & 3{,}000; 4 & hand-set (48 GB card) \\
History; image size; action chunk & 4 frames; 224; 4 & hand-set \\
\end{longtable}
\endgroup

\section{Extended results}
\label{app:extended}

\subsection{Montezuma's Revenge: the 200-seed test and what each arm does}
\label{app:montezuma}

\begin{figure}[H]
\centering
\includegraphics[width=\linewidth]{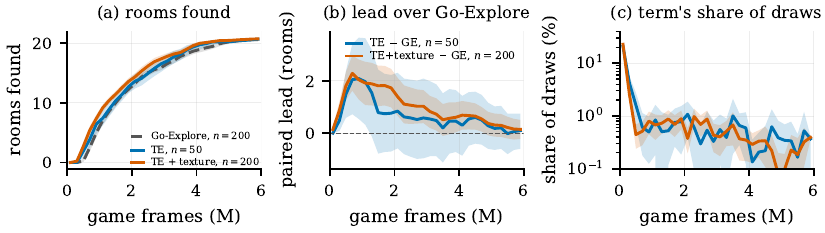}
\caption{Where the term acts on Montezuma's Revenge, all Phase-1
seeds. (a) Rooms found against game frames, IQM with a 95\% bootstrap
band. (b) Paired lead over \GE{} in rooms found, mean with a 95\%
band. (c) Share of draws the homology term carries, per 200k-frame
bin, log scale.}
\label{fig:regime}
\end{figure}

\paragraph{The 200-seed test.}
\emph{Hypothesis:} \TE{}+texture achieves a higher best-trajectory
score than \GE{} at matched budget (15M steps, 60M frames,
Phase~1). \emph{Design:} arms \GE{} and \TE{}+texture at
$n = 200$ seeds each, the same tooling, budget and stride throughout.
\emph{Metric:} best archived trajectory return. \emph{Test:} paired
Wilcoxon signed-rank on the 200 shared seeds, two-sided,
$\alpha = 0.05$; secondary, rliable IQM with a stratified bootstrap
95\% CI and Cohen's $d$ on the paired differences. Power is about 95\% at
$d = 0.26$ and 80\% at $d = 0.20$. \emph{Outcome:} $W = 8029$,
$p = 0.032$, paired $d = 0.18$, texture higher on 113 pairs, \GE{} on
84, 3 ties; IQMs in \Cref{tab:montezuma}. Plain \TE{}, ricci and
both were not taken past 50 seeds: they are null there
(\Cref{tab:montezuma_all}) and nothing in their mechanism predicts
otherwise on this game.

\begin{table}[h]
\centering\small
\caption{Every Montezuma arm, same statistics as \Cref{tab:montezuma};
the local-signal arms were not extended past the 50-seed wave.}
\label{tab:montezuma_all}
\begin{tabular}{lcccl}
\toprule
Method & $n$ & Median & IQM [95\% CI] & paired $p$ vs.\ \GE{} \\
\midrule
\GE{} & 200 & 31{,}400 & 31{,}629 [28{,}456, 34{,}382] & reference \\
\TE{} (plain) & 50 & 27{,}500 & 28{,}173 [23{,}308, 32{,}765] & $p = 0.699$ ($n = 50$) \\
\TE{} + texture & 200 & \textbf{36{,}000} & \textbf{35{,}057} [31{,}125, 38{,}653] & $p = 0.032$ ($n = 200$) \\
\TE{} + ricci & 49 & 28{,}500 & 26{,}892 [19{,}744, 34{,}820] & $p = 0.739$ ($n = 49$) \\
\TE{} + both & 44 & 27{,}000 & 26{,}141 [20{,}818, 32{,}946] & $p = 0.731$ ($n = 44$) \\
\bottomrule
\end{tabular}

\end{table}

\begin{figure}[t]
\centering
\includegraphics[width=\linewidth]{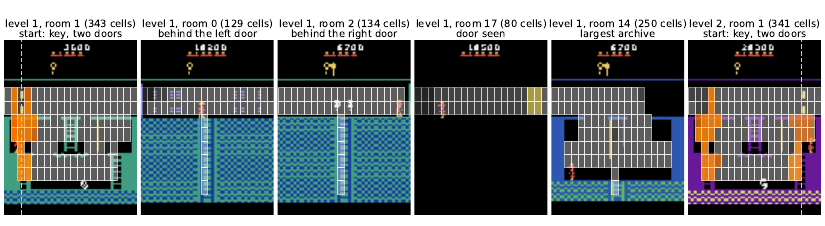}\\[3pt]
\includegraphics[width=0.55\linewidth]{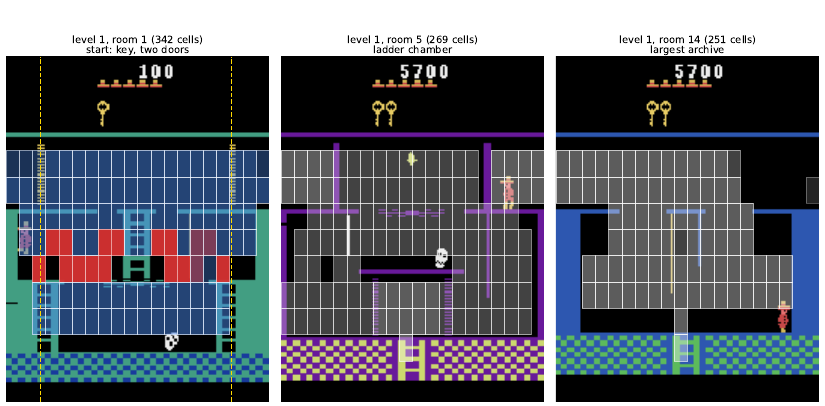}
\caption{Inside the archive on Montezuma's Revenge. Each frame is an
archived state re-rendered by replaying its stored action string;
boxes are the archived cells of that room in screen coordinates
(x = RAM[42], y from RAM[43]). Top, a \TE{}+texture run: cells
carrying \texttt{door\_adjacent} (yellow) and \texttt{openable\_door}
(orange), dashed lines the door columns the detector found in the
live frame; the rooms behind the doors and the level-2 copy of the
start room with its own doors. Bottom, a \TE{}+ricci run: cells with
negative Ollivier--Ricci curvature on the archive graph (red) against
the rest (blue); the negative cells sit at the room's pinches.}
\label{fig:app:mz_states}
\end{figure}

\paragraph{Where the texture signal comes from.}
The adaptor emits per-cell slots from the live frame and RAM: key
possession (\texttt{has\_key}, from the inventory bits), and door
adjacency, which is \emph{seen} rather than read: the detector
matches the door sprite's colour in the frame, requires at least 12
pixels in a column and three adjacent columns, which separates a
door from a key and from the one-column rope, and since the sprite
disappears when a door opens the same detector observes the door's
state, so an opened door stops being an entrance without any rule
saying so. \texttt{openable\_door} is door adjacency while holding a
key. The arm acts in two ways. Each slot adds a staleness-weighted
bonus to the cell's score (\Cref{app:impl}; the door slots carry the
TopoGym-tuned weight of about 20, key possession 1). And when a live
class's active rim contains a cell carrying \texttt{openable\_door}
or \texttt{door\_adjacent}, the class's whole mass goes to those
cells, turning ``somewhere on this rim'' into ``this door, and only
while you hold the key'' (\Cref{fig:app:mz_states}, top). Where no
door is in view the arm is inert and the term is plain \TE{}.

\paragraph{Who carried the door draws.}
\label{app:montezuma:doors}
The two ways can be told apart from the selection logs. Every draw
records the selected cell, its total score $S$, and the homological
mass $H$ on it; the persisted archive records each cell's slots. On
Montezuma's Revenge $S = M(\text{base} + H' + T')$ with $M$ the level
multiplier, base the neighbour term (in $\{1.1, 11.1\}$ up to $0.8$,
the large value while the ``same cell, one more key'' neighbour is
missing, which is almost always) and $T'$ the texture bonus. $H/S$ is
exact. $T'$ is not logged, so a door draw with $H/S \le 1/2$ is binned
by $S/M - H/M = \text{base} + T'$: above $22.9$ the texture bonus
exceeds the base whichever base applied, below $2.2$ there is no
room for a bonus, and between the two the answer depends on which
base applied. \GE{} has $T' = 0$, so its texture bin must be empty,
and it is. \Cref{tab:door_draws} gives the result over every
Phase-1 seed. Narrowing is engaged (the door cell carries
homological mass, and that mass is the majority of its score) on
$1.2\%$ of the texture arm's door draws, against $2.3\%$ of plain
\TE{}'s, where a door sits on a rim only by chance; the arm does not
raise the count of homology-carried door draws. What it raises is
the door-draw rate, from $1.9\%$ and $1.8\%$ of all draws in \GE{} and plain
\TE{} to $2.9\%$, and every one of those extra draws is carried by
the slot bonus: $35\%$ of its door draws are unambiguously
texture-carried where \GE{} has none. The two kinds of door draw lead to different outcomes.
Homology-carried door draws enter the class they were drawn for
($23$ entries per $1{,}000$) but seldom collect reward ($22$
reward-bearing episodes per $1{,}000$); slot-carried door draws
enter nothing ($0.4$ per $1{,}000$) and collect reward at $83$ per
$1{,}000$. Across all draws the arm also shifts about four points of
selection mass from unslotted cells to key-holding and
skull-adjacent ones. Read together, the Montezuma gain is what the
door and key cues buy when added to \GE{}'s score inside the \TE{}
loop, and the logs does not attribute it to the homological
term; a \GE{}+texture arm with the same slot bonus and no
topological pass is the control that would settle the attribution,
and it has not been run. The table is recomputed from the episode
and archive logs by \texttt{analysis/montezuma\_door\_draws.py}.

\begin{table}[h]
\centering\small
\caption{Door draws on Montezuma's Revenge, Phase 1, all seeds.
A door draw is a selection landing on a cell whose slots include
\texttt{door\_adjacent}. Carried by: homology when the cell's
homological mass is the majority of its score; texture when the
non-homological part of the score exceeds $22.9$ (the texture bonus
exceeds the largest possible base); either when it lies between
$2.2$ and $22.9$; base below $2.2$. Yield: episodes launched from
the draw that collected reward, and class entries, per $1{,}000$
door draws in each bin.}
\label{tab:door_draws}
\resizebox{\linewidth}{!}{\begin{tabular}{lcccccccccc}
\toprule
& & Door draws & \multicolumn{4}{c}{carried by (\% of door draws)} & \multicolumn{3}{c}{reward episodes / class entries per 1{,}000 door draws} \\
\cmidrule(lr){4-7}\cmidrule(lr){8-10}
Method & $n$ & (\% of draws) & homology & texture & either & base & homology & texture & either \\
\midrule
\GE{} & 200 & 1.88 & 0 & 0 & 89.6 & 10.4 & 0 / 0 & 0 / 0.0 & 55 / 0.00 \\
\TE{} & 50 & 1.85 & 2.3 & 0 & 87.5 & 10.1 & 22 / 21 & 0 / 0.0 & 60 / 0.01 \\
\TE{} + texture & 200 & 2.86 & 1.2 & 34.6 & 63.9 & 0.4 & 22 / 23 & 83 / 0.4 & 84 / 0.01 \\
\bottomrule
\end{tabular}}

\end{table}

\begin{figure}[h]
\centering
\includegraphics[width=\linewidth]{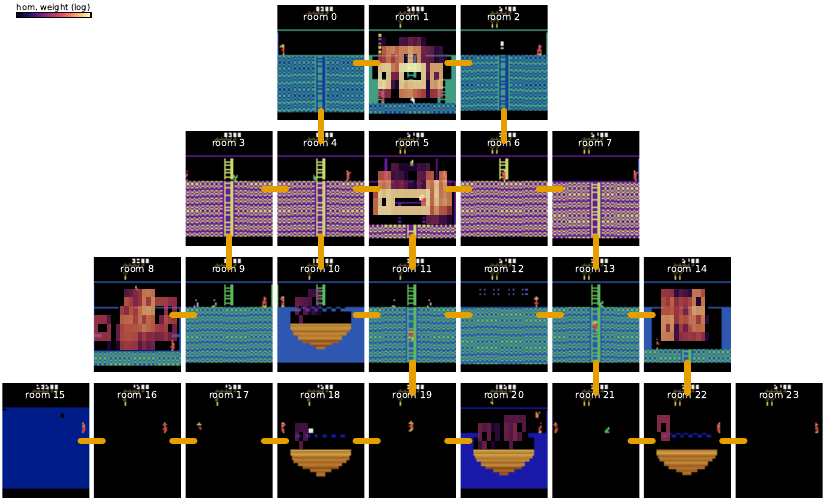}
\caption{Level~1 of Montezuma's Revenge in the layout learned from
the passages the agent crossed (bars across shared edges), each room
re-rendered from an archived state, with every lattice cell tinted by
the homological weight the term drew through it over the 50 plain
\TE{} seeds (log scale; untinted where the archive never surrounded
anything). Every tinted structure is a solid: the map shows where the
decoy cost was incurred.}
\label{fig:app:mz_map}
\end{figure}

\paragraph{Where the term's weight went.}
\label{app:montezuma:map}
\Cref{fig:app:mz_map} lays out the 24 rooms of level~1 in the
arrangement learned from the passages the agent crossed (a room
change inside one explore phase at the left or right edge, or through
the floor or ceiling; \texttt{analysis/montezuma\_room\_layout.py})
and tints every lattice cell by the homological share of the draws
that landed on it, summed over the 50 plain \TE{} seeds. The weight
sits on four solids: the central platform of the start room
(room~1), the block of room~5, the wall of room~8 and the block of
room~14, with traces on the lava-bridge rooms 10, 18, 20 and 22.
None of these is a pocket of free space. Of the 83{,}819
homology-carried draws on level~1, 1.9\% launched an episode that
entered its class and 9.6\% ended in a retirement; room~5's block,
which holds the skull, took the most draws and entered 1.5\% of the
time. This is the bounded cost of decoys of \Cref{app:theory}, incurred on the only
structures a per-room complex can enclose, and it is the mechanism of
the term's neutrality here: the doors are room exits on the
perimeter and never lie on a rim, and the corridor rooms enclose
nothing at all. The enclosure that would matter is one level up, and on this level
it is rare. Treating rooms as cells of a cubical complex on the learned map,
an unvisited room is a pocket when the four rooms sharing an edge
with it have been visited; only rooms 4, 5, 6 and 9 to 13 have four
such neighbours inside the pyramid, and over the 190 runs with room-entry logs the ring closed
before entry around room~12 in 15 and around room~5 in 4
(\Cref{fig:room12}a). Room~14,
the room most often found last, is a dead end on the perimeter, not
an enclosure.

\begin{figure}[p]
\centering
\includegraphics[width=\linewidth]{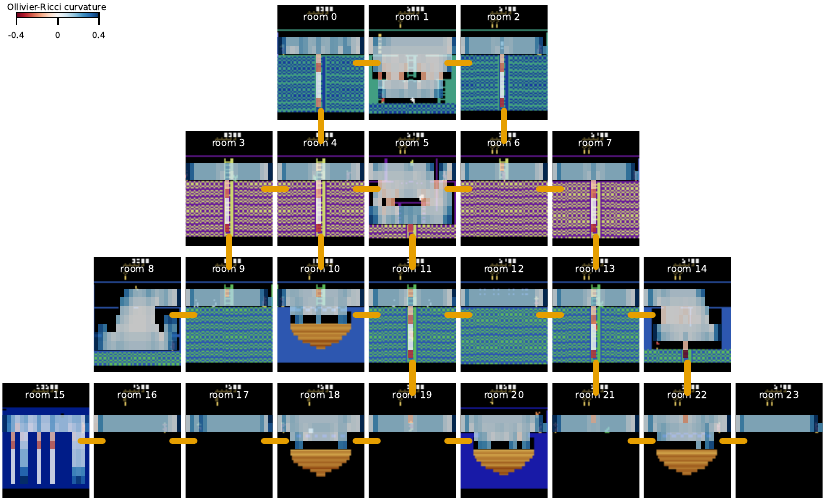}
\caption{Ollivier--Ricci curvature on level~1 of Montezuma's Revenge:
the curvature of the archive graph of one plain \TE{} seed, lattice
neighbours inside a room plus the crossed passages, on the learned
room layout (red negative, blue positive;
\texttt{analysis/montezuma\_ricci\_map.py}).}
\label{fig:app:ricci_mz}
\end{figure}

\paragraph{The room-scale complex, run.}
\label{app:montezuma:rooms}
We ran that complex as an arm: rooms as the cells of a cubical complex
on the learned map, a pocket's support the archived cells of each
neighbouring room nearest to it, the term otherwise unchanged; alone
(\TE{} + room scale) and on top of the curvature arm, 50 seeds each,
paired against the campaign's \GE{} and \TE{} seeds
(\Cref{tab:rooms_wave,tab:rooms_wave_paired}). The ring closed before entry in 7 of the 100
runs carrying the term, 8 classes in all (\Cref{tab:room_term,tab:room_term_b}): six
around room~12 and two around room~5, each with 56 to 86 cells of
support, drawing between 1 and 448 times before the pocket was
entered. While a class was live the term ran the selection: 33 to
45\% of all draws in that window landed on its rim, and on those
draws the homology term was 72 to 82\% of the selected cell's
score (0.01 for the one-draw room~5 class of seed 29). The two room~5 pockets were entered within two thousand
frames of closing.
Of the six room~12 pockets, four were entered by an episode launched
from the rim, and five were entered sooner than the same seed
reached the room without the term, by 186k to 341k frames in four
and by 853k in the fifth (seed 29, whose two runs had already
parted at its room~5 pocket), and one later, by 96k
(\Cref{fig:room12}). None of it reaches the score: \TE{} + room scale equals plain \TE{} on 49 of 50 seeds, and
every arm is even with \GE{} (24 to 26 wins of 50, $p \geq 0.6$).
The curvature arm here runs on the passage-aware graph, where a
doorway or ladder is a one-cell-wide edge between rooms
(\Cref{fig:app:ricci_mz}); at 50 seeds it is even with \GE{} and with
\TE{} as well. The event the theory needs is real on this game and
the term does what it should when it occurs; it is too rare, and
its pocket too cheap, to move a 60M-frame score.

\begin{table}[h]
\centering
\small
\caption{The room-scale wave: best archived trajectory return at 60M
frames, IQM [percentile-bootstrap 95\% CI] over seeds 0--49
(\texttt{analysis/montezuma\_rooms\_wave.py}).}
\label{tab:rooms_wave}
\setlength{\tabcolsep}{5pt}
\begin{tabular}{lrr}
\toprule
Arm & $n$ & IQM score [95\% CI] \\
\midrule
\GE{} & 50 & 31,138 [25,800, 35,581] \\
\TE{} & 50 & 28,173 [23,511, 33,073] \\
\TE{} + room scale & 50 & 27,800 [22,950, 33,116] \\
\TE{} + ricci & 50 & 26,265 [20,976, 32,512] \\
\TE{} + ricci + room scale & 50 & 26,069 [19,942, 33,078] \\
\bottomrule
\end{tabular}

\end{table}

\begin{table}[h]
\centering
\small
\caption{The room-scale wave, paired: wins/losses/ties and the
two-sided Wilcoxon $p$ of each arm against \GE{} and against \TE{}
on the same seeds.}
\label{tab:rooms_wave_paired}
\setlength{\tabcolsep}{5pt}
\begin{tabular}{lrrr}
\toprule
Arm & $n$ & vs \GE{} (W/L/T, $p$) & vs \TE{} (W/L/T, $p$) \\
\midrule
\GE{} & 50 & -- & 24/26/0, 0.70 \\
\TE{} & 50 & 26/24/0, 0.70 & -- \\
\TE{} + room scale & 50 & 26/24/0, 0.69 & 0/1/49, -- \\
\TE{} + ricci & 50 & 24/24/2, 0.70 & 24/24/2, 0.87 \\
\TE{} + ricci + room scale & 50 & 24/24/2, 0.62 & 23/25/2, 0.98 \\
\bottomrule
\end{tabular}

\end{table}

\begin{table}[h]
\centering
\small
\caption{Every room-scale class born before its pocket was entered,
over the 100 runs carrying the term: the enclosed room, the frame at
which the ring closed (millions), the support in cells, the draws the class
received before entry, and the share of all selections in that
window that landed on the rim.}
\label{tab:room_term}
\setlength{\tabcolsep}{5pt}
\begin{tabular}{llrrrrr}
\toprule
Arm & Seed & Room & Closed (M) & Support & Draws & On rim (\%) \\
\midrule
\TE{} + room scale & 18 & 12 & 2.59 & 83 & 331 & 40 \\
\TE{} + room scale & 46 & 12 & 1.60 & 61 & 448 & 42 \\
\TE{} + ricci + room scale & 14 & 12 & 4.32 & 59 & 30 & 39 \\
\TE{} + ricci + room scale & 29 & 5 & 1.08 & 82 & 1 & 33 \\
\TE{} + ricci + room scale & 29 & 12 & 1.77 & 75 & 80 & 36 \\
\TE{} + ricci + room scale & 33 & 12 & 1.45 & 56 & 10 & 45 \\
\TE{} + ricci + room scale & 35 & 12 & 3.84 & 86 & 416 & 37 \\
\TE{} + ricci + room scale & 42 & 5 & 0.72 & 79 & 5 & 33 \\
\bottomrule
\end{tabular}

\end{table}

\begin{table}[h]
\centering
\small
\caption{The same classes: the homological share of the selected
cell's score on the rim draws, and the frame at which the room was entered in
this run (starred where the entering episode was launched from the
rim) and in the same seed run without the term.}
\label{tab:room_term_b}
\setlength{\tabcolsep}{5pt}
\begin{tabular}{llrrrl}
\toprule
Arm & Seed & Room & Hom.\ share & Entered (M) & Without term (M) \\
\midrule
\TE{} + room scale & 18 & 12 & 0.82 & 2.75* & 3.09 \\
\TE{} + room scale & 46 & 12 & 0.79 & 1.81 & 1.71 \\
\TE{} + ricci + room scale & 14 & 12 & 0.76 & 4.34 & 4.57 \\
\TE{} + ricci + room scale & 29 & 5 & 0.01 & 1.08* & 1.08 \\
\TE{} + ricci + room scale & 29 & 12 & 0.76 & 1.82* & 2.67 \\
\TE{} + ricci + room scale & 33 & 12 & 0.75 & 1.45* & 1.75 \\
\TE{} + ricci + room scale & 35 & 12 & 0.80 & 4.05* & 4.23 \\
\TE{} + ricci + room scale & 42 & 5 & 0.72 & 0.73 & 0.73 \\
\bottomrule
\end{tabular}

\end{table}

\begin{figure}[t]
\centering
\includegraphics[width=\linewidth]{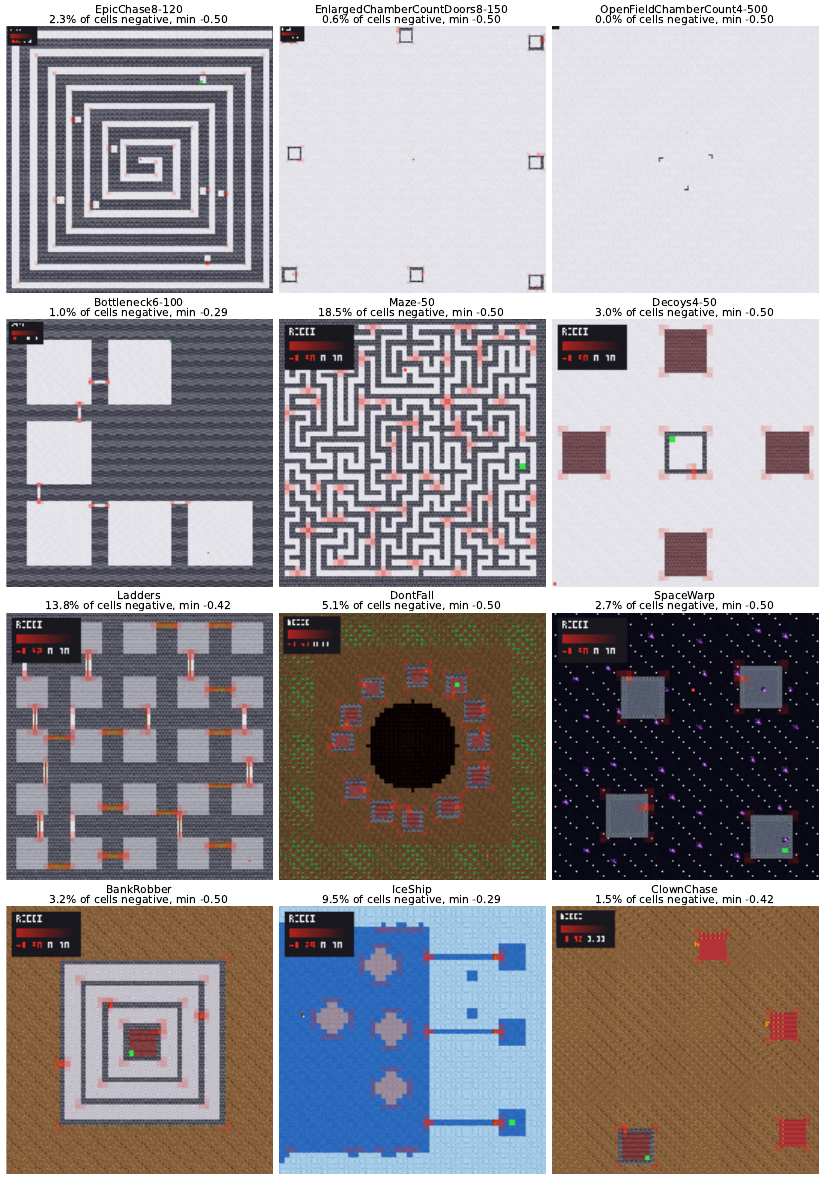}
\caption{Ollivier--Ricci curvature on TopoGym: twelve worlds
rendered by TopoGym with its curvature heat map, every free cell
tinted red in proportion to how negative its curvature is over the
world's free-cell graph (\texttt{analysis/topogym\_ricci\_panels.py});
the negative cells are doorways, corners and junctions.}
\label{fig:app:ricci}
\end{figure}

\paragraph{What curvature sees.}
The ricci arm computes Ollivier--Ricci curvature on the graph the
archive induces (the lattice relation restricted to archived cells,
never the unexplored world), with exact per-edge $W_1$ by assignment
on unit-expanded measures, and gives a bonus proportional to how far
below zero a cell's mean incident curvature sits. Negative curvature
marks pinches. In the starting room of a \texttt{te-ricci} run
(342 archived cells, curvature defined on all) 13\% of cells are
negative, with the two most negative at the symmetric left and right
descents, exactly where the room pinches
(\Cref{fig:app:mz_states}, bottom). That is the signal the
chamber worlds reward (\Cref{sec:results:chambers}); on this game a
pinch is rarely where the key is, and the arm is null.
\Cref{fig:app:ricci,fig:app:ricci_mz} show the field over whole
worlds. On TopoGym
the negative cells are exactly the chamber doorways, corridor corners
and maze junctions: at most $2.3\%$ of the free cells in the chamber
families, $14$--$19\%$ in the maze and the Ladders scenario. On
Montezuma's Revenge, over a full level-1 archive with the crossed
passages added as edges of the archive graph
(\texttt{analysis/montezuma\_ricci\_map.py}), $6.7\%$ of sites are
negative and they sit where a ladder meets a corridor; the passages
between rooms join one corridor to another and are as flat as the
corridors themselves (mean curvature $+0.11$ at passage sites against
$+0.13$ elsewhere). At the archive's $8 \times 16$ binning the game is
a graph of thin corridors, which is why a pinch detector finds little
to reward there and much in a world of rooms.

\subsection{Pitfall!: more rooms, no more score}
\label{app:pitfall}

\begin{figure}[t]
\centering
\includegraphics[width=\linewidth]{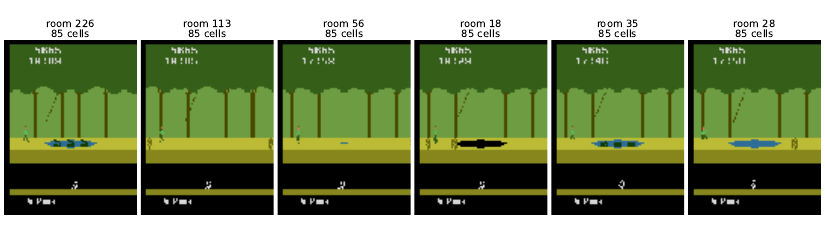}
\caption{Archived states on Pitfall!, replayed from a \TE{} run: six
of the 52 rooms it reached, each a thin strip with one hazard
(crocodiles, rolling logs, a tar pit, quicksand) and a tunnel below.
Every room saturates at the same $\sim$85 lattice cells, so the
archive has almost no per-room shape for the homology term to
read.}
\label{fig:app:pf_states}
\end{figure}

Pitfall! is a chain of 255 thin, nearly linear rooms with sparse
positive reward and dense small negative rewards; cells are
$(\text{room}, x_{\mathrm{bin}}, y_{\mathrm{bin}})$ with the room
taken from the world-generating LFSR byte, which disambiguates the
visually aliased rooms that defeat downscaled-frame cells. It is the
regime with the least for the homology term to find, since the
environment has almost no enclosed structure, and most rooms carry
no reward, so even rooms found need not raise the score. The campaign was run at
the flagship budget on every arm so that the result is a measurement
rather than an assumption.

\begin{table}[h]
\centering\small
\caption{Pitfall!, Phase 1, 60M game frames, matched budgets,
$n = 10$ seeds per arm. Frames to $k$ rooms (median, 95\% bootstrap
CI, in millions) and rooms discovered; the score is in
\Cref{tab:pitfall_score}. Frames-to-$k$ are indistinguishable across
arms at every $k$.}
\label{tab:pitfall}
\setlength{\tabcolsep}{4pt}\footnotesize
\begin{tabular}{lccc}
\toprule
Method & Frames to 15 rooms (M) & Frames to 20 rooms (M) & Rooms \\
\midrule
\GE{} & 0.6 [0.4, 0.9] & 1.6 [1.1, 2.1] & 67 [58, 73] \\
\TE{} & 0.5 [0.5, 0.6] & 1.6 [1.3, 2.1] & 67 [56, 76] \\
\TE{} + ricci & 0.6 [0.4, 0.7] & 1.3 [0.8, 2.2] & 74 [57, 78] \\
\TE{} + texture & 0.5 [0.4, 0.7] & 1.6 [1.3, 2.2] & \textbf{78} [73, 83] \\
\TE{} + both & 0.6 [0.3, 0.7] & 1.7 [1.0, 2.9] & 72 [55, 78] \\
\TE{} + debounce & 0.5 [0.3, 1.0] & 2.3 [1.7, 3.3] & 75 [57, 80] \\
\midrule
PPO+RND (100M) & --- & --- & 9 [9, 10] \\
PPO+ICM (100M) & --- & --- & 7 [7, 7] \\
\bottomrule
\end{tabular}

\vspace{8pt}
\caption{Pitfall!, the same runs as \Cref{tab:pitfall}: best archived
trajectory score (median, 95\% bootstrap CI). The PPO baselines keep
no archive and so have no archived score.}
\label{tab:pitfall_score}
\begin{tabular}{lc}
\toprule
Method & Score \\
\midrule
\GE{} & 6{,}926 [4{,}000, 6{,}956] \\
\TE{} & 6{,}646 [3{,}976, 6{,}856] \\
\TE{} + ricci & 6{,}739 [3{,}861, 6{,}910] \\
\TE{} + texture & \textbf{6{,}956} [5{,}426, 7{,}340] \\
\TE{} + both & 5{,}427 [0, 11{,}970] \\
\TE{} + debounce & 5{,}803 [3{,}855, 7{,}328] \\
\bottomrule
\end{tabular}

\end{table}

\paragraph{The plain term neither helps nor hurts.} At five seeds plain \TE{}
trailed \GE{} (median 56 rooms against 67); at ten the deficit is
gone (67 against 67, overlapping intervals) and frames to every
milestone coincide. The mechanism logs
(\Cref{sec:results:montezuma}) says why nothing happens in either
direction: Pitfall!'s rooms are thin horizontal strips --- the
archive occupies a median of 5 of 16 vertical bins per room --- so
sparse visits create one-cell ``holes'' that are coverage
artefacts, not chambers. About 100 such classes are born per run (mean over ten seeds), 82
of which die naturally as coverage fills them, and the remainder
absorb bonus draws until retirement; rim draws account for
$\sim$1.6\% of selections. This is the intended behaviour when there is nothing to detect, and its cost is the
bounded per-class cost and the throughput cost of the pass
(\Cref{sec:results:compute}).

\paragraph{Texture finds more rooms but no more reward.} As on
Montezuma's Revenge, the texture arm is the one that moves: its
\texttt{underground} slot gives a bonus on tunnel cells, the only
route past the walls that partition the surface, and it discovers
more rooms than every other arm (78 [73, 83] against 67 [58, 73] for
\GE{}, 16\% more), an effect that survived the seed extension. Score
does not follow (6{,}956 against 6{,}926): Pitfall!'s reward sits in
a handful of treasure rooms, so the extra rooms add hazards but no reward. The two-term arm is within
\GE{}'s interval and the curvature arm just above it (74 against
[58, 73]), so the two terms do not compose here, and the debounce arm is slowest to 20 rooms. Read together with Montezuma's Revenge, the two games suggest the signal's value depends
on enclosed structure to detect and on that structure being worth
entering, and that the first condition is measurable from the agent's
own logs rather than requiring prior knowledge of the
environment.

\subsection{Negative controls: cleaning the signal}
\label{app:negative}

Pitfall!'s artefact classes (\Cref{app:pitfall}) invite a filter that
removes them and leaves the real chambers. Two such filters were
designed, calibrated, and rejected. Both are reported here rather
than in the main text because neither is part of \TE{}: they are
ablations of $\Sloc$ that failed.

\paragraph{Spatial magnitude thresholding (rejected before running).}
The natural first remedy for Pitfall!'s artefact classes is to ignore
classes below a magnitude threshold. Calibrating that threshold
against a coverage-noise null --- classes arising in \GE{} runs'
reconstructed visited sets, plus synthetic Bernoulli-visited lattices
matched to each game's room geometry --- shows that null classes
concentrate at inradius 1 (99th percentile 1 on Pitfall-shaped
lattices, 2 on Montezuma-shaped ones), \emph{and so do the real
chambers} at this cell resolution. A threshold that removes the
artefacts also silences the signal, so the variant was recorded as a
negative control rather than run.

\paragraph{Temporal debouncing (run, \texttt{debounce}).}
This variant weights a class in proportion to how improbable its
survival is under a coverage-noise null: the arm multiplies a
class's mass by the empirical CDF of null class lifetimes evaluated
at the class's current age, so a class earns funding by outliving
what noise would predict. The null was measured treatment-blind ---
from \GE{} logs and synthetic lattices only, never from \TE{}
runs --- which leaves the rule with no threshold or percentile
parameter to tune. It fails, and the calibration says why: measured
null lifetimes are long (median $\approx 480$ topology passes, 90th
percentile in the thousands), because a noise class persists until
exploration happens to fill its one cell, so withholding funding
until a class outlives that distribution also withholds it from
genuine chambers. On Montezuma's Revenge the arm completes level~1
in 6 of its 10 seeds, against 8 of 10 for plain \TE{} on the same
seeds; on Pitfall! it is slowest to 20 rooms (\Cref{tab:pitfall}). We report it because a
plausible fix that fails under its own calibration is more
informative than omitting it.

\subsection{Chamber worlds: survival, cost, and grids}
\label{app:chambers}

\begin{figure}[p]
\centering
\includegraphics[width=\linewidth]{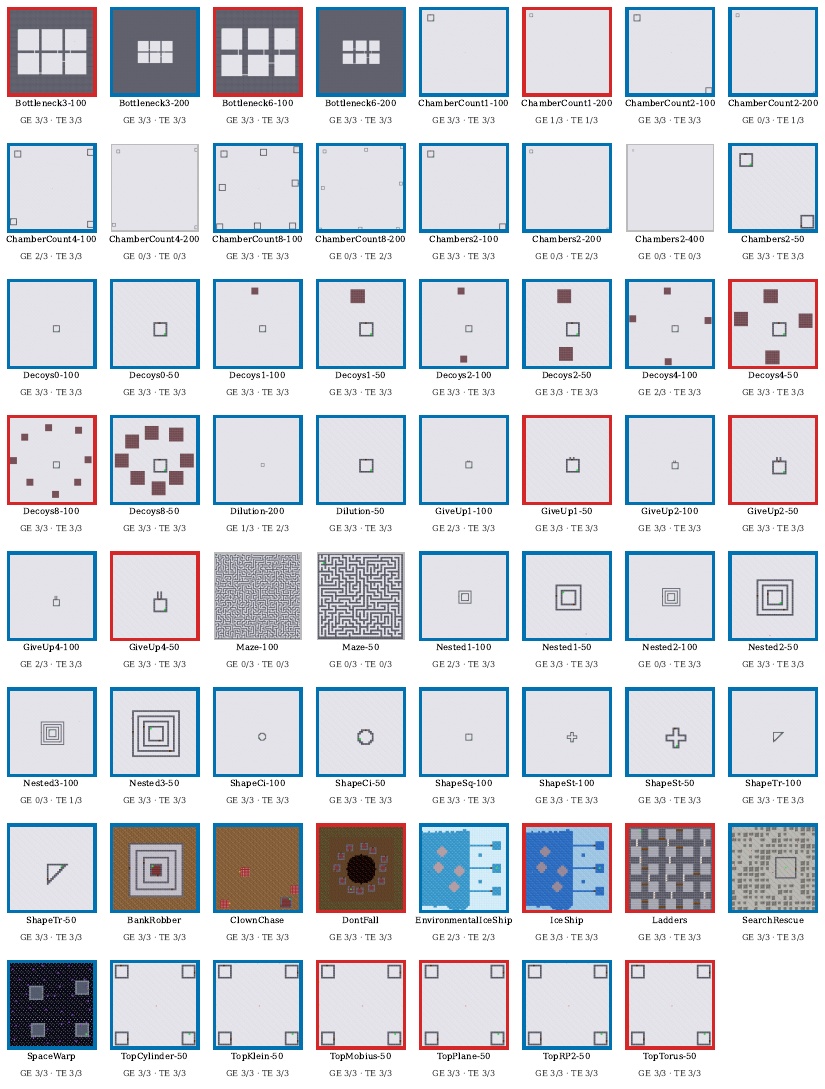}
\caption{The 63 units of the TopoGym hold-out benchmark
(\Cref{tab:holdout}), one of each unit's three layout seeds, drawn
whole at their true extent and ordered by slice (GridWorld2D,
Texture, non-trivial topologies) and family. Under each: the number
of its three seeds in which \GE{} and plain \TE{} found the goal.
Frame: who found it first, i.e.\ the method that solved more seeds or,
on a tie, the lower median steps-to-goal over the seeds both solved;
blue \TE{} (45 units), red \GE{} (14), grey neither solved any
seed (4).}
\label{fig:app:holdout_gallery}
\end{figure}

\begin{figure}[h]
\centering
\begin{minipage}[c]{0.52\linewidth}
\centering
\includegraphics[width=\linewidth]{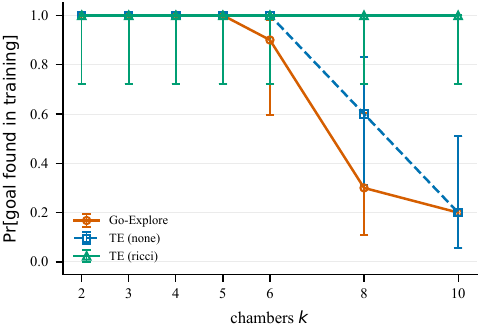}\\[2pt]
\includegraphics[width=\linewidth]{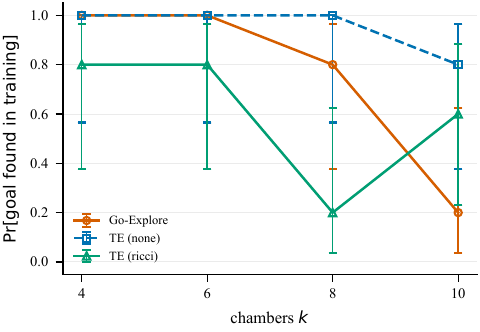}
\end{minipage}\hfill
\begin{minipage}[c]{0.44\linewidth}
\centering
\includegraphics[width=\linewidth]{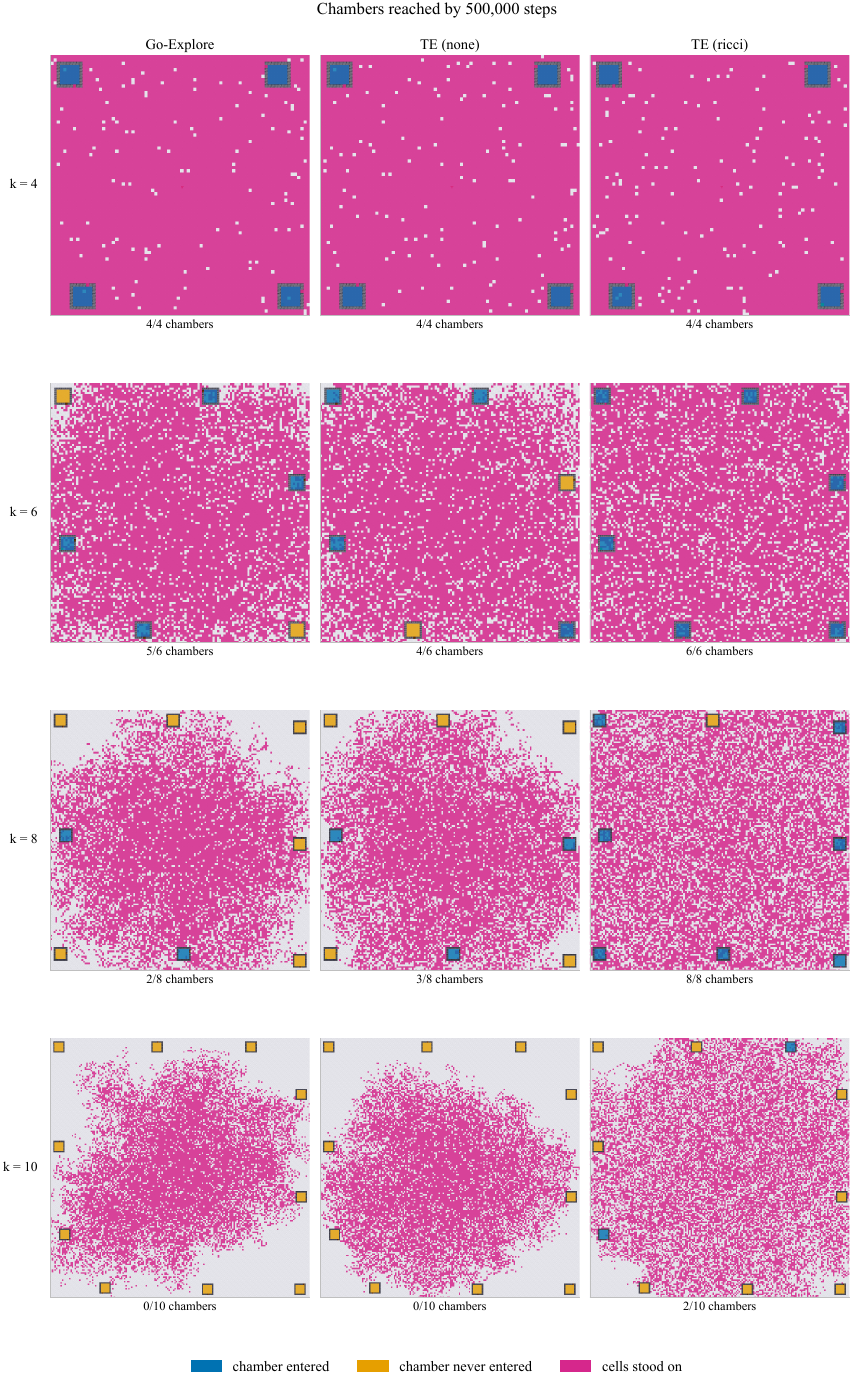}
\end{minipage}
\caption{Left, top and bottom: probability of finding the goal during
training against $k$ on EnlargedChamberCount and
OpenFieldChamberCount (Wilson 95\% CIs). Right: which rooms each
method had opened at half budget on EnlargedChamberCount (layout seed
0 at each $k$, 500k of 1M steps): blue chambers were
entered, amber never were, magenta marks cells stood on. At $k = 8$
\GE{} and plain \TE{} have opened 2 and 3 of 8 chambers and their
visited sets are still a disc around the start; the curvature arm has
opened all 8 and reached the walls. One row is one world and
illustrates the statistic; the claim lives in \Cref{fig:ksweep}.}
\label{fig:grid}
\end{figure}

\begin{figure}[h]
\centering
\includegraphics[width=0.32\linewidth]{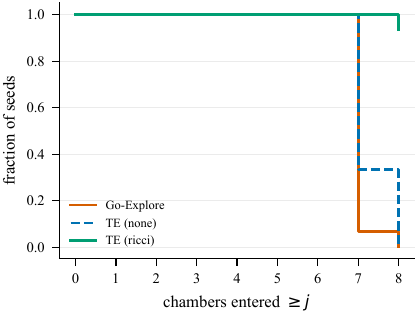}
\includegraphics[width=0.32\linewidth]{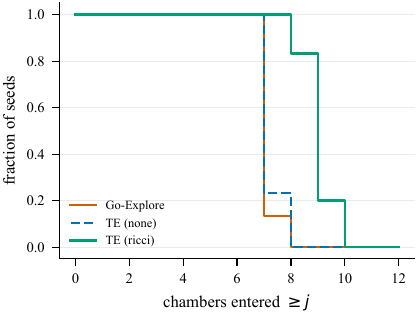}
\includegraphics[width=0.32\linewidth]{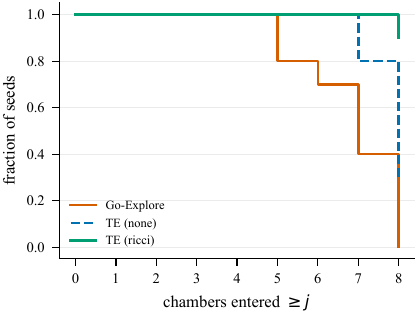}\\
\includegraphics[width=0.32\linewidth]{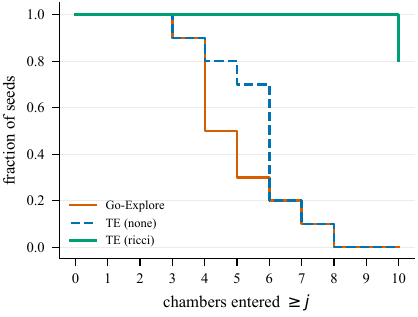}
\includegraphics[width=0.32\linewidth]{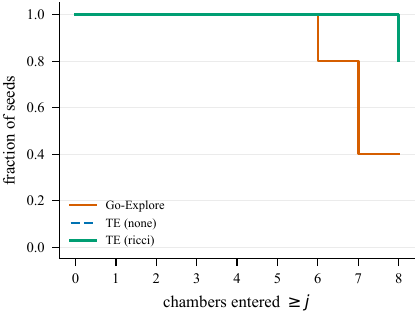}
\includegraphics[width=0.32\linewidth]{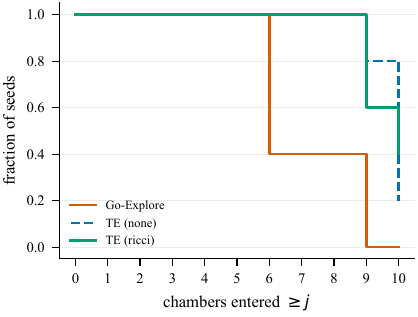}
\caption{Survival curves at the high $k$ values: fraction of seeds
with at least $j$ chambers entered at budget end. Top: EpicChase
$k = 8$ and $12$ ($n = 30$), EnlargedChamberCount $k = 8$ ($n = 10$).
Bottom: EnlargedChamberCount $k = 10$, OpenFieldChamberCount $k = 8$
and $10$ ($n = 5$, 2M steps).}
\label{fig:app:survival}
\end{figure}

\begin{figure}[h]
\centering
\includegraphics[width=0.32\linewidth]{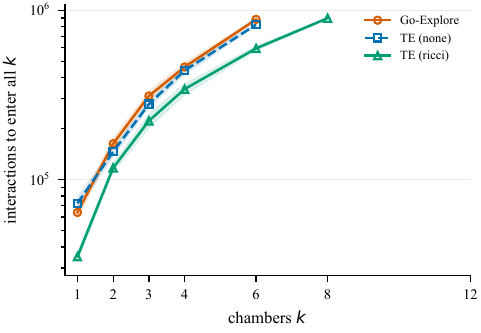}
\includegraphics[width=0.32\linewidth]{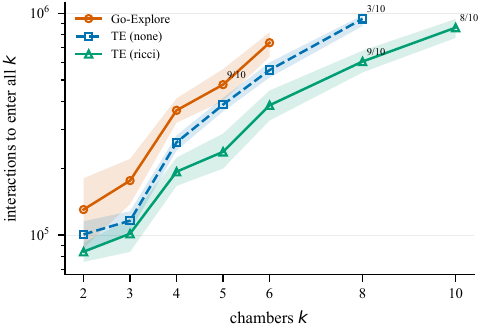}
\includegraphics[width=0.32\linewidth]{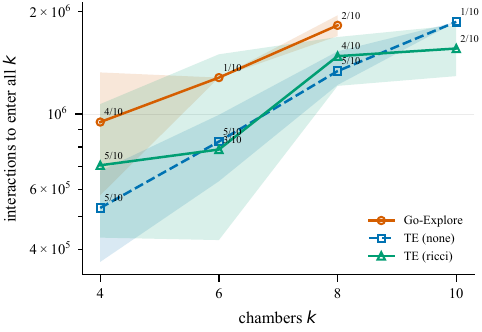}\\
\includegraphics[width=0.32\linewidth]{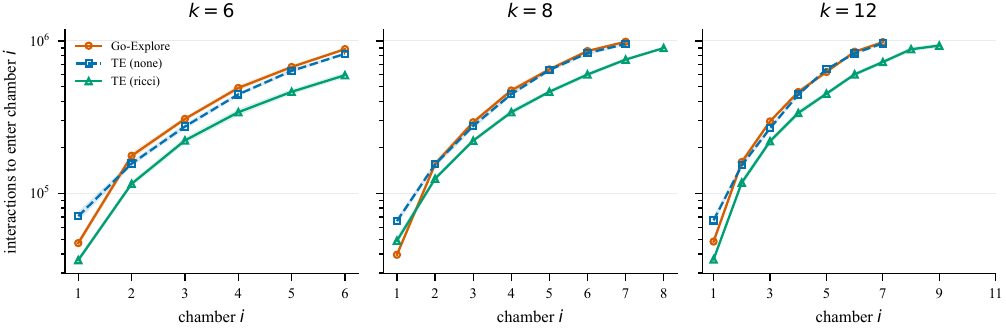}
\includegraphics[width=0.32\linewidth]{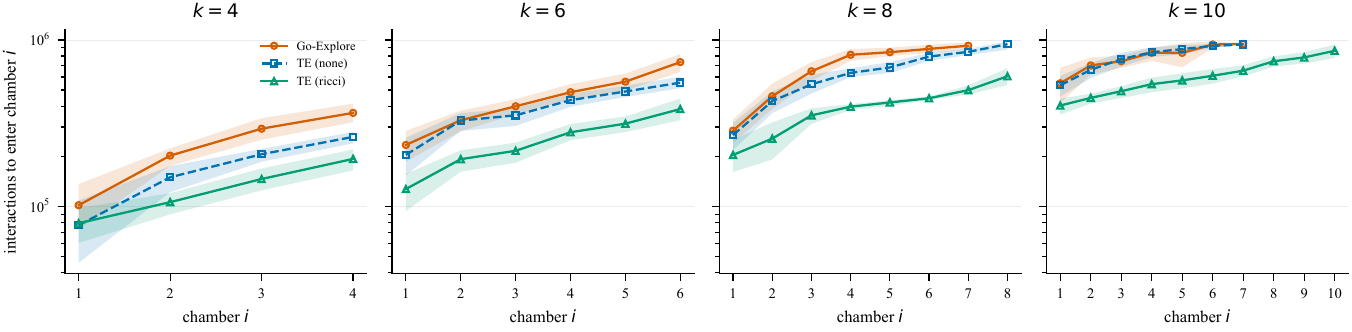}
\includegraphics[width=0.32\linewidth]{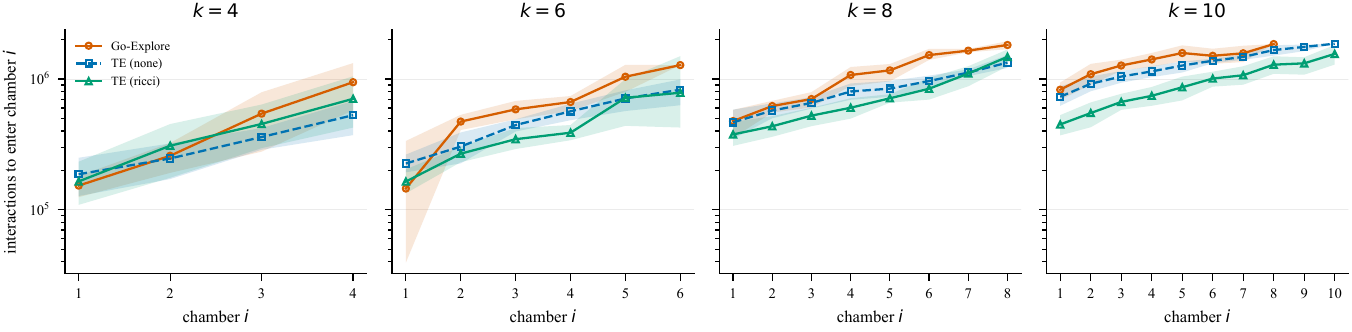}
\caption{Cost figures for EpicChase, EnlargedChamberCount, and
OpenFieldChamberCount (left to right). Top: interactions to enter all
$k$ chambers against $k$, successful seeds only, annotated with
success counts. Bottom: interactions to enter the $i$-th chamber
against $i$, one panel per $k$; where a method's curve bends upward
is where it stops.}
\label{fig:app:cost}
\end{figure}

\begin{figure}[h]
\centering
\includegraphics[width=0.48\linewidth]{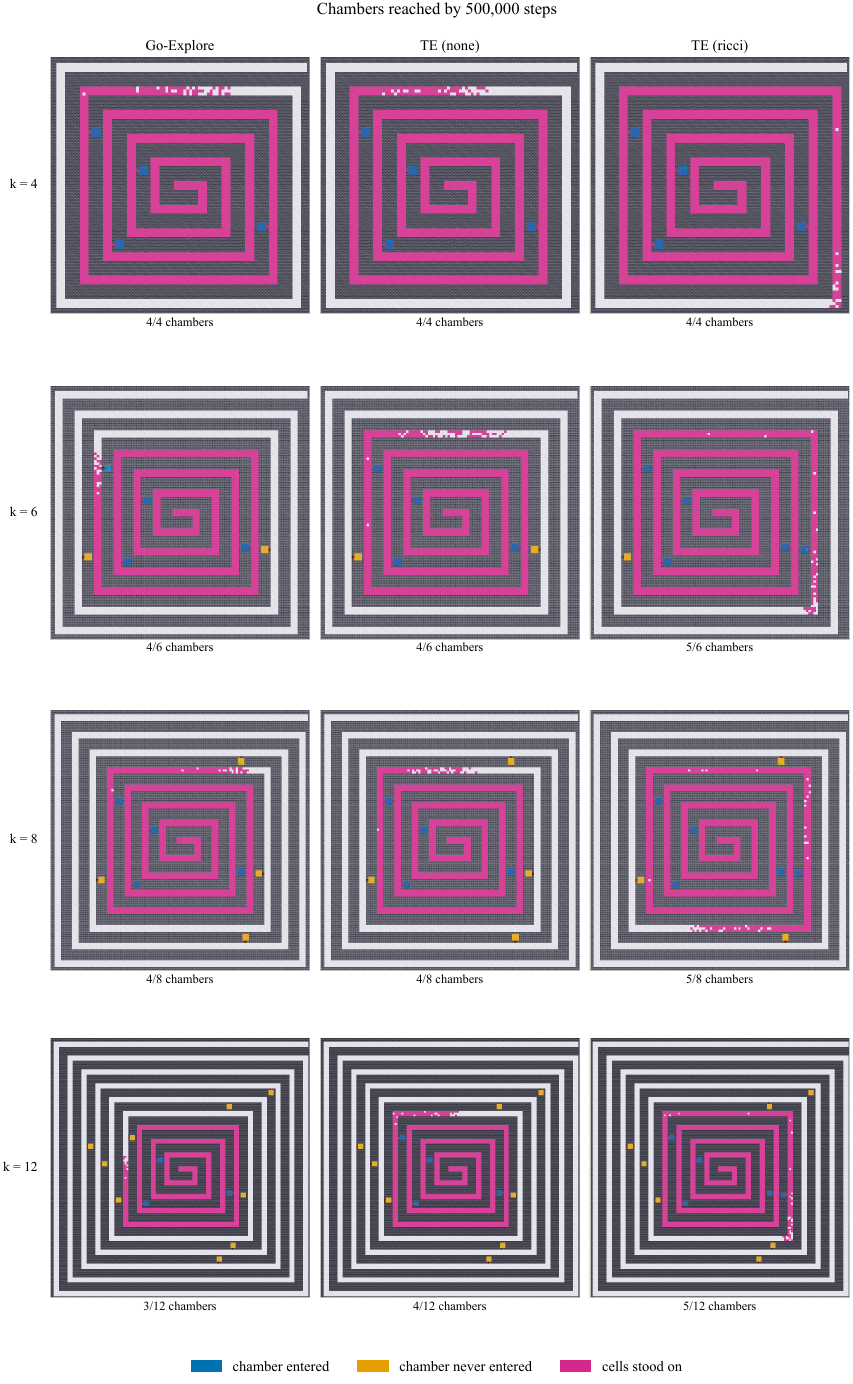}
\includegraphics[width=0.48\linewidth]{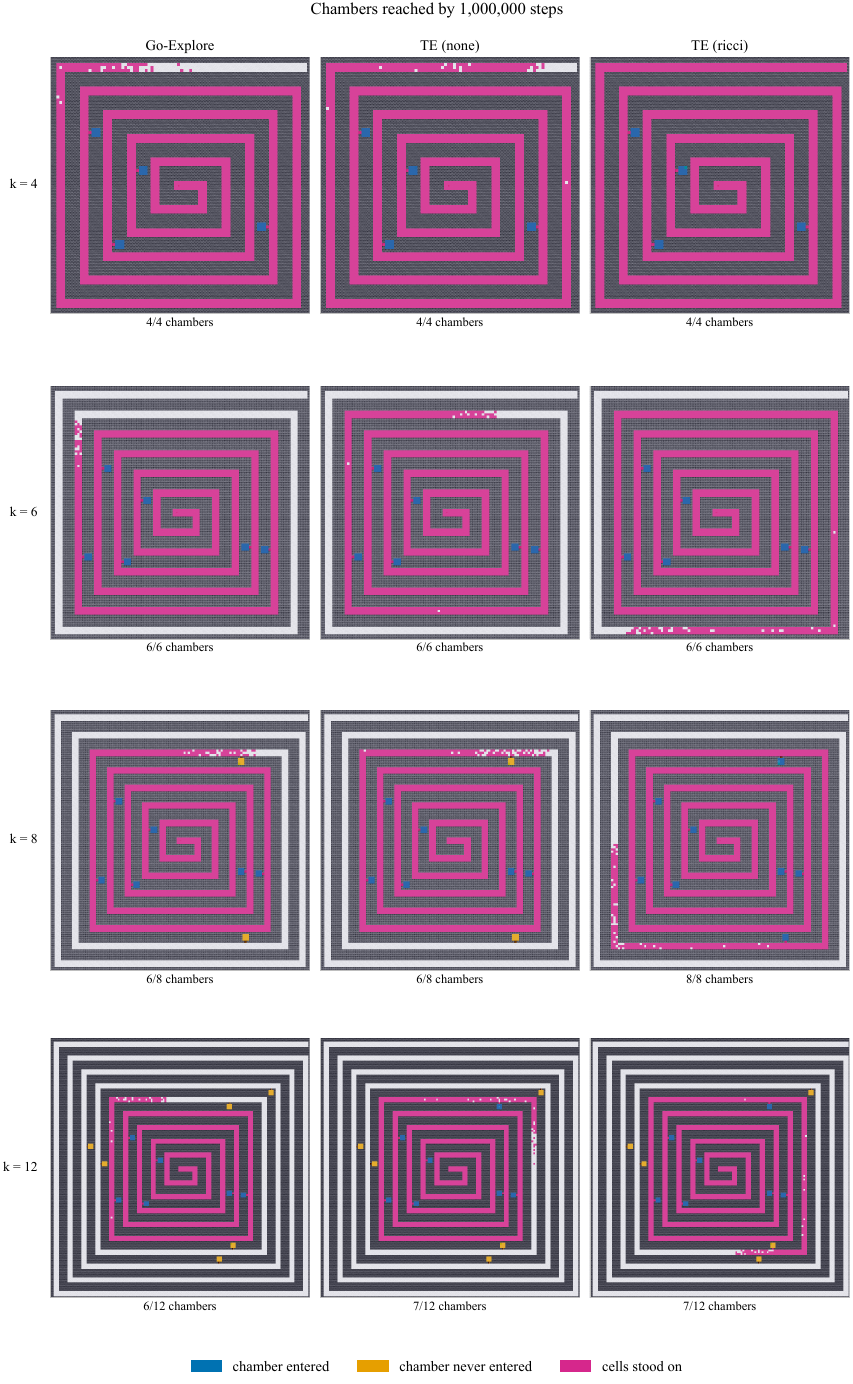}\\
\includegraphics[width=0.48\linewidth]{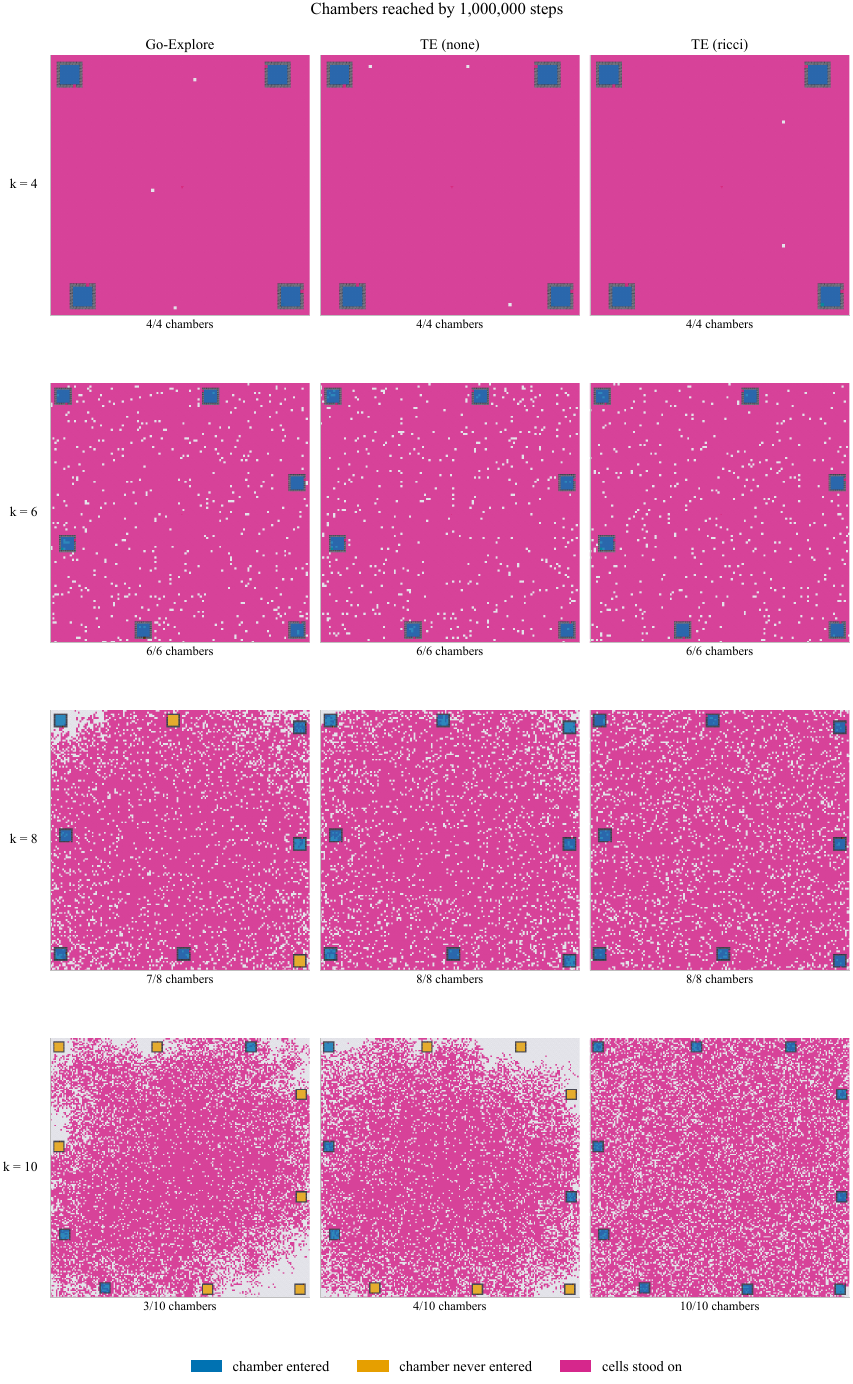}
\includegraphics[width=0.48\linewidth]{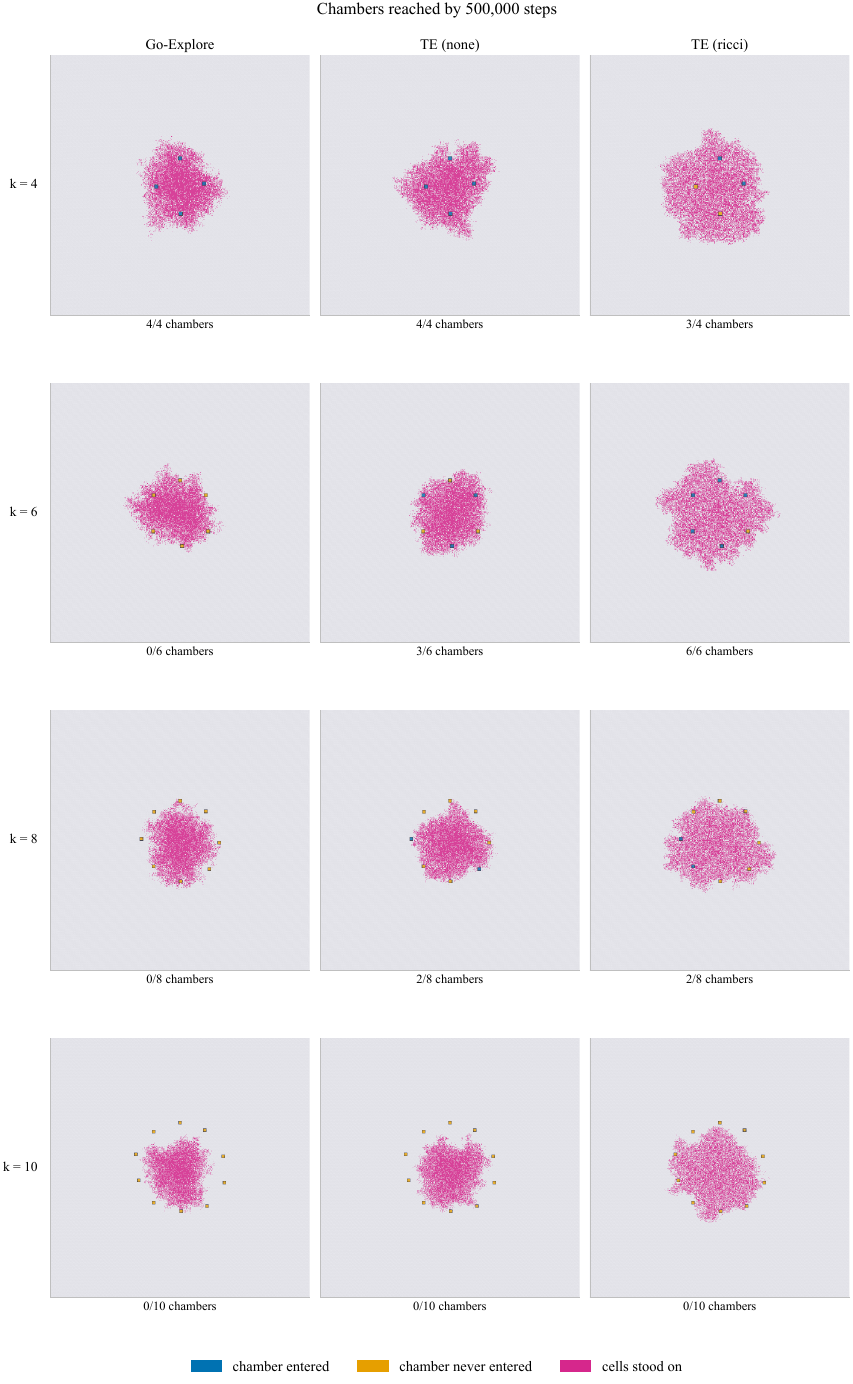}
\caption{Chamber grids. Top: the EpicChase spiral at 500k
and 1M steps (seed 0; the spiral is layout-seed invariant, so one
panel per $k$ is the family). Bottom left: EnlargedChamberCount at
the full 1M budget. Bottom right: OpenFieldChamberCount at
500k steps. Blue chambers were entered, amber never were;
magenta marks cells stood on.}
\label{fig:app:grids}
\end{figure}

The $k$-sweep suites behind \Cref{fig:ksweep} emit the same set of figures for every family; the main text shows
only the all-$k$ probability panels. \Cref{fig:grid} gives the
goal-found curves and a chamber grid, \Cref{fig:app:survival} gives the
survival curves at the high $k$ values (fraction of seeds with at
least $j$ chambers at budget end), \Cref{fig:app:cost} the
interactions to enter all $k$ (successful seeds only) and the
per-chamber cost curves, whose bend locates where each method dies,
and \Cref{fig:app:grids} the remaining chamber grids, including the
EpicChase spiral. Every statistic plotted, with its confidence
interval and $n$, is in \texttt{figures/topogym/*/suite\_stats.csv};
the per-world outcome behind every cell of \Cref{tab:holdout} is in
\texttt{figures/topogym/holdout/holdout\_goal\_found\_by\_world.csv},
and \Cref{fig:app:holdout_gallery} shows every unit with its outcome.

\subsection{Cost against archive size}
\label{app:cost_archive}\label{sec:results:arms}\label{sec:results:baselines}\label{sec:results:compute}

The gradient baselines spend their whole budget on policy learning;
the archival methods use it on discovery and turn what they find into a policy afterwards
(Phase~2, \Cref{sec:method:algorithm}, not analysed here). The
topological pass costs 16--21\% of wall-clock at every iteration on
archives of $\sim$6{,}000 cells, a ceiling since the cadence is
configurable. On Montezuma's Revenge RND and ICM required
$\sim$7.5--8.3 GPU-hours per seed to reach 1--3 rooms; a \TE{} seed
reached 24 rooms in $\sim$8.0 CPU-hours.

\begin{figure}[H]
\centering
\includegraphics[width=\linewidth]{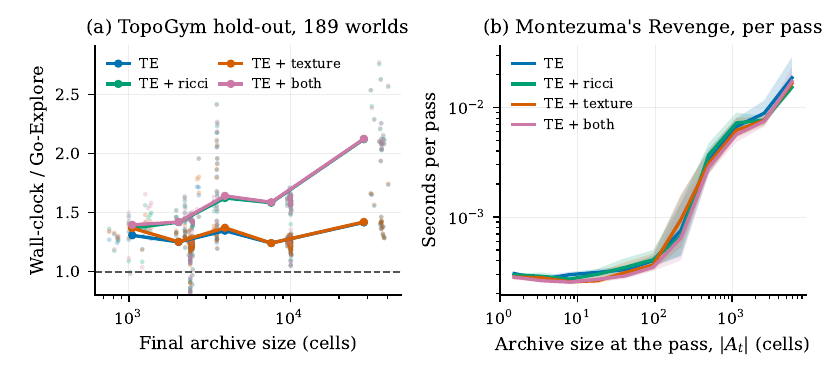}
\caption{What the topological pass costs and how the cost scales.
(a) Hold-out benchmark, 189 worlds: wall-clock of each \TE{} arm
divided by \GE{}'s on the same world (one point per world, medians
per archive-size bin), against the archive size the run ended with.
The texture arm coincides with the plain term and the both arm with
the curvature arm. (b) Montezuma's Revenge, every topology pass of
every Phase-1 seed: seconds per pass against the archive size at that
pass, medians with interquartile bands per bin, log axes.}
\label{fig:app:cost_archive}
\end{figure}

\begin{table}[H]
\centering
\small
\caption{Overhead of each arm. Left: wall-clock relative to \GE{} on
the 189 hold-out worlds, the median with its interquartile range and
the medians over the worlds in the smallest and largest quartile of
final archive size (756 to 2,392 and 9,946 to 39,560 cells). Right:
the log-log slope $b$ of per-pass latency against archive size on
Montezuma's Revenge over the passes with at least 100 cells.}
\label{tab:app:cost_archive}
\begin{tabular}{lcccc}
\toprule
Arm & \multicolumn{3}{c}{Wall-clock / \GE{}, 189 worlds} & Pass latency $\propto |\archive|^{\,b}$, $|\archive| \ge 100$ \\
\cmidrule(lr){2-4}\cmidrule(lr){5-5}
 & median [IQR] & smallest quartile & largest quartile & $b$ \\
\midrule
TE & 1.27 [1.22, 1.32] & 1.28 & 1.27 & 0.82 \\
TE + ricci & 1.49 [1.30, 1.62] & 1.28 & 1.61 & 0.80 \\
TE + texture & 1.27 [1.23, 1.34] & 1.29 & 1.27 & 0.76 \\
TE + both & 1.50 [1.31, 1.63] & 1.28 & 1.62 & 0.93 \\
\bottomrule
\end{tabular}

\end{table}

\Cref{fig:app:cost_archive} and \Cref{tab:app:cost_archive} answer
two questions the compute paragraph of \Cref{sec:results:compute}
leaves open: what the pass adds over \GE{} when both run the same
world, and whether that addition grows with the archive.

\paragraph{Per world, against \GE{}.}
On the hold-out benchmark every archive arm ran on every world in the
same shard, one after the other, so dividing an arm's wall-clock by
\GE{}'s on the same world pairs out the machine's load; the runs
shared an 18-core machine with 16 shards, so absolute times are only
indicative (\GE{} took a median of 210 seconds per world, interquartile
range 155 to 350) and are shipped in
\texttt{figures/topogym/holdout/cost\_vs\_archive\_topogym.csv}. The
plain term costs a median $1.27\times$ \GE{}'s wall-clock
(interquartile range 1.22 to 1.32, ninetieth percentile 1.42, worst
world 2.11), and the ratio is flat in archive size: 1.28 on the
quartile of smallest worlds and 1.27 on the quartile of largest, from
about 800 to about 40,000 cells. The texture arm costs the same as
the plain term, since a slot is a per-cell attribute read at scoring
time and adds no pass of its own. The curvature arm costs a median
$1.49\times$ and its ratio does grow, from 1.28 on the smallest
quartile to 1.61 on the largest (worst world 2.76), and the both arm
tracks it. In steps per second the plain term runs at 3,510 against
\GE{}'s 4,439 (medians over worlds), the curvature arms at about
3,470.

\paragraph{Per pass, against the archive.}
The flat ratio of the plain term is not because the pass is cheap in
the limit; it is because \GE{}'s own iteration is already linear in
the archive. Each iteration draws a cell from the whole archive, so
\GE{} pays $O(|\archive|)$ per iteration, and panel (b) shows the pass
paying about the same order. Over the 5.1 million passes of the 16
plain-term seeds on Montezuma's Revenge, latency sits on a floor of
about 0.3\,ms while the archive is below 100 cells (the fixed cost of
building the complex), then grows as $|\archive|^{\,b}$ with
$b = 0.82$ (0.76 to 0.93 across arms): 0.75\,ms at about 200 cells,
3.3\,ms at 500, 6.7\,ms at 1,100, and 19\,ms at about 6,000, the
largest archives a Phase-1 seed reaches. At the pass cadence this is
the 14 to 21\% share of wall-clock of \Cref{sec:results:compute}.
Pitfall! gives the same curve (0.25\,ms floor, 20\,ms at about 4,000
cells; \texttt{cost\_vs\_archive\_pass\_pitfall.csv}). A pass whose
latency grew faster than the archive would show as a rising ratio in
panel (a); the plain term's does not, and the curvature arm's, whose
Ollivier--Ricci computation touches every edge of the archive graph,
does. Both figures are produced by
\texttt{analysis/iclr/cost\_vs\_archive.py} from the hold-out result
files and the epoch logs.

\section{Reproducibility}
\label{app:repro}

The method is stated in full in \Cref{alg:te}, with every
hyperparameter, backend and class-identity rule in \Cref{app:impl};
the theory's model, assumptions and complete proofs are in
\Cref{app:theory}. The hold-out benchmark is a public, frozen split
whose tuning and test worlds are disjoint (\Cref{sec:setup}); the
Montezuma test's hypothesis, seeds, statistic and decision rule are
stated in \Cref{app:montezuma}; the
embodied pilot's per-seed evaluation rows, trajectory pool recipe and
training configuration are in \Cref{app:embodied}.

Every figure and table in this paper is generated from the logs
written during training or from the per-run summaries derived from
them; no number is transcribed by hand. The paper-level script
\texttt{papers/iclr/reproduce.sh} has tiered targets:

\begin{itemize}[leftmargin=1.6em,itemsep=1pt]
  \item \texttt{setup} --- virtual environment and dependencies,
    including the TopoGym release the paper used;
  \item \texttt{tables} --- every table from the committed per-run
    summaries, with no training;
  \item \texttt{figures} and \texttt{topogym-figs} --- every figure
    from the released logs, with no training;
  \item \texttt{check} --- verifies every number the prose quotes
    against the generated tables and summaries;
  \item the campaigns themselves, per domain: \texttt{runs},
    \texttt{baselines} and \texttt{phase2} in
    \texttt{experiments/\{montezuma,pitfall\}/reproduce.sh}, resumable,
    so that any (environment, method, seed, budget) combination
    already complete is skipped and an interrupted run costs only the
    jobs in flight;
  \item \texttt{experiments/topogym/topogym\_benchmarks.sh} --- the
    TopoGym studies behind \Cref{tab:holdout} and
    \Cref{fig:ksweep,fig:grid}: the 189-world hold-out benchmark, the
    three chamber-count $k$-sweeps, the chamber grids, their figures,
    and the port into the paper tree. Every shard's exact argv and the
    TopoGym commit it ran at are shipped as run manifests, and the
    script replays those argv verbatim; \Cref{tab:holdout} is
    recomputed from the episode logs by
    \texttt{analysis/topogym/holdout\_table.py} (verified identical to
    the shipped CSVs);
  \item \texttt{figures} also runs the Montezuma diagnostics over the
    same episode rows and archives:
    \texttt{analysis/montezuma\_door\_draws.py} (\Cref{tab:door_draws}),
    \texttt{analysis/montezuma\_regime.py} (\Cref{fig:regime}) and \texttt{analysis/montezuma\_room\_heatmap.py}
    (\Cref{fig:app:mz_map}; re-renders rooms in the ALE, on the room
    layout \texttt{analysis/montezuma\_room\_layout.py} learns from
    the step logs), with
    \texttt{analysis/montezuma\_where\_topo\_helps.py} printing the
    paired frames-to-$k$ tests quoted in \Cref{sec:results:montezuma},
    \texttt{analysis/montezuma\_rooms\_wave.py} the room-scale
    wave's tables (\Cref{tab:rooms_wave,tab:rooms_wave_paired,tab:room_term,tab:room_term_b}), and
    \texttt{analysis/montezuma\_room12.py} \Cref{fig:room12} from
    the first-visit events of every run that logged them;
  \item the embodied pilot (\Cref{tab:embodied}) lives in a separate
    repository whose recipes R14--R18 name the trajectory pool, the
    arm lists, the training recipe, and the evaluation quotas;
    \texttt{analysis/embodied/pilot\_table.py} recomputes the table
    from the per-episode evaluation rows, the composite and
    training-example figures come from that repository's
    \texttt{make\_pilot\_figures.py} and
    \texttt{make\_n1\_paper\_assets.py}, and the per-seed summaries,
    source commit, and handoff notes ship under
    \texttt{figures/embodied/}.
\end{itemize}

Seeds, budgets, and hyperparameters are pinned in code. Archival
runs are deterministic given a seed (deterministic emulator, seeded
selection), and the determinism survives a change of platform: a
200{,}000-step run of the curvature arm executed on x86-64 Linux
reproduces the arm64 macOS run of the same seed exactly, agreeing on
the action taken and the resulting cell at every one of the 200{,}000
steps. Seeds produced on either platform are therefore poolable, which
is what lets the seed extension run on cloud CPU. The learned
baselines are not bit-deterministic on GPU,
so every training episode logs its environment reset seed and action
sequence, making any logged episode exactly replayable. Runs persist
their archives, so trajectories, demonstrations, and per-room
first-entry frames are recoverable after the fact rather than only
during the run.

\paragraph{Metrics and hardware.}
The evidence throughout is training-time exploration: frames to the
$k$-th distinct room, right-censored at the budget and always
reported with the fraction of seeds that reached $k$; completion;
mechanism logs (live and pruned classes, the realised share
$\delta_t$, attempt counters); and cost. PPO+RND and PPO+ICM use dual
value heads, non-episodic normalised intrinsic returns, observation
whitening and sticky actions, as in their reference implementations.
The Atari and embodied campaigns ran on one 18-core Apple-silicon
workstation, with archival methods CPU-bound and PPO and Phase~2 on
its GPU; the seed extensions and the TopoGym waves ran on cloud CPU
(\texttt{cloud/README.md}).

\clearpage  
\section{AI use statement}
\label{app:ai}
We used Claude (Anthropic), operated by the authors through an agentic
coding interface, for tasks with required disclosure: to help develop
the theoretical model of \Cref{app:theory} by turning the claims the
authors intended to prove into concrete candidate statements (the
proofs themselves were written by the authors), to give feedback on
experimental design, and to implement the method and the analysis
code. We did not use it to generate synthetic data, to prove
mathematical claims, or to interpret results; translation, data
cleaning and reformatting, and qualitative or thematic analysis are
not applicable to this work. Additionally, for tasks with recommended
disclosure, it drafted and edited text, proposed the paper's
structure, created and modified figures, suggested experimental
parameters, and identified and formatted references. We have reviewed
all AI-assisted work: the authors checked every proof, reviewed and
tested all code, verified references against the cited venues, and
recomputed every reported number from the run logs. We take
responsibility for the final content of this work, including text,
claims and artifacts produced with the aid of generative AI.

\end{document}